\documentclass[lettersize,journal]{IEEEtran}
\usepackage{amsmath,amsfonts}
\usepackage{algorithmic}
\usepackage{algorithm}
\usepackage{array}

\usepackage[caption=false,font=normalsize,labelfont=sf,textfont=sf]{subfig}
\usepackage{textcomp}
\usepackage{stfloats}
\usepackage{url}
\usepackage{verbatim}
\usepackage{graphicx}

\def\BibTeX{{\rm B\kern-.05em{\sc i\kern-.025em b}\kern-.08em
    T\kern-.1667em\lower.7ex\hbox{E}\kern-.125emX}}
\usepackage{balance}

\usepackage{cite}
\usepackage{multirow}
\usepackage{makecell}
\usepackage{booktabs}
\usepackage{multirow}
\usepackage{booktabs}

\usepackage{hyperref}

\usepackage{soul}
\usepackage{xcolor}
\usepackage{amssymb}
\usepackage{arydshln}
\usepackage{threeparttable}

\usepackage{chapterbib}
\usepackage{docmute}

\begin{document}

\title{CrowdTransfer: Enabling Crowd Knowledge Transfer in AIoT Community}

\author{Yan~Liu,
        Bin~Guo\IEEEauthorrefmark{1},~\IEEEmembership{Senior Member,~IEEE,}
        Nuo~Li, \\
        Yasan~Ding, 
        Zhouyangzi Zhang,
        and~Zhiwen~Yu,~\IEEEmembership{Senior Member,~IEEE}

\thanks{
This work was supported in part by the National Science Fund for Distinguished Young Scholars under Grant 62025205, in part by the National Natural Science Foundation of China under Grant 62032020, in part by the Young Scientists Fund of the National Natural Science Foundation of China under Grant 62302017, and in part by the China Postdoctoral Science Foundation under Grant 2023M730058. (\textit{Corresponding author: Bin Guo.})}
\thanks{Y. Liu, B. Guo, N. Li, Y. Ding, and Z. Zhang are with the Northwestern Polytechnical University, Xi’an, Shaanxi 710129, China. (e-mail: yan\_emily@outlook.com, guob@nwpu.edu.cn, linuo@mail.nwpu.edu.cn, yasanding@mail.nwpu.edu.cn, zzyangzi29@gmail.com).}
\thanks{Z. Yu is with the Harbin Engineering University, Harbin, Heilongjiang 150001, China, and with the Northwestern Polytechnical University, Xi’an, Shaanxi 710129, China. 
(e-mail:zhiwenyu@nwpu.edu.cn).}
}

\markboth{IEEE COMMUNICATIONS SURVEYS \& TUTORIALS,~Vol.~0, No.~0, 2023}%
{Yan \MakeLowercase{\textit{et al.}}: CrowdTransfer: Enabling Crowd Knowledge Transfer in AIoT Community}


\maketitle

\begin{abstract}
Artificial Intelligence of Things (AIoT) is an emerging frontier based on the deep fusion of Internet of Things (IoT) and Artificial Intelligence (AI) technologies.
The fundamental goal of AIoT is to establish a self-organizing, self-learning, self-adaptive, and continuous-evolving AIoT system by orchestrating intelligent connections among Humans, Machines, and IoT devices. 
Although advanced deep learning techniques enhance the efficient data processing and intelligent analysis of complex IoT data, they still suffer from notable challenges when deployed to practical AIoT applications, such as constrained resources, dynamic environments, and diverse task requirements. 
Knowledge transfer, a popular and promising area in machine learning, is an effective method to enhance learning performance by avoiding the exorbitant costs associated with data recollection and model retraining. 
Notably, although there are already some valuable and impressive surveys on transfer learning, these surveys introduce approaches in a relatively isolated way and lack the recent advances of various knowledge transfer techniques for AIoT field. 
This survey endeavors to introduce a new concept of knowledge transfer, referred to as Crowd Knowledge Transfer (CrowdTransfer), which aims to transfer prior knowledge learned from a crowd of agents to reduce the training cost and as well as improve the performance of the model in real-world complicated scenarios. 
Particularly, we present four transfer modes from the perspective of crowd intelligence, including \textit{derivation}, \textit{sharing}, \textit{evolution} and \textit{fusion modes}. 
Building upon conventional transfer learning methods, we further delve into advanced crowd knowledge transfer models from three perspectives for various AIoT applications: intra-agent knowledge transfer, centralized inter-agent knowledge transfer, and decentralized inter-agent knowledge transfer. 
Furthermore, we explore some applications of AIoT areas, such as human activity recognition, urban computing, multi-robot system, and smart factory. 
Finally, we discuss the open issues and outline future research directions of knowledge transfer in AIoT community.  
\end{abstract}

\begin{IEEEkeywords}
AIoT, crowd intelligence, crowd knowledge transfer, transfer learning.
\end{IEEEkeywords}

\section{Introduction}

Internet of Things (IoT), a well-known term, refers to a vast network connecting the billions of physical devices (e.g., smartphones, vehicles, and robots) embedded with sensors and actuators throughout the world via the Internet, which enables these devices to communicate with each other \cite{gubbi2013internet, al2015internet, stoyanova2020survey}.
A range of technologies (e.g., sensor networks, wireless communication, and cloud computing.) are harnessed to achieve real-time data communication and information exchange, effectively bridging the realms of the digital and physical world.
IoT has influenced various domains, including cities, industries, transportation, healthcare, and so on. 
According to Cisco’s projections \footnote{https://www.cisco.com/c/en/us/solutions/service-provider/a-network-to-support-iot.html}, the number of IoT devices will be up to 500 billion globally by 2030. 
With the increasing number of devices connecting to IoT, it is likely to play a pivotal role in enhancing the intelligence of our world by providing a variety of intelligent services.

In recent years, the rapid advancement of artificial intelligence technologies and the growing computation capabilities of IoT devices have accelerated the rapid growth of the IoT. This, in turn, has given rise to the promising emergence of a new frontier: \textbf{Artificial Intelligence of Things (AIoT)} \cite{zhang2020empowering, liu2021adaspring, liu2023AIoT}. 
Unlike traditional IoT, AIoT aims to establish a comprehensive and intelligent connection between Humans, Machines, and IoT devices by combining advanced artificial intelligence techniques and IoT techniques, to improve the quality and efficiency of service management with minimal human intervention. 
Especially, IoT excels in establishing extensive connectivity for millions of physical devices to collect multi-source data, while AI techniques are harnessed to analyze and extract valuable knowledge from massive data for sophisticated data processing and intelligent decision-making. 
Consequently, the deep fusion of AI and IoT will bring various potentials in ubiquitous sensing, collaborative computing, distributed learning, and effective decision-making, to enhance more intelligent services for a wide range of applications, including smart cities, intelligent manufacturing, etc.

Generally, AIoT primarily consists of three components to enable real-time data processing and efficient information extraction. 
\textit{Embedded computing module} 
deploys various IoT devices (e.g., robots, wearable devices, and smart vehicles) to collect sensing data and perform tasks. 
\textit{Edge computing module} 
processes the obtained data on edge devices located close to the terminals to reduce latency and provide real-time services. 
\textit{Cloud computing module}
 integrates real-time data streams from IoT devices and edge devices, and further facilitates a variety of services based on vast amounts of data and abundant computing resources. 
For example, IoT devices first collect different types of data (environmental data, business data, etc.) in real-time, and then intelligently process and analyze them through data mining and machine learning methods at terminals, edge devices, or cloud servers.
Compared to traditional IoT where the data are mainly processed on cloud servers, in the era of AIoT, each device from cloud servers and edge network nodes to IoT devices can participate in the process of sensing, computing, and decision-making.

Deep learning \cite{lecun2015deep}, one of the most popular AI techniques, has achieved great success in many fields, such as computer vision \cite{voulodimos2018deep} and natural language processing \cite{otter2020survey}. 
Specifically, deep learning is based on artificial neural networks with single or multiple layers \cite{schmidhuber2015deep}, which is capable of learning complex patterns and higher-level representations within data. 
For AIoT, utilizing deep learning models enables the efficient data processing and analysis of complex data to extract valuable knowledge. 
However, due to the unique characteristics of AIoT scenarios and the complexity of neural networks, deploying deep learning models for practical AIoT applications still faces notable challenges:

\begin{itemize}

    \item \textbf{Constrained resources:}
    With the popularity of a large number of smart devices with computing capabilities, deep learning models are gradually decentralized from the cloud to the edge servers or terminal devices in AIoT, which provides safer and more convenient smart services to humans on wearables, mobile devices, and other embedded devices. 
    However, the resources (e.g., computing and storage) of IoT devices are often limited, making it difficult to run complex deep learning models.

    \item \textbf{Dynamic environments:}
    Deployment environments of each IoT device in the real-world scenario are different and dynamic (e.g., design style and weather conditions), leading to distribution discrepancies between training and testing data. 
    Generally, it is infeasible to collect all potential data due to the considerable cost and complicated environments, thus the model performance tends to suffer from sharp degradation in deployment environments.

    \item \textbf{Incremental tasks:}
    Some practical applications deploy a sequence of tasks continuously over a period of time, such as intelligent manufacturing and autonomous driving. All these tasks should be completed incrementally because they could share some similarities. Therefore, IoT devices not only need to adapt to new tasks, while avoiding forgetting knowledge related to previous tasks.
    
\end{itemize}

Knowledge transfer \cite{pan2009survey} is a promising learning methodology to improve the performance of learning by transferring knowledge across domains, which has the potential to solve the problem mentioned above.
The concept of transfer learning may initially come from educational psychology. 
For example, a person who has learned the violin can learn the piano faster than others, since both the violin and the piano are musical instruments and may share some common knowledge. 
At present, transfer learning techniques have been widely used in many real-world applications. 
For example, knowledge from a data-sufficient corpus can be transferred to the data-sparse corpus, and the pre-trained image classification model can be fine-tuned for different backgrounds.

It is worth mentioning that while some transfer learning methods have exhibited remarkable achievements in some tasks, the above-mentioned challenges still cannot be tackled effectively due to a lot of complicated factors in real-world AIoT applications.
Therefore we present a new concept of knowledge transfer for AIoT, named \textbf{Crowd Knowledge Transfer (CrowdTransfer)}, which transfers prior knowledge learned from a crowd of AIoT agents to a target AIoT agent to reduce the cost of learning as well as improve the model performance. 
Specifically, the AIoT agent embodies a ubiquitous physical object with the capability to execute deep learning models based on the collected data by its own hardware resources. 
Different from general IoT devices, the AIoT agent integrates sensing, communication, storage, and computational capabilities into a unified entity to comprehend semantic information from the surrounding environment and engage in inferential decision-making.
In real-world AIoT scenarios, a lot of factors could affect the performance of the model deployed on the AIoT agent. 
To account for key factors affecting AIoT agents, we categorize AIoT contexts into three aspects: computation context, sensing context, and task context.
\textit{In general, the objective of CrowdTransfer is to facilitate the self-learning, self-adaptive, and continuous-evolving AIoT agent for a variety of AIoT applications under complex AIoT contexts.} 
In this paper, we introduce four transfer modes to delve into how crowd knowledge is transferred among AIoT agents to achieve crowd intelligence in AIoT systems, including derivation, sharing, evolution and fusion modes. 
Furthermore, we propose a general framework of CrowdTransfer to illustrate the key transfer techniques in AIoT applications, such as intra-agent knowledge transfer, decentralized inter-agent knowledge transfer, and centralized inter-agent knowledge transfer

This survey aims to give readers a comprehensive understanding of knowledge transfer in AIoT community. Although there are already some valuable and impressive surveys on transfer learning, these surveys ignore the unique challenges faced by real-world scenarios and lack the recent advances of various knowledge transfer methods for AIoT applications. Fig. \ref{fig:organization_diagram} presents the overview of this survey. 
Particularly, the key contributions of this work can be summarized as follows:

\begin{itemize}

    \item Presenting a new concept of knowledge transfer for AIoT community, namely CrowdTransfer, and then proposing a generic framework of crowdTransfer. 

    \item Reviewing the state-of-art research knowledge transfer methods in AIoT field, including intra-agent knowledge transfer, decentralized inter-agent knowledge transfer and centralized inter-agent knowledge transfer. 

    \item Exploring some AIoT applications based on knowledge transfer, including human activity recognition, urban computing, connected vehicles, multi-robot systems and smart factory.

    \item Investigating the open issues and future research directions of crowd knowledge transfer in AIoT community, such as cognitive foundations, transferability measurement mechanisms, and learning in resource-constrained AIoT devices.

\begin{figure*}[!t]
  \centering
  \includegraphics[width=0.87\linewidth]{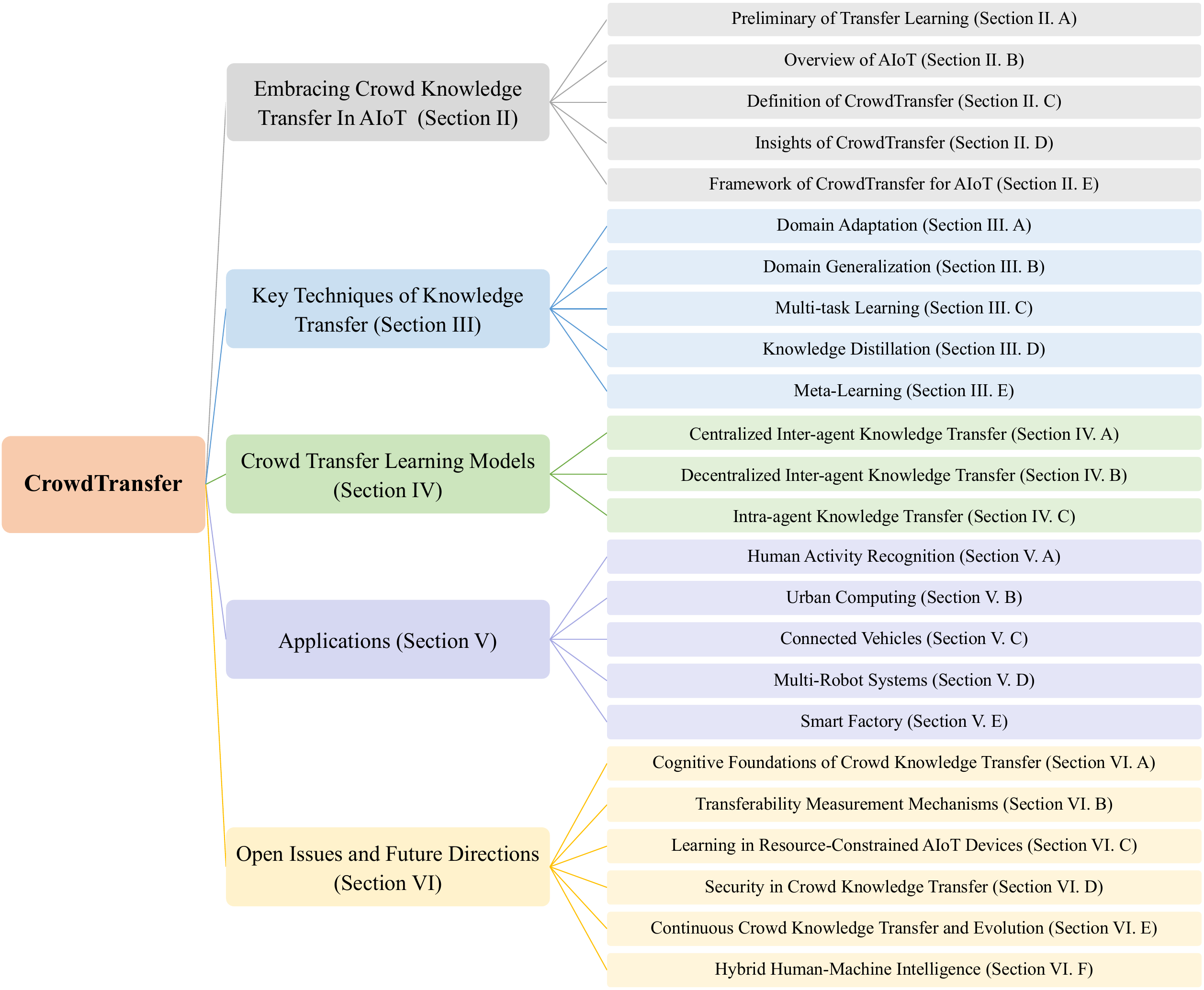}
  \caption{Overview of paper organization. }
  \vspace{-3mm}
  \label{fig:organization_diagram}
\end{figure*}

\end{itemize}

\section{Embracing Crowd Knowledge Transfer in AIoT}

In this section, we first describe the basic concept of traditional transfer learning. 
Next, we introduce an overview of the AIoT framework, which consists of the embedded computing layer, edge computing layer, and cloud computing layer. 
Furthermore, we present the definition of CrowdTransfer, a new concept of knowledge transfer, to 
empower AIoT agents with the capability of self-learning, self-adaptation, and continuous evolution across a diverse range of AIoT applications.  
In addition, we delve into the intricacies of CrowdTransfer and provide insights to explore how crowd knowledge is transferred among  AIoT agents in different scenarios through four fundamental modes: derivation, sharing, evolution, and fusion modes.
Finally, we present the general framework of CrowdTransfer for AIoT field to illustrate its key modules and transfer techniques. 
For brevity, we provide a table of notations used in our work in Table \ref{tab:Notations}.

\subsection{Preliminary of Transfer Learning}

Recently, with the rapid growth of data size and computational resources, machine learning has achieved great success in many areas. However, it has limited ability in some real-world applications where there is insufficient data to train the model. 
In addition, the model trained in one domain can only be directly utilized for another domain with the same data distribution, because many machine learning methods assume that the training and future data must be in the same feature space and have the same distribution, which may not hold in practical scenarios. Knowledge transfer or transfer learning, which aims to transfer knowledge across domains or tasks, is an effective way to solve the above-mentioned problems without much expensive data-labeling efforts.

Transfer learning \cite{pan2009survey, yang2020transfer, tan2018survey} is an important research problem in machine learning.
The objective of transfer learning is leveraging knowledge learned from one task or domain to improve the performance of a related, but different, task or domain.
This section introduces some basic definitions of transfer learning.

\newtheorem{definition}{Definition}
\begin{definition}[\textbf{Domain}]
    A domain $\mathcal{D}_o$ is composed of two components: a feature space $\mathcal{X}$ and a marginal probability distribution $P(X)$, where $X$ is the particular instance set in the feature space of all possible instances, $X=\{x_1, x_2, \dots, x_n\} \in \mathcal{X}$. Thus, the domain $\mathcal{D}_o$ can be denoted as $\mathcal{D}_o = \{\mathcal{X}, P(X)\}$.
\end{definition}

\begin{definition}[\textbf{Task}]
    A task $\mathcal{T}_a$ consists of two components: a label space $\mathcal{Y}$ and an objective decision function $f$, that is, $\mathcal{T}_a=\{\mathcal{Y}, f(.) \}$. The decision function $f$ is an implicit one, which is expected to be learned from the training data, such as a number of labeled pair $D=\{(x_1,y_1 ),\dots,(x_n,y_n )\}$, where $x_i\in \mathcal{X}$ and $y_i\in \mathcal{Y}$. From a probabilistic viewpoint, $f(x)$ can be written as conditional distributions of instances $P(y|x)$, and a task can also be defined as $\{\mathcal{T}_a, P(y|x)\}$.
\end{definition}

In general, a domain is often observed by a number of instances with or without the label data. The source domain  $\mathcal{D}^S_o$ usually contains a number of instance-label pairs, denoted as $\mathcal{D}^S=\{(x_1^S,y_1^S),\dots,(x_n^S,y_n^S)\}$. Similarly, the target domain $\mathcal{D}^T_o$ consists of the training data  $\mathcal{D}^T=\{(x_1^T,y_1^T),\dots,(x_n^T,y_n^T)\}$. Note that the observation of the target domain usually consists of unlabeled instances and/or a limited number of labeled instances.

\begin{definition}[\textbf{Transfer Learning}]
    Given a source domain $\mathcal{D}^S_o$ and learning task $\mathcal{T}^S_a$, a target domain $\mathcal{D}^T_o$ and learning task  $\mathcal{T}^T_a$, transfer learning aims to utilize the knowledge implied in $\mathcal{D}^S_o$ and $\mathcal{T}^S_a$ to improve the learning of the predictive function $f_\mathcal{T}$ in $\mathcal{D}^T_o$, where $\mathcal{D}^S_o \neq \mathcal{D}^T_o$ or $\mathcal{T}^S_a \neq \mathcal{T}^T_a$. 
\end{definition}

There are several scenarios of transfer learning according to the definition \cite{zhuang2020comprehensive}. For example, the domains $\mathcal{D}^S_o \neq \mathcal{D}^T_o$ imply that either the feature spaces are different ($\mathcal{X}^S \neq \mathcal{X}^T$) or the distributions are different ($P_s(X) \neq P_T(X)$), and the tasks  $\mathcal{T}^S_a \neq \mathcal{T}^T_a$ implies that either the label spaces are different ($\mathcal{Y}^S \neq \mathcal{Y}^T$) or conditional probability distributions are different ($P(y^S|x^S) \neq P(y^T|x^T)$). In view of the varying differences between source domain/task and target domain/tasks, some effective approaches are developed to improve the model performance accordingly, including instance-based, feature-based, parameter-based, and relational-based approaches.

\begin{table}[!t] \small
  \caption{{Description of Notation}}
  \label{tab:Notations}
  \centering
  \begin{tabular}{p{0.25\columnwidth}p{0.6\columnwidth}}
    \hline
    Symbol & Description \\
    \hline  
    $\mathcal{D}_o$      &  Domain   \\
    $\mathcal{T}_a$ & Task  \\
    $\mathcal{X}$  &  Feature space  \\   
    $\mathcal{Y}$ & Label space \\
    $P(X)$ & Marginal probability distribution \\
    $P(y|x)$  &  Conditional probability distribution       \\
    $f(\dot)$  &   Decision function     \\
    $\mathcal{A}$  &  AIoT agent    \\
    $\mathcal{H}$ &  IoT hardware      \\
    $\mathcal{D}=\{x,y\}$  &  Dataset     \\
    $\mathcal{C}^\mathcal{R}$  &  Computation context     \\
    $\mathcal{C}^\mathcal{D}$  &  Sensing context     \\
    $\mathcal{C}^\mathcal{T}$  &  Task context     \\
  \hline
\end{tabular}
\end{table}

\subsection{Overview of AIoT}

The deep fusion of the Internet of Things (IoT) and artificial intelligence techniques has given rise to a promising emerging frontier field of Artificial Intelligence of Things (AIoT) \cite{zhang2020empowering, GuoBin2020AIoT, baker2023artificial}. 
On one hand, the widespread deployment of IoT devices and the exponential growth of data collected by these devices create an opportunity for artificial intelligence to enable intelligent sensing, communication, and computing within the IoT system. 
Particularly, it also supports efficient data processing and analysis to provide more intelligent services for users, which can be referred to as \textit{AI for IoT}. 
On the other hand, the increasing prevalence of IoT applications provides vast real-world data, which could enhance the deployments of most artificial intelligence models. 
With the continuous development of embedded chips, processors, and sensors, IoT devices are endowed with enhanced capabilities for intelligent data processing. Collaborative sensing and computing among heterogeneous entities such as portable terminals (e.g., smartphones and wearables), embedded IoT devices (e.g., cameras and smart vehicles), and Internet applications (e.g., edge and cloud servers) bestow new attributes upon artificial intelligence, which can be referred to as \textit{IoT for AI}. 
In general, different from IoT, AIoT builds the comprehensive connection of Humans, Machines, and Things to enable more intelligent IoT applications and provide more efficient services.
Especially, IoT serves to establish extensive connections for hundreds of millions of physical devices to collect data, while AI algorithms are harnessed for analyzing and mining the potential patterns and strategies from massive amounts of collected data on the end devices, edge nodes, or cloud servers.

\begin{definition}[\textbf{AIoT}]
   Based on the deep fusion of artificial intelligence, edge computing, IoT, and other advanced technologies, AIoT aims to establish a more comprehensive connection and intelligent collaboration among Humans, Machines, and Things, and further empower the sensing, communication, computing, and application with  AI algorithms, to 
   achieve a self-organizing, self-learning, self-adaptive, and continuous-evolving intelligent computing system.
\end{definition}

\begin{figure*}[!t]
  \centering
  \includegraphics[width=0.77\linewidth]{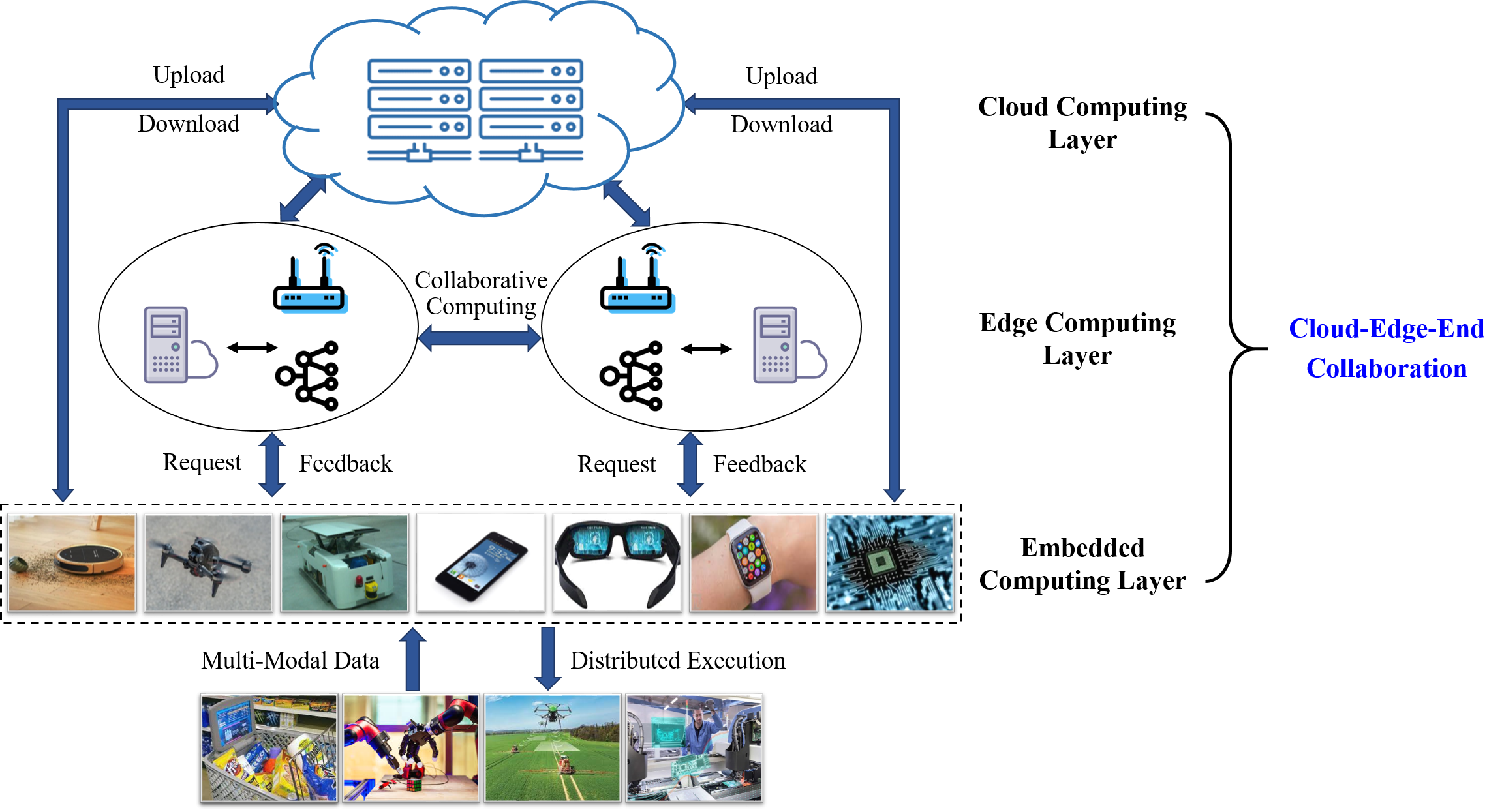}
  \caption{An overview of AIoT. }
  \vspace{-2mm}
  \label{fig:AIoT}
\end{figure*}

The core of AIoT is real-time and efficient data collection and information processing \cite{baker2023artificial}.
Benefiting from the introduction of edge intelligence into the AIoT system, AIoT forms a cloud-edge-end architecture, which consists of three layers: embedded computing layer, edge computing layer, and cloud computing layer.
The architecture of AIoT is illustrated in Fig. \ref{fig:AIoT}.
In contrast to the traditional centralized cloud-based data processing approach, AIoT systems leverage a range of computational capabilities across IoT devices, edge networks, and cloud servers distributed across various layers, enabling them to actively engage in computing tasks.
Notably, the collaboration of these three layers in the cloud-edge-end architecture not only alleviates the burden of data processing, but also enhances the efficiency of computing and real-time response on terminal and edge devices.

\begin{itemize}

    \item \textbf{Embedded Computing Layer:}
    It comprises various IoT devices embedded with sensors, processors, etc., which serve as the sensing and execution modules of AIoT. 
    In real-world environments, the embedded computing layer for data collection and intelligent analysis not only enhances the comprehensiveness of the AIoT system, but also saves manpower as well as reduces resources and costs. 
    Unlike sensing devices in traditional IoT systems only for data collection, the smart terminal layer is capable of performing some data processing tasks, and some lightweight deep learning models could be deployed at the embedded devices due to limited resources.

    \item \textbf{Edge Computing Layer:}
    It deploys the data processing and analysis tasks on edge servers near the terminals. 
    Especially, the tasks can be offloaded from terminals to edge nodes due to constrained resources and limited data, to provide real-time services by edge-end collaboration.
    In addition to enabling data transmission and data process, the edge intelligence layer also has the capabilities of balancing load, cooperating with terminals, and learning models distributively to enhance the training and inference tasks.

    \item \textbf{Cloud Computing Layer:}
    It contains the central clouds to support a variety of AIoT services based on large amounts of data.
    Similar to traditional IoT systems, collected data streams from distributed IoT devices and edge devices are transmitted over the network to remote cloud centers, where they're further integrated, processed, and stored. 
    Based on vast data and abundant computing resources, training and deploying large-scale machine learning models with good generalization performance become possible on the cloud.
    
\end{itemize}

\subsection{Definition of CrowdTransfer}

With the popularity of a large number of smart devices with sensing and computing capabilities, machine learning or deep learning models can be deployed on a diverse range of devices, including edge servers and terminal devices. 
This enables AIoT to provide safer and more convenient services to humans on wearables, mobile devices, and embedded devices.
Different from the traditional centralized learning mechanism whose performance mainly depends on the data collected in advance, the performance of learning models in AIoT applications not only depends on the real-time data obtained by devices, but also is impacted by the practical AIoT scenarios, such as the computing and storage of the device. 
To understand the main characteristics of machine learning models in the AIoT community, we present some definitions to describe the key factors that could affect the learning process of machine learning models in AIoT applications.

\begin{definition}[\textbf{AIoT Agent}]
   An AIoT agent, referred to as $\mathcal{A}$, embodies a ubiquitous object (e.g., IoT devices, edge devices, and cloud servers) with the capability to execute AI algorithms $f$ based on the collected data $\mathcal{D}=\{x,y\}$ by utilizing IoT hardware $\mathcal{H}$. The AIoT agent integrates sensing, communication, storage, and computational capabilities into a unified entity to comprehend semantic information from the surrounding environment and engage in inferential decision-making. Notably, the learning process of AIoT agent $\mathcal{A}$ can be succinctly represented as $f:\mathcal{H},x \rightarrow y$.
\end{definition}

Intuitively, \textit{hardware}, \textit{data}, and \textit{algorithms} are three key elements that typically influence the performance of the AIoT agent in complex and dynamic AIoT scenarios. 
At the hardware level, the agent encompasses sensor, memory, processor, and communication units, granting it profound sensing, communicative, storage, and computational capabilities. 
At the data level, the agent is capable of acquiring and storing diverse modalities of data from various sources, including text, images, videos, etc., wherein lies rich information about the target object. 
At the model level, the agent first trains models based on extensive data to acquire valuable knowledge, then optimizes the models during the inference stage and ultimately deploys the learned models at hardware devices.

\begin{definition}[\textbf{AIoT Context}]
   The AIoT context, denoted as $\mathcal{C}$, contains the information related to AIoT scenarios, including the sensing context involving environmental information, computation context involving hardware resources, and task context involving task-specific requirements. 
   The comprehensive context facilitates a deeper understanding of AIoT scenarios, empowering agents to make more intelligent decisions and responses. 
\end{definition}

In traditional cloud-based centralized learning frameworks, it is assumed that the training and deployment scenarios are consistent, so pre-trained models can achieve satisfactory performance during the inference phase.
Conversely, in AIoT scenarios, each device can participate in both the learning and inference processes of the model. 
However, due to the dynamic nature of AIoT contexts, cloud-based centralized learning methods struggle to achieve optimal model performance.
More specifically, the sensing context, primarily concerned with physical conditions of deployment environments (e.g., lighting, noise, etc.) and data types from diverse sensors (e.g., text, images, videos, etc.), exerts a significant influence on the domain of AIoT agents $\{\mathcal{X}, P(x)\}$.  
The computation context, mainly encompassing hardware information such as memory and processor, directly impacts the resources of AIoT agents $\mathcal{H}$. 
The task context, mainly involving task-related details such as label categories, performance requirements, etc., notably shapes the task undertaken by AIoT agents $\{y,f\}$.

Transfer learning is an effective solution to improve model performance by transferring knowledge learned from one source task to another related target task. 
For example, domain adaptation \cite{yang2020mobileda} endeavors to build a model in a source domain that can yield robust results in a target domain with different data distributions.
Multi-task learning \cite{smith2017federated}, on the other hand, seeks to learn common features across multiple tasks to effectively transfer information. 
Meanwhile, meta-learning \cite{hospedales2021meta} aims at harnessing prior knowledge derived from multiple tasks to guide learning in new tasks, effectively embracing the concept of learning to learn.
Although these transfer learning methods have demonstrated remarkable accomplishments in many tasks, the practical implementation of deep learning models for real-world AIoT applications continues to confront some challenges:

\begin{itemize}

    \item \textbf{\emph{Heterogeneous Data:}}  
   The data distributed on a large number of AIoT agents is heterogeneous, which could cause various data types and data distribution discrepancies in feature spaces. For example, the deployment environment of the model often undergoes continuous changes, which could lead to performance degradation of the pre-trained model due to varying data distributions in practical applications.

    \item \textbf{\emph{Unlabeled Data:}}  
    AIoT agents deployed in the environment typically collect a vast amount of sensing data, and obtain labeled information for these data can be challenging due to its high manual cost, such as the extensive surveillance videos. Therefore, AIoT applications often encounter the problem of sparse label data, making it difficult to train high-performing models through traditional supervised learning methods.

    \item \textbf{\emph{Real-time Data Streams:} } 
    In real-world AIoT scenarios, data is gradually generated over time instead of being instantly available in its entirety.
    However, many specialized AIoT applications require real-time responses, such as autonomous driving and online healthcare.
    It is infeasible to train the model over the entire dataset due to extended time delays.
    Therefore, AIoT agents need to facilitate online learning based on real-time data streams to quickly adapt to the current scenario.

    \item \textbf{\emph{Data Privacy:} } 
   Transmitting data among distinct AIoT agents poses a risk of exposing crucial and sensitive information. 
   While conducting local training on individual devices within AIoT can help circumvent the direct sharing of raw data, the shared model might still inadvertently disclose private information.

    \item \textbf{\emph{Incremental Tasks:} } 
    Practical applications often involve a series of tasks to be finished over time. 
    AIoT agents should have the ability to continually learn models for these tasks. 
    This entails two key aspects: firstly, effectively leveraging prior knowledge from old tasks to aid in learning new ones, and secondly, preventing the forgetting of learned knowledge from old tasks.

    \item \textbf{\emph{Mobility:} } 
    Some AIoT agents have the capability to travel in different environments to perform various tasks, such as inspection by robot dogs in multiple factories. 
    However, most environments are typically heterogeneous, and mobile AIoT agents need to facilitate the adaptability and robustness of their models to maintain consistent performance across different environments.

    \item \textbf{\emph{Limited Computing and Storage Resources:} } 
    Most embedded AIoT agents have limited resources for training and inferring deep learning models, including computing and storage resources. For example, tasks involving object detection could exhaust substantial computing resources.

    \item \textbf{\emph{Communication Bottleneck:} } 
    Some AIoT agents face a notable challenge in terms of communication bandwidth, and energy consumption when transmitting data with other agents. To mitigate communication costs, uploading learned model parameters instead of collected data becomes an option. However, it still demands a considerable allocation of communication resources, since the size of neural network parameters continues to expand.

    \item \textbf{\emph{Communication Dynamics:} } 
    The distributed AIoT agent can communicate with other agents to collaborate on complex tasks that a single agent alone cannot accomplish, such as drone swarms.
    However, the AIoT agents, within AIoT systems, typically exhibit diversity in terms of their capabilities, controllability, availability, and reliability. 
    The primary challenge lies in collaborating with multiple agents adaptively to achieve enhanced performance in view of dynamic communication patterns for varying scenarios.

\end{itemize}

In general, traditional transfer learning approaches have a lot of limitations, and it is necessary to develop an effective way to improve the performance of machine learning models in real-world AIoT scenarios.
To solve the above-mentioned challenges, we introduce a new concept of knowledge transfer for AIoT community, namely Crowd Knowledge Transfer (CrowdTransfer), which aims to transfer prior knowledge learned from a crowd of source AIoT agents to target AIoT agents to reduce the cost of learning as well as improve the model performance. 
The definition of CrowdTransfer is described as follows:

\begin{definition}[\textbf{Crowd Knowledge Transfer in AIoT}]
   Given a crowd of source AIoT agents with the computation context $\mathcal{C}^\mathcal{R}_S$, sensing context $\mathcal{C}^\mathcal{D}_S$, and task context $\mathcal{C}^\mathcal{T}_S$, and a crowd of target AIoT agents operating in contexts $\mathcal{C}^\mathcal{R}_T,\mathcal{C}^\mathcal{D}_T,\mathcal{C}^\mathcal{T}_T$, Crowd transfer learning utilizes the knowledge implied in those source agents to enhance the model performance of the target agents, where  $\mathcal{C}^\mathcal{R}_S \neq \mathcal{C}^\mathcal{R}_T$, $\mathcal{C}^\mathcal{D}_S \neq \mathcal{C}^\mathcal{D}_T$, or $\mathcal{C}^\mathcal{T}_S \neq \mathcal{C}^\mathcal{T}_T$.   
\end{definition}

CrowdTransfer goes beyond being a mere extension of transfer learning. The primary objective of CrowdTransfer is to empower AIoT agents with the ability for self-learning, self-adaptation, and continuous evolution through transferring and sharing knowledge among crowd agents, to address challenges prevalent in most AIoT scenarios, such as constrained resources, dynamic environments, and incremental tasks. 
Specifically, CrowdTransfer encompasses a variety of key techniques to transfer knowledge for a range of AIoT applications, with transfer learning being one of these techniques.

From the perspective of agents, CrowdTransfer typically involves multiple agents with the primary goal of fostering collaboration among agents to accomplish complex tasks. Unlike conventional transfer learning methods, which tend to focus on single source agent or target agent, CrowdTransfer acknowledges and leverages the interrelationships among numerous agents in real-world environments. This acknowledgment is critical in practical AIoT applications where terminal devices may struggle to develop powerful models due to limited resources and data. In such instances, CrowdTransfer can enable devices to achieve better performance by transferring knowledge from multiple available devices.

Furthermore, from the perspective of AIoT contexts, CrowdTransfer takes into consideration a variety of complex factors to enhance the adaptability and evolutionary capabilities of agents in AIoT scenarios, such as disparities in resource allocation and variations in sensing environments. Traditional transfer learning methods often focus narrowly on enhancing performance within specific target scenarios, potentially overlooking the broader applicability and generalizability of agents. In contrast, real-world applications typically involve a sequence of tasks that unfold over time, and AIoT agents should possess the ability to continually learn models for these evolving tasks. In this scenario, CrowdTransfer can utilize prior knowledge from old tasks to facilitate learning new ones and prevent the forgetting of previously acquired knowledge.

\subsection{Insights of CrowdTransfer}

We first compare the differences between CrowdTransfer and closely-related concepts.

\begin{itemize}

    \item \textbf{\emph{Collective Intelligence:}}  
    It is a broad concept, referring to the ability of a group to accomplish tasks that are beyond the capabilities of individual members through collaboration, sharing, and competition. The concept of "collective intelligence" originates from sociobiology, depicting the collective intelligent behaviors of social animals that emerge through collaboration, such as ant foraging. It emphasizes that while individual wisdom is limited, collective intelligence can emerge from the collaboration, collective efforts, and competition of many individuals. Intuitively, any group's general ability to perform a wide range of tasks can be considered as collective intelligence.

    \item \textbf{\emph{Crowd Intelligence:}}  
    It emerges from the collective intelligent efforts of a large scale of individuals, which leverags diverse sensing capabilities and computational resources of crowds in a blend of collaboration and competition to address complex tasks. Early manifestations of crowd intelligence were mostly seen through internet-based crowd intelligence, and a lot of participants are organized on online platforms to deal with complex problems, such as Wikipedia, Web Q\&A. With the rapid development of IoT technology and embedded devices, a large number of ordinary users ultize IoT terminal devices as basic sensing and computing units. Crowd intelligence further integrates the complement capability of machine and human to complete large-scale sensing and computing tasks, referred to as crowd sensing and computing.

    \item \textbf{\emph{Crowd Knowledge Transfer:} }
    It is one of the fundamental and crucial technology to promote and achieve crowd intelligence. Transferring knowledge is a core ablity of human to continuously learn and adapt in complex and dynamic environments. For individual agents, limitations in data, resources, and other factors often prevent them from achieving optimal performance in new tasks or scenarios. Therefore, CrowdTransfer facilitates the transfer of prior knowledge from some individuals to others, and enhance the self-learning, self-adaptive, and continuous-evolving ability of individual agents for unlocking the full potential of crowd intelligence.

\end{itemize}

\begin{figure*}[!t]
  \centering
  \includegraphics[width=0.63\linewidth]{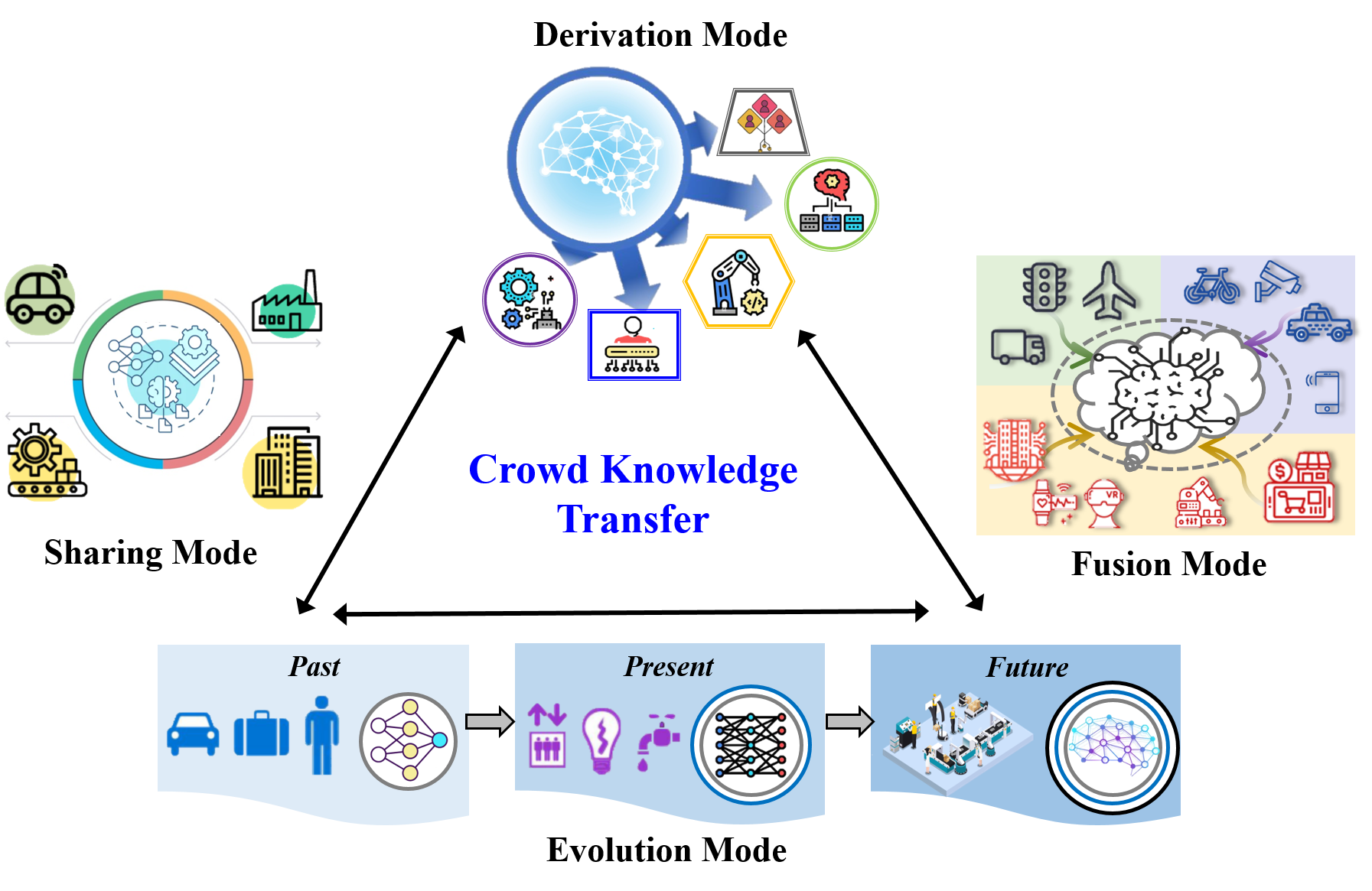}
  \caption{Four CrowdTransfer modes. }
  \vspace{-3mm}
  \label{fig:2_CrowdTransfer_Mode}
\end{figure*}

Inspired by the biological communities, where crowd intelligence emerges through collaboration among individual behaviors, we elaborate on how the knowledge is transferred at different stages throughout the life cycle of individual and their groups to enhance their capabilities.
From the perspective of biological evolution, knowledge transfer exhibits patterns similar to species evolution. 
We posit that, over an extended period of development, a crowd system learns and accumulates a wealth of diverse knowledge. 
Building upon this foundation, the crowd can then be divided into different individual agents or smaller groups, with knowledge subsequently \textit{deriving from the crowd system} and transferring to each individual agent. 
For newly emerging agents, rapid adaptation to the environment can also occur through the \textit{sharing of knowledge from other agents}. 
As time progresses, the environment in which agents find themselves may undergo changes, necessitating the \textit{evolution of their knowledge} to adapt to the new environment. 
Ultimately, each individual possesses unique knowledge, which can \textit{be fused} to foster the development of the entire crowd system. 
The process of transferring knowledge not only aids in the further development of agents but also elevates the knowledge level of the entire crowd system. 
The crowd knowledge transfer pattern underscores profound similarities between biological evolution, emphasizing the significance of transferring knowledge in the crowd system.

In summary, the methodology of CrowdTransfer can be summarized into the following four modes, which are illustrated in Fig. \ref{fig:2_CrowdTransfer_Mode}. 

\begin{itemize}

    \item \textbf{\emph{Derivation Mode:}}  
    The individual agent derives their own knowledge from the accumulated knowledge of the crowd. 
    This pattern emphasizes individual learning and innovation, similar to genetic mutations and adaptive changes in evolution.

    \item \textbf{\emph{Sharing Mode:}}  
    Agents transfer shared knowledge and experiences to enhance the overall knowledge level of the crowd. 
    This is similar to knowledge transfer in social animal groups, where members share information to improve the survival and reproductive prospects of the entire group.

    \item \textbf{\emph{Evolution Mode:} } 
    The knowledge of agents evolves in response to environmental challenges to better adapt to new environments. 
    This mode is akin to the evolution of species, where the characteristics and behaviors of the entire species gradually adapt to a changing environment.

    \item \textbf{\emph{Fusion Mode:} } 
    The knowledge of different agents can fused to facilitate a more robust and diverse knowledge repository. 
    This resembles cooperative interactions among diverse agents, allowing for the integration of different knowledge sources to better address complex problems and challenges.

\end{itemize}

\begin{figure*}[!t]
  \centering
  \includegraphics[width=0.78\linewidth]{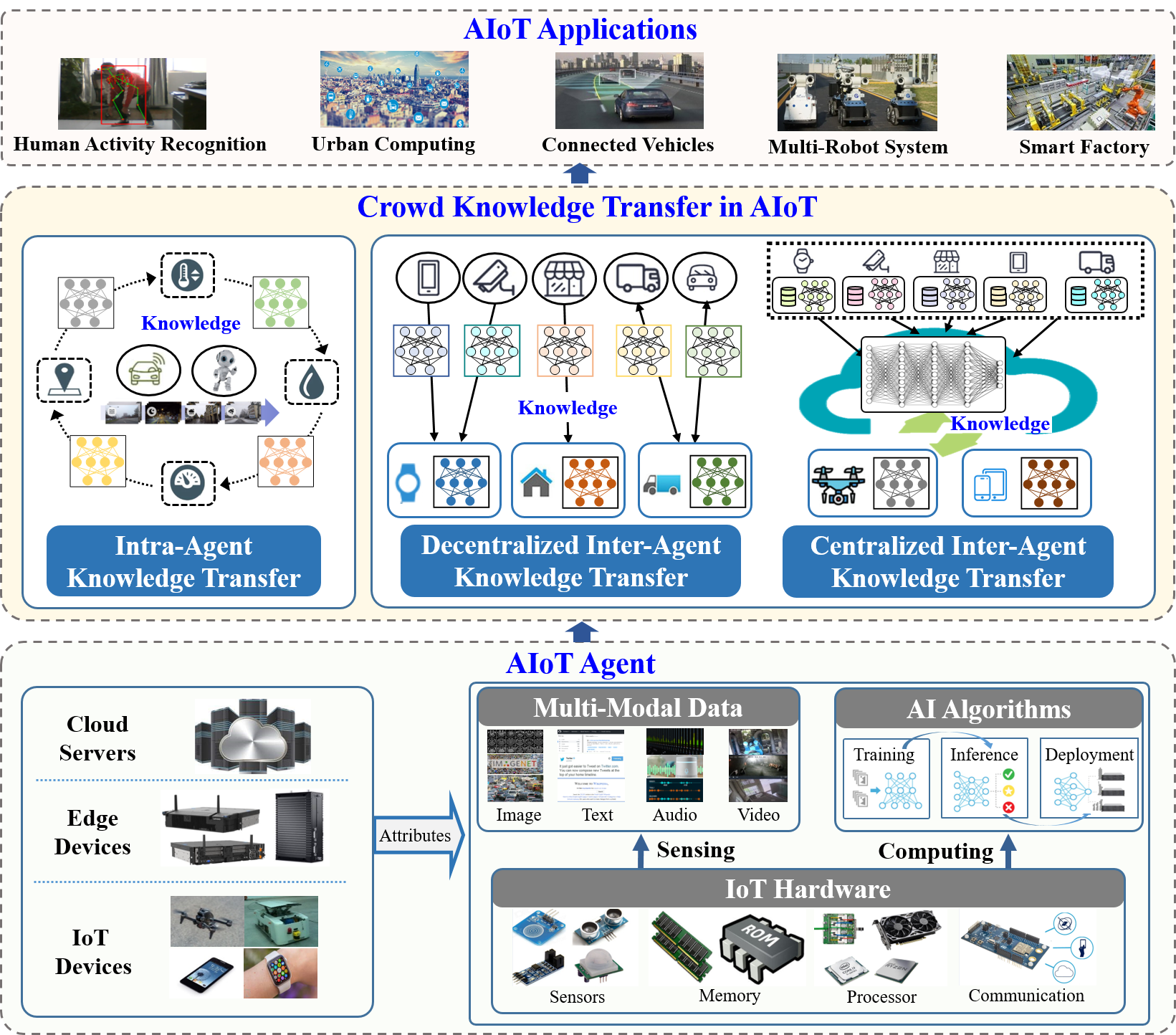}
  \caption{The general framework of CrowdTransfer for AIoT. }
  \vspace{-3mm}
  \label{fig:Crowd_Transfer}
\end{figure*}

\subsection{Framework of CrowdTransfer for AIoT}

We present a general framework of CrowdTransfer for AIoT, as shown in Fig. \ref{fig:Crowd_Transfer}.

\subsubsection{AIoT Agent}

AIoT agent refers to physical objects with abilities of sensing, computing, communicating, decision-making, etc.
Especially, AIoT agent is capable of executing AI algorithms based on the collected data by utilizing IoT hardware, such as IoT devices, edge devices, and cloud servers.

\begin{itemize}

    \item \textbf {\emph{IoT Hadware.}}
   The agent comprises sensor, memory, processor, and communication units, endowing it with profound sensing, computing, and communication proficiencies.  
    
    \item \textbf {\emph{Multi-source Data.}}
   The agent is capable of sensing and acquiring diverse modalities of data from various sources, including text, images, audio, and video.
    
    \item \textbf {\emph{AI algorithms.} }
   The agent executes various types of AI models based on different task requirements, such as deep learning algorithms, reinforcement learning algorithms, and transfer learning algorithms.

\end{itemize}

\subsubsection{CrowdTransfer in AIoT}

In AIoT scenarios, due to various complex factors such as limited resources, heterogeneous data, and incremental tasks, a single isolated agent may not be able to effectively complete a specific task. 
Typically, there are multiple agents concurrently executing diverse yet related tasks, such as tasks with different data distributions.
 Considering the relationship among these tasks performed by multiple agents, it becomes crucial to enhance the task-specific learning of individual target agents by transferring knowledge from other agents. 
The relationship between the different settings of CrowdTransfer and the related areas are summarized in Table \ref{tab:Crowd_Transfer}.

\begin{itemize}

    \item \textbf {\emph{Intra-agent knowledge transfer.}}
   The tasks to be performed by the agent generally change continuously over time, and the agent should optimize and adapt their own models according to the state of IoT devices and environmental information.  
   Therefore, intra-agent knowledge transfer aims to improve the model performance of the target agent corresponding to different scenarios by transferring its previous knowledge. 
   For instance, knowledge transfer between different types of data, known as multi-modal learning, as well as the continuous transfer of knowledge between distinct tasks.

    \item \textbf {\emph{Decentralized inter-agent knowledge transfer.}}
  When a single AIoT agent is unable to achieve better performance due to data and resource limitations, it can enhance its performance by transferring knowledge from other AIoT agents. 
  Specifically, AIoT agents can directly communicate with each other, and they leverage the shared knowledge of one or more other agents without the need for other central computing nodes.  
  For example, multiple agents collaborate with others in a shared environment to accomplish a complex task, like UAV swarm rescue.

    \item \textbf {\emph{Centralized inter-agent knowledge transfer.} }
    As the scale of AIoT agents and the volume of data continues to increase, a central cloud can be used to learn and store a vast amount of knowledge. 
    Unlike direct communication between AIoT agents, each agent interacts with the cloud and leverages the generalized knowledge from the cloud to enhance their own performance. 
    For example, the large-scale deep learning model is pre-trained in the cloud, which is then fine-tuned at the edge/terminal devices for their local tasks.

\end{itemize}

\subsubsection{AIoT Applications}

CrowdTransfer techniques could be applied in many applications to improve model performance.

\begin{itemize}

    \item \textbf {\emph{Human Activity Recognition.}}
   It aims to detect human behaviors in a real-world setting based on the data acquired by some IoT devices. 
   However, it is very time-consuming and laborious to obtain the labeled data to train the recognition models, for example, a large number of video streams should be watched to obtain the action information. 
   Knowledge transfer is one effective way to reduce the need and effort to collect the labeled data.  
    
    \item \textbf {\emph{Urban Computing.}}
    It focuses on tackling complex and practical issues widely existing in cities by using and analyzing the data that has been generated in cities, such as traffic congestion, energy consumption, etc. 
    Generally, the data distribution varies by city, resulting in different performance of the model in different cities. However, training a model from scratch for each city is expensive or impossible due to data scarcity problems. 
    It would be beneficial to improve the model performance and reduce the need to rebuild the model by knowledge transfer.

    \item \textbf {\emph{Connected Vehicles.} }
   The connected vehicle is capable of connecting over wireless networks to nearby devices to share important safety and mobility information. 
   The autonomous car is a special type of vehicle, which is capable of sensing the environment and operating without human involvement. 
   However, road conditions in real environments are very complex, and pre-trained models of autonomous cars may not be able to adapt to changing scenarios. 
   It is very effective to improve the generalization of the model in the face of various environments through knowledge transfer methods.

    \item \textbf {\emph{Multi-Robot System.}}
      It aims to cooperate with multiple intelligent robots (e.g., mobile robots, UAV swarm) to accomplish complex tasks that would otherwise be impossible for one single powerful robot to perform. 
      For each robot, its observations are insufficient to learn effective actions in complex environments. 
      It is necessary to allow a robot to learn from other robots to improve its own behavior, which could reduce the time of training.


    \item \textbf {\emph{Smart Factory.} }
      The production environment in the smart factory that relies on smart manufacturing is automatized and intelligent without human intervention. The data in the production environment generally suffers from the imbalance problem, and the model fails to learn valuable knowledge from these imbalanced samples. 
      To improve the performance of the model, we could transfer useful knowledge from other similar products to reduce labor costs and improve operational efficiency.

\end{itemize}

\begin{table*}[htbp]
  \centering
  \caption{Different Categories of CrowdTransfer in AIoT}
   \label{tab:Crowd_Transfer}
  \resizebox{1.0\textwidth}{!}
  {
    \begin{tabular}{m{0.08\textwidth} | m{0.25\textwidth} | m{0.32\textwidth} | m{0.145\textwidth}}
        \hline
        \multicolumn{1}{c|}{\textbf{CrowdTransfer Category}} & \multicolumn{1}{c|}{\textbf{Characteristic}} & \multicolumn{1}{c|}{\textbf{Common Scenario}} & \multicolumn{1}{c}{\textbf{Related Area}} \\
        \hline
        \multicolumn{1}{c|}{\multirow{4}[8]{*}{\textbf{ \makecell{ \\ Intra-agent \\  knowledge transfer}}}} & \multicolumn{1}{c|}{\multirow{4}[8]{*}{ \makecell{ \\ The individual AIoT agent \\ autonomously performs a variety of \\ tasks. It leverages its prior knowledge \\ to adapt to different scenarios. }}} & The edge/terminal device trains the model across different modalities.  & Multimodal Learning \\
    \cline{3-4}          &       & The edge/terminal device learns new tasks without forgetting knowledge from previous tasks. & Continual Learning \\
    \cline{3-4}          &       & The edge/terminal device learns from simple tasks to complex tasks gradually.  & Curriculum Learning \\
    \cline{3-4}          &       & The edge/terminal device adapts to the unlabeled target scenario without accessing source data.  & Test-Time Adpatation \\
        \hline
        \multicolumn{1}{c|}{\multirow{3}[6]{*}{\textbf{\makecell{\\  \\ \\ Decentralized inter-agent \\ knowledge transfer}}}} & \multicolumn{1}{c|}{\multirow{3}[6]{*}{\makecell{ \\ Multiple AIoT agents directly \\ communicate with each other. They \\ harness the shared knowledge \\ among agents to enhance \\ their performance. }}} & Multiple edge/terminal devices collaborate with other devices to accomplish a task, and each device operates within its own observations throughout the training and execution processes. & Distributed MARL \\
    \cline{3-4}          &       & The edge/terminal device transmits the model to other devices locally instead of global model aggregation by server coordination.  & Decentralized Federated Learning \\
    \cline{3-4}          &       & The target agent mimics and learns behaviors from other expert agents to acquire new skills. & Imitation Learning \\
        \hline
        \multicolumn{1}{c|}{\multirow{4}[6]{*}{\textbf{\makecell{ \\   \\ \\  \\  Centralized inter-agent \\ knowledge transfer}}}} & \multicolumn{1}{c|}{\multirow{3}[6]{*}{\makecell{\\    The central cloud is utilized to learn \\ a wealth of knowledge. AIoT agents \\ leverage the general knowledge \\ from the cloud to improve \\ their performance. }}} & The large-scale model is pre-trained on vast amounts of data in the cloud, while the edge/terminal device adjusts the model based on its own data.  & Fine-tuning of Large Language Models \\
    \cline{3-4}          &       & Multiple edge/terminal devices coordinate to train a global model via the central cloud without sharing the raw data.  & Federated Transfer Learning \\
    \cline{3-4}          &       & The central cloud is used to guide the training process of the model on the edge/terminal device due to limited data and constrained resources. & Teacher-Student Learning \\
    \cline{3-4}          &       & Multiple edge/terminal devices share a centralized training environment, but each device acts independently in its own environment. &  CTDE-based MARL \\
        \hline
    \end{tabular}
  }
\end{table*}%



\section{KEY TECHNIQUES OF KNOWLEDGE TRANSFER} \label{TL_3}

In recent years, transfer learning methods have experienced substantial advancements and have led to the emergence of various learning paradigms. 
This article embraces a comprehensive understanding of transfer learning, encompassing the sharing of diverse forms of knowledge, e.g., network parameters, typical instances, features, or network structures, between source and target domains/tasks within the scope of this paper. In this section, the key techniques for facilitating knowledge transfer comprise \textit{domain adaptation/generalization}, \textit{multi-task learning}, \textit{knowledge distillation}, and \textit{meta learning} (as shown in Table \ref{comparison_tra_methods}).

\subsection{Domain Adaptation}

Domain adaptation (DA) aims to build a model that can learn knowledge from semantically related source domains with different distributions to perform the tasks in the target domain \cite{zhang2022transfer}. In AIoT scenarios, agents frequently encounter tasks with identical conditional probability distributions (CPDs) but differing marginal probability distributions (MPDs). For instance, a target model deployed in an intelligent surveillance camera exhibits proficient performance under sunny weather conditions, yet experiences a sudden decline in performance during snowy weather conditions. This issue can be resolved through DA, wherein the model is adapted to transition from sunny to snowy weather conditions. The entire process of adaptation necessitates aligning the distribution of data from the source domain with that of the target domain, encompassing the alignment of CPDs, MPDs, or both. 

Given the feature space of the source domain, $X_{S}$, the feature space of the target domain, $X_{T}$, the label space of the source domain, $Y_{S}$, the label space of the target domain, $Y_{T}$, as well as the training sets of the source domain, $\mathcal{D}_{S}=\left\{ \left ( x_{i},y_{i} \right )\right\}$, and the target domain, $\mathcal{D}_{T}=\left\{ x_{j}\right\}$, the objective of DA is to learn a mapping function $f: X_{S} \rightarrow Y_{S}$. This mapping function aims to minimize the distribution discrepancy between the domains, $D_{KL}\left ( p_{S}\parallel p_{T} \right )$, and a combination with the loss function $L\left ( f\left ( x_{i} \right ), y_{i} \right )$:

\begin{equation}
    \label{eq_DA}
    \min_{f}\left ( \lambda D_{KL} \left ( p_{S}\parallel p_{T} \right ) + \sum _{\left ( x_{i},y_{i} \right )\in\mathbb{D}_{s}} L\left ( f\left ( x_{i} \right ), y_{i} \right ) \right ),
\end{equation}
where $\lambda$ serves as a weighting parameter. At present, dominant methods employed to facilitate DA primarily comprise instance-based adaptation, feature-based adaptation, and model-based adaptation.

\begin{table*}
\caption{Statistics of the transfer learning methods}
\label{comparison_tra_methods}
\centering
\renewcommand\arraystretch{1.1}{
\resizebox{\textwidth}{!}{
\begin{tabular}{ccccc}
\toprule
\multicolumn{1}{c}{\textbf{Concept}} & \multicolumn{1}{c}{\textbf{Sub-type}} & \multicolumn{1}{c}{\textbf{Description\&Objective}} & \multicolumn{1}{c}{\textbf{Key Focus}} & \multicolumn{1}{c}{\textbf{Related Works}}  \\ \midrule

\multirow{3}{*}{\begin{tabular}[c]{@{}c@{}}Domain\\ Adaptation\end{tabular}}      
& Instance-based                                                            & \multirow{3}{*}{\begin{tabular}[c]{@{}c@{}}Map data from various source and \\ target domains into a shared feature space, \\ minimizing the distances between them\end{tabular}}               & \multirow{3}{*}{\begin{tabular}[c]{@{}c@{}}Addressing distributions differences \\ between source and target domains\end{tabular}}                 & \multirow{3}{*}{\begin{tabular}[c]{@{}c@{}}TJM\cite{long2014transfer}, SGF\cite{gopalan2011domain}, GFK\cite{gong2012geodesic}, \\ DAN\cite{long2015learning}, DANN\cite{ganin2015unsupervised}\end{tabular}}                              \\ 
& Feature-based                                            &                                                          &                                                          &                                                                                                                   \\ 
& Model-based                                                              &                                                                                                                                                                                                 &                                                                                                                                                    &                                                                                                                   \\ \midrule
\multirow{3}{*}{\begin{tabular}[c]{@{}c@{}}Domain \\ Generalization\end{tabular}} 
& Data manipulation       & \multirow{3}{*}{\begin{tabular}[c]{@{}c@{}}Train the model using data from the \\ source domain for generalization to different \\ target domains with diverse data distributions\end{tabular}} & \multirow{3}{*}{\begin{tabular}[c]{@{}c@{}}Emphasizing shared features, enabling \\ broad cross-domain generalization\end{tabular}}                & \multirow{3}{*}{\begin{tabular}[c]{@{}c@{}}DARLING\cite{zhang2022towards}, L2A-OT\cite{zhou2020learning}, \\ DAML\cite{shu2021open}\end{tabular}}                                 \\ 
& Representation learning &                                                             &                                                                                                                                                    &                                                                                                                   \\ 
                                                                                  & Learning strategies     &                                                                                                                                                                                                 &                                                                                                                                                    &                                                                                                                   \\ \midrule
\multirow{2}{*}{\begin{tabular}[c]{@{}c@{}}Multi-task\\ Learning\end{tabular}}   & Joint task   & \multirow{2}{*}{\begin{tabular}[c]{@{}c@{}}Simultaneous training multiple \\ correlated tasks within a single model\end{tabular}}                                & \multirow{2}{*}{\begin{tabular}[c]{@{}c@{}}Navigating task interdependencies, fostering \\ generalized learning across diverse tasks\end{tabular}} & \multirow{2}{*}{\begin{tabular}[c]{@{}c@{}}DRN\cite{long2017learning}, SCL\cite{yu2016learning}, \\ Sluice Network\cite{ruder2017sluice}, AMT\cite{shinohara2016adversarial}\end{tabular}}   \\ 
                                                                                  & Auxiliary task &                                                                                                                                                                                                 &                                                                                                                                                    &                                                                                                                   \\ \midrule
\multirow{2}{*}{\begin{tabular}[c]{@{}c@{}}Knowledge\\ Distillation\end{tabular}} & Offline learning      & \multirow{2}{*}{\begin{tabular}[c]{@{}c@{}}Improve the student's performance by having a \\light student model imitate complex teacher models\end{tabular}}   & \multirow{2}{*}{\begin{tabular}[c]{@{}c@{}}Channeling knowledge from teachers to \\ students, facilitating model compression\end{tabular}}         & \multirow{2}{*}{\begin{tabular}[c]{@{}c@{}}Fitnets\cite{romero2014fitnets}, \\ BAN\cite{furlanello2018born}, DML\cite{zhang2018deep}\end{tabular}}       \\ 
                                                                                  & Online learning        &                                                                                                                                                                                                 &                                                                                                                                                    &                                                                                                                   \\ \midrule
\multirow{3}{*}{\begin{tabular}[c]{@{}c@{}}Meta \\ Learning\end{tabular}}         & Optimization-based                                                       & \multirow{3}{*}{\begin{tabular}[c]{@{}c@{}}Equipping models to swiftly adapt to \\ fresh tasks, acquiring strategies for \\ rapid adaptation or parameter setup\end{tabular}}                   & \multirow{3}{*}{\begin{tabular}[c]{@{}c@{}}Facilitating agile learning, enabling \\ efficient adjustment to novel tasks\end{tabular}}              & \multirow{3}{*}{\begin{tabular}[c]{@{}c@{}}MAML\cite{finn2017model}, Reptile\cite{nichol2018first}, \\ MANN\cite{santoro2016meta}, MetaNet\cite{munkhdalai2017meta}, \\ Prototypical Network\cite{snell2017prototypical}\end{tabular}} \\ 
                                                                                  & Model-based                                                              &                                                                                                                                                                                                 &                                                                                                                                                    &                                                                                                                   \\ 
                                                                                  & Metric-based                                                             &                                                                                                                                                                                                 &                                                                                                                                                    &                                                                                                \\ \bottomrule
\end{tabular}}
}
\end{table*}

\textit{1) Instance-based DA}: The instance-based DA aims to reduce the discrepancy between the source and target domains by re-weighting the source instances, which can be achieved through direct instance weighting or instance kernel mapping weighting. Direct instance weighting adjusts the weights of the source instances to align the weighted source data distribution with the target data distribution \cite{jiang2007instance}. For example, Chen et al. \cite{chen2016visual} assign a weight vector to the source instances and subsequently re-weight those instances located in the vicinity of the target domain subspace for aligning the source and target domain sub-spaces. The aforementioned methods are implemented within the original data space, while researchers also propose methods of instance kernel mapping weighting for achieving DA by a non-parametric approach \cite{zhong2009cross}. For example, Long et al. \cite{long2014transfer} propose the transfer joint matching (TJM) for unsupervised DA, which sparsely samples the utilized transformation matrix from the source domain to the target domain. The corresponding coefficient values in this matrix increase in proportion to the strength of the correlation between source and target instances. 

\textit{2) Feature-based DA}: Feature-based DA aims to extract general feature representations from multiple sources using linear or non-linear mapping methods. Researchers achieve knowledge transfer from source domains to target domains by constructing low-dimensional feature subspaces, e.g., manifold spaces \cite{zhang2011adaptive}. For example, Gopalan et al. \cite{gopalan2011domain} propose an unsupervised low-dimensional subspace knowledge transfer method called sampling geodesic flow (SGF), which samples a limited number of sub-spaces along geodesic lines between the source and target domain data to find feature representations with minimal between-domain discrepancies. Gong et al. \cite{gong2012geodesic} propose the geodesic flow kernel (GFK) method, which models domain discrepancy by sampling infinitely many sub-spaces. Fernando et al. \cite{fernando2013unsupervised} directly use an alignment matrix to bring the source and target domain sub-spaces closer to the data points in Grassmann manifold space. In addition, several works utilize distribution metrics between source and target domains to learn transformations or projections of the data, aiming to reduce differences in feature distributions across domains. For example, Zhang et al. \cite{zhang2017joint} incorporate joint discriminative subspace learning and maximum mean discrepancy (MMD) minimizing, and propose the joint geometrical and statistical alignment model to minimize the difference in conditional distributions between the projected source and target domain data.

\textit{3) Model-based DA}: Model-based DA aims to enhance generalization in the target domain by adjusting the parameters or structure of models. Yosinski et al. \cite{yosinski2014transferable} initially investigate the transferability of knowledge across different layers (i.e., bottom, middle, and top) of deep neural networks (DNNs), observing a reduction in transferability as the distribution discrepancies between domains increase. Subsequently, Long et al. \cite{long2015learning} propose the deep adaptation network (DAN) for model-based DA, which leverages non-parametric kernel matching techniques, e.g., MMD, to align the source and target domains by incorporating high-level features into reproducing kernel Hilbert space (RKHS). Ganni et al. \cite{ganin2015unsupervised} propose the DANN for unsupervised DA, consisting of a feature extractor, task classifier, and a domain classifier. The principle behind DANN is that if a shared feature space is learned between the source and target domains, a discriminative model trained on the source domain can also effectively capture target domain features in this shared space,  referred to as domain-transferable features. Zhang et al. \cite{zhang2018collaborative} propose the CAN, comprising of domain-distinguishable information extraction, domain-invariant information extraction, and cooperative confrontation training combining both types of information. It does not align samples by category but utilizes the output vector of each feature extraction block for domain classification.

\subsection{Domain Generalization}
Domain Generalization (DG) aims to learn a model with strong generalization ability from several domains with different data distributions and achieve better results on the unknown test set \cite{shu2021open}. DG methods, in contrast to DA approaches, only necessitate access to the training set during the model training phase. Unlike DA, which demands both source and target domain data from the training and testing sets, DG solely relies on source domain data. Moreover, DG possesses the capacity to handle an unlimited number of potential target domains in the future.

Given multiple feature spaces $X_{S}$ of source domains, multiple label spaces $Y_{S}$ of source domains, and a feature space $X_{T}$ of a target domain, the objective of DG is to learn universal mapping function $f: X_{S} \rightarrow Y_{S}$, enabling accurate predictions on unseen target domains. It is typically to combine the prediction loss and the domain discrepancy:
\begin{equation}
    \label{eq_DG}
    \min_{\theta}\sum_{i=1}^{n}L \left ( D_{S}^{i},\theta \right ) +\lambda D_{KL}\left ( p_{S}^{i}\parallel p_{T} \right ),
\end{equation}
where $L\left ( D_{S}^{i},\theta \right )$ represents the prediction loss of the source domain $D_{S}^{i}$, $\theta$ denotes the model parameters, $D_{KL}\left ( p_{S}^{i}\parallel p_{T} \right)$ measures the distribution discrepancy between the source domains and target domain, and $\lambda$ is a balancing parameter.

Current DG methods can be primarily classified into the following three major categories: (1) Data manipulation-based DG: It involves enhancing the training data through the use of augmentation and variations. By manipulating the data, the training set can be enriched, which results in improved generalization performance; (2) Representation learning-based DG: It focuses on learning domain-transferable features, also known as domain-transferable representation learning. By learning representations that are resilient to domain discrepancies, deep models become more effective in addressing variations across different domains. (3) Learning strategies-based DG: It refers to the integration of other machine learning patterns into multi-domain training, e.g., meta-learning. 

For instance, Zhang et al. \cite{zhang2022towards} introduce a method called domain-aware representation learning (DARLING). The model undergoes pre-training on unlabeled data from various source domains, followed by training on labeled data from the source domain. Chen et al. \cite{chen2021style} leverage intra-domain style invariance to enhance the learning of inter-domain semantic invariance, thereby improving its generalizability. During training, they incorporate intra-domain style invariance at the instance level to enable the network to capture variations in semantic features. Furthermore, they utilize a memory-based semantic feature bank mechanism, where the final class of a pair of instances’ features is determined based on the direction of instances previously stored in the memory bank. Li et al. \cite{li2019feature} suggest two approaches for DG: The first method involves training individual models for each source domain. When a testing domain is encountered, the most relevant model for that domain is estimated, and its classifier is employed for prediction. The second approach is founded on the assumption that every domain comprises a globally shared factor and a domain-specific component. During training on source domains, the domain-specific and domain-agnostic components are disentangled. By extracting and transferring the domain-agnostic component as a standalone model, it is expected to exhibit good performance on new source domains. Wang et al. \cite{wang2022contrastive} consider the causal invariance of the average causal effect (ACE) \cite{angrist1995two} of features on labels. ACE is defined as the difference in expected output between the intervention of a specific input feature and the baseline output obtained when the same feature is consistently perturbed within a fixed value range. Zhou et al. \cite{zhou2020learning} propose the learning to augment by optimal transport model (L2A-OT) to address DG by enhancing the diversity of available source domains. The main idea is to learn a conditional generative network that maps source domain images to a pseudo-new domain, and then combine the source domain images with the pseudo-new domain images to train the target task model.

\subsection{Multi-task Learning}
The underlying concept of multi-task learning lies in the possibility of knowledge or feature sharing among related tasks. Through the sharing of these task-transferable features, models can effectively transfer information across tasks, enhancing the learning process. Particularly in situations with limited training data, multi-task learning can aid models in acquiring additional information from analogous tasks. Given multiple tasks $\mathcal{T}_a= \left\{{T_{1}, T_{2},\cdots, T_{n}}\right\} $, each task including a feature space $X$ and a label space $Y$, along with their corresponding training datasets $D_{i}=\left\{ \left ( x_{i}, y_{i} \right )\right\}$, the objective of multi-task learning is to learn a set of models $f_{1},f_{2},\cdots,f_{n}$, where each model corresponds to a task $T_{i}$, with the aim of minimizing the loss function across tasks:
\begin{equation}
    \label{eq_Multi-task}
    \min_{f_{1},f_{2},\cdots,f_{n}}\sum_{i=1}^{n}\sum _{\left ( x_{j},y_{j} \right )\in D_{i}} L\left ( f_{i}\left ( x_{j} \right ), y_{j} \right ),
\end{equation}

The design of multi-task learning methods primarily encompasses two approaches: One is learning a shared feature representation for multiple tasks based on shallow or deep models, which can be a subset or transformation of the original feature representation; the other is reducing distribution differences between tasks by task clustering or analyzing of task correlations. In conclusion, this paper divides multi-task learning methods into two types:

\textit{1) Joint task learning-based}: The goal of joint task learning is to simultaneously train multiple tasks with similar data distribution, and enhance specific task performance through knowledge sharing. Parameter sharing is a straightforward method to achieve knowledge sharing, primarily involving the sharing of hidden layer parameters in DNNs, mainly categorized into hard parameter sharing and soft parameter sharing.  The fundamental concept behind hard parameter sharing is the sharing of certain hidden layers among different tasks. Conversely, soft parameter sharing involves each task having its own model and parameters, with the regularization of model parameter distances to promote similarity, e.g., through the utilization of L2 norm \cite{duong2015low} and trace norm \cite{yang2016trace}. Moreover, Long et al. \cite{long2017learning} propose a deep relationship network (DRN) to strengthen the relationship between tasks, and leverage the relationships for effective feature transfer in deep networks. Misra et al. \cite{misra2016cross} introduce the cross-stitch network. It incorporates a cross-stitch unit between the feature layers of the two networks, allowing them to automatically learn the relevant shared features. Specifically, each task learns a linear mapping as a shared representation at each layer, followed by a nonlinear transformation in the subsequent layer. Furthermore, Ruder et al. \cite{ruder2017sluice} add switch units between neural network layers, and propose the sluice network for the joint task learning.

\textit{2) Auxiliary task learning-based}: In an environment with multiple agents, each agent is responsible for similar tasks, and employing multi-task joint learning allows for leveraging shared knowledge across tasks to reinforce each agent’s capabilities. Girshick et al. \cite{girshick2015fast} use CNN for traditional object detection tasks, enabling simultaneous prediction of object categories and locations in images. Yu et al. \cite{yu2016learning} propose the structural correspondence learning (SCL) method for cross-domain sentiment classification. The SCL incorporates two auxiliary binary prediction tasks: identifying whether a sentence contains positive or negative emotion words, and inferring whether the auxiliary task labels can be inferred from unlabeled data in the source and target domains. In addition, Shinohara et al. \cite{shinohara2016adversarial} put forward the anti-multi-task learning framework, abbreviated as AMT, which leverages anti-task information from auxiliary tasks to mitigate the noise in the main tasks and learn representations that closely resemble the real underlying data. The AMT consists of three sub-networks: the main task output sub-network, the secondary task output sub-network and the input network shared by primary and secondary tasks. It aims to learn an adversarial representation of the secondary task and eliminate irrelevant domain-dependent information that may hinder the primary task’s feature representation. To facilitate intelligent agents in determining task similarity, Rei \cite{rei2017semi} proposes that if the data from two tasks are generated by applying a fixed probability distribution through the same class of function transformations, then the two tasks are functionally correlated.

\subsection{Knowledge Distillation}
Knowledge distillation (KD) \cite{hinton2015distilling} aims to facilitate knowledge transfer by allowing a small/simple model to closely approximate or even outperform complex/large models, thereby achieving comparable predictive results with reduced complexity. Given the teacher model $W$ and student model $w$, along with a set of training data $D$, the objective of KD is to learn the student model by minimizing the student loss on the training data, denoted as $L_{w}$, while utilizing the output of teacher model as an additional signal of supervision:
\begin{equation}
    \label{eq_KD}
    \min_{w}\left ( L_{w}\left ( w,W \right ) + \lambda L_{K}\left ( T,w,W \right ) \right ),
\end{equation}
where $L_{K}$ refers to the loss of knowledge transfer, and $\lambda$ represents a balancing scalar. This paper categorizes KD methods into the following two types:

\textit{1) Offline learning-based KD}: Offline learning-based KD approaches involve distilling knowledge from a pre-trained large teacher model to train a smaller student model. The primary objective is to train the student model to achieve comparable performance to the teacher model under the guidance of the teacher model. The supervision signal from the teacher usually refers to the teacher’s knowledge, including logits and intermediate feature representations, which aids the student model in emulating the teacher. For example, Hinton et al. \cite{hinton2015distilling} introduce the concept of softmax temperature to soften the predicted labels, which act as knowledge transferred from the teacher model to guide the training of the student model. The main components of the KD are the distillation loss and the student loss. The distillation loss consists of a soft target loss and a temperature parameter, while the student loss is calculated based on the cross-entropy between the predicted class probabilities of the student network and the true labels. In addition to utilizing logits as distilled knowledge, researchers \cite{romero2014fitnets} also utilize the outputs of intermediate layers in the network, such as feature maps obtained from convolutions, to supervise the training of the student model. For instance, Heo et al. \cite{heo2019knowledge} introduce knowledge transfer by leveraging the activation boundaries of hidden layer neurons, with an activation boundary being a distinct hyperplane that determines the activation or deactivation of a neuron. This method demonstrates that the activation transfer loss is minimized when the boundaries generated by the student model align with those generated by the teacher model.

For crowd agents, leveraging the knowledge of multiple teachers can benefit their learning process. One direct approach for knowledge transfer from multiple teacher networks is to utilize the average logits of all teachers as the supervision signal, and the other is to integrate each teacher’s feature vector. To leverage both logits and intermediate features, Chen et al. \cite{chen2019two} use two teacher networks, with one teacher transferring knowledge based on logits and the other teacher transferring knowledge based on features. Wu et al. \cite{wu2019distilled} also propose using a learnable transformation matrix in the student network to solve cross-domain knowledge transfer. Radosavovic et al. \cite{radosavovic2018data} propose a distillation method that applies multiple transformations to unlabeled data to construct different teacher models while sharing the same network structure. 

\textit{2) Online learning-based KD}: Online learning-based KD approaches involve simultaneously training a group of student models to learn from each other. For example, Zhang et al. \cite{zhang2018deep} propose a method in which a group of untrained student networks, with identical structures, learn the target task together through alternating iterations. Specifically, each network has two loss functions during the learning process, i.e.,  the conventional supervised loss and the interaction loss between the networks. Chung et al. \cite{chung2020feature} not only transfer knowledge regarding class probabilities but also utilize an adversarial learning framework to transfer knowledge about feature maps. Multiple networks are trained simultaneously, and a discriminator is employed to differentiate the feature map distributions of different networks. 

Some researchers propose utilizing multiple student models to aggregate intermediate predictions, creating a dynamic "teacher" or "leader" that guides all student networks in a closed-loop manner to enhance student learning. For instance, Chen et al. \cite{chen2020online} introduce an online KD approach that employs a two-level distillation process involving multiple auxiliary student networks and a leader. In the first level of distillation, each student network possesses a unique set of aggregation attention weights to derive its own targets from the predictions of other auxiliary student networks. In the second level of distillation, the integrated knowledge from the auxiliary student networks is further transferred to the "leader" model, which serves as the inference model.

\subsection{Meta-Learning}
Meta-learning \cite{hospedales2021meta} aims at leveraging prior knowledge acquired from multiple tasks to guide its learning in new tasks, i.e., learning to learn. Meta-learning consists of the base learning stage and the meta-learning stage. During the base learning stage, internal algorithms are utilized to address a learning task with provided data and optimization objectives. In the meta-learning stage, external algorithms are utilized to update internal learning algorithms, enabling them to improve external optimization objectives, such as generalizability and learning efficiency. Given a meta-training dataset $D_{meta}$, where each meta-task consists of a training set $D_{train}$ and a testing set $D_{test}$, the objective is to learn a model $f$ that achieves high performance on a new task $T_{new}$ with only a small amount of instances $D_{train,new}$:
\begin{equation}
    \label{eq_meta}
    \max_{f}\sum_{T_{new}\in D_{meta}}\sum _{\left ( x_{j},y_{j} \right )\in D_{train,new}} L_{f}\left ( f, D_{test,new} \right ),
\end{equation}
where $L_{f}$ is the loss function during meta-training.

\textit{1) Optimization-based meta-learning}: This type of methods primarily treats the internal tasks as an optimization problem, emphasizing the extraction of meta-knowledge (i.e., optimization parameters) that enhance the model in the target agent. For instance, Finn et al. \cite{finn2017model} propose a model-agnostic meta-learning approach known as MAML, whose main idea is to search for a set of highly generalized initial parameters that allow the model to efficiently adapt itself through gradient updates based on limited data. Reptile \cite{nichol2018first} directly updates the parameters of the meta-learning network by multiplying a learnable parameter with the difference between the meta-learning network parameters and the base learning network parameters. The base learning network parameters are updated using first-order gradients computed after sampling multiple tasks, thus reducing the computational and storage cost of knowledge fusion. 

\textit{2) Model-based meta-learning}: This class of methods encapsulates the learning of internal tasks within a single-model feed-forward process, often using black-box models. It enables rapid model updates based on specific network structures to adapt to new tasks efficiently. For example, Santoro et al. \cite{santoro2016meta} propose a meta-learning knowledge fusion method called memory-augmented neural network (MANN), which utilizes an external storage space to explicitly retain feature information from historical task data. MANN constructs the storage space by a neural Turing machine (NTM), which rapidly encodes and retrieves information. The meta-learning algorithm is then used to optimize the reading and writing processes of the NTM, ultimately enabling its application to few-shot tasks. Similarly, MetaNet \cite{munkhdalai2017meta} adds an additional memory module to the meta-learner, whose training process involves the acquisition of meta information, the generation of fast weights (using another neural network to predict network parameters), and the optimization of slow weights (using stochastic gradient descent). 

\textit{3) Metric-based Meta-learning}: This type of methods is to learn a metric space that captures similarities between data from various tasks, enabling internal task models to perform non-parametric learning and predict the labels directly of corresponding training samples. For example, the Siamese Network \cite{koch2015siamese} consists of two twin networks sharing the same weights and parameters, which are trained to learn the relationship function between input sample pairs. The Matching Network \cite{vinyals2016matching} has a similar structure to Siamese Networks, but it incorporates attention and memory mechanisms to recognize unlabeled samples when only limited labeled samples are available. The attention weights between two samples are determined based on the cosine similarity of their feature embeddings. Moreover, the Prototypical Network \cite{snell2017prototypical} aims to learn a metric space for diverse tasks and complete model training by computing the distance between each sample and its prototype representation of each task. The Relation Network \cite{sung2018learning} differs from previous studies by emphasizing learning a transferable embedding representation, and it goes a step further by learning a transferable metric for deep task similarity. The whole network is divided into two stages: an embedding module that extracts feature information of different samples, and a relation module that calculates the similarity scores between samples to determine their class membership.



\begin{table*}[!t]
\scriptsize
\centering
\caption{Summary of Crowd Knowledge Transfer Models.}
\label{tab:CrwodTransfer_models}
\renewcommand\arraystretch{1.1}{
\begin{tabular}{m{1.8cm}<{\centering} | m{2.5cm}<{\centering} | m{2cm}<{\centering} | m{2.8cm}<{\centering} | m{6.9cm}}
\hline
\textbf{CrwodTransfer}       & \textbf{Related Area}           & \textbf{Methodoloy Mode}        & \textbf{AIoT Challenge}                 & \textbf{Application Example}                                                                   \\ \hline
\multirow{4}{*}{\shortstack{ \\ \\  \\  \\ \\   Centralized \\ Inter-agent  \\ Knowledge \\ Transfer}} 
& Federated Transfer Learning   & Fusion Mode   & Communication Bottleneck; Data privacy; Mobility  & Multi-agent real-time collaborative transfer for obstacle avoidance \cite{liang2022federated}; Multi-user collaborative transfer for personalized health services \cite{chen2020fedhealth}                     \\
\cline{2-5} & Teacher-Student Learning      & Derivation Mode & Unlabeled Data; Limited Computing and Storage Resources     & A active teacher method for semi-supervised object detection \cite{mi2022active}; A progressive teacher-student method for early action prediction \cite{wang2019progressive}         \\
\cline{2-5} & Fine-tuning of Large Language Models   & Derivation Mode    & Heterogeneous Data  &  Integrating PEFT and adaption method for medical image segmentation  \cite{wu2023medical}; An adapt pre-trained Image Model for video action recognition \cite{yang2023aim};     
  \\ 
  \cline{2-5} & CTDE-based MARL   &   Fusion Mode    &  Unlabeled Data;  Communication Dynamics   &   A multi-agent reinforcement learning method considering the energy efficiency of AGVs for AGVs scheduling and routing \cite{ye2023toward}; A constrained multi-agent reinforcement
learning for CAVs(connected and autonomous vehicles (CAVs) driving \cite{zhang2023spatial}        
  \\ \hline
\multirow{3}{*}{\shortstack{ \\ \\  \\  \\  \\  Decentralized \\ Inter-agent \\ Knowledge \\ Transfer}}
&  Distributed MARL     & Sharing Mode   & Unlabeled Data;  Mobility   &  A Distributed-Training-Distributed-Execution MARL framework for the electric vehicle charging schedules \cite{zhang2023distributed} ;  A Distributed-Training-Distributed-Execution MARL scheme for power control in heterogeneous networks \cite{xu2023distributed}    \\
\cline{2-5} & Imitation Learning       & Sharing Mode    & Unlabeled Data;  &  A hierarchical interpretable imitation learning for end-to-end autonomous driving \cite{teng2022hierarchical}; A generative adversarial imitation learning method for human pose prediction \cite{wang2019imitation}
   \\
\cline{2-5} & Decentralized Federated Learning         & Sharing Mode
& Communication Bottleneck; Data Privacy; Mobility  &  A decentralized federated learning approach for human-computer interaction in IoT applications \cite{elayan2021deep}; A deep decentralized federated learning approach for IoT-based healthcare systems \cite{chhikara2020federated} 
 \\ \hline
\multirow{3}{*}{\shortstack{ \\ \\  \\  \\  \\  \\ \\   \\  
 \\  \\Intra-agent \\ Knowledge \\ Transfer}}
& Multimodal Learning     & Fusion Mode  & Heterogeneous Data   &  Combining RNA and miRNA Sequencing data, and DNA methylation data for survival predicting in patients \cite{lv2020survival}; Fusing RGB visual stream and skeleton stream for human action predicting \cite{li2021toward};                \\
\cline{2-5} & Curriculum Learning     & Evolution Mode
& Incremental Tasks       &  An end-to-end curriculum learning approach for autonomous driving \cite{anzalone2022end}; A curriculum reinforcement learning method based on the curiosity model for agent competition and cooperative \cite{lin2022learning}     \\
\cline{2-5} & Continual Learning     & Evolution Mode
& Incremental Tasks       &  Dynamic gradient scenario memory for continual automatic driving  \cite{li2023continual}; The dynamic architecture approach and rehearsal approach  for agents continual learning and private unlearning \cite{liu2022continual} \\
\cline{2-5} & Test-time Adaption      & Evolution Mode      & Real-time Data Streams; Incremental Tasks; Unlabeled Data       &  Test-time dynamic adaptation for on-the-fly medical image segmentation \cite{valanarasu2022fly}; Test-time aggregating diverse experts with self-supervision for test-agnostic long-tailed recognition \cite{zhang2022self}
       \\ \hline
\end{tabular}
}
\end{table*}

\section{CROWD TRANSFER LEARNING MODELS}\label{TL_4}

Crowd knowledge transfer plays an increasingly prominent role in AIoT. Benefiting from the self-adaptive ability of crowd transfer learning models, they optimize their own modules according to the state of IoT devices and the changes in the environment to improve specific task performances. This section mainly analyzes the crowd transfer learning models in detail to manifest the mechanisms of knowledge usage manners behind various methods. 

Due to the sensing/perception ability of agents, a single modal or task data is not enough to successfully complete a specific task. In AIoT scenarios, there are multiple agents executing different yet related tasks, such as tasks with varying data distributions or different data modalities. Taking into account the interrelated tasks of multiple agents and leveraging shared factors or representations among these tasks is also an important approach to enhance the generalization of individual task learning and facilitate crowd knowledge transfer. This section primarily focuses on introducing the research paradigm of crowd transfer learning models, mainly including \textit{centralized inter-agent knowledge transfer}, \textit{decentralized inter-agent knowledge transfer}, and \textit{intra-agent knowledge transfer}. 
Table \ref{tab:CrwodTransfer_models} summarizes the CrowdTransfer Models. 

Specifically, the centralized inter-agent knowledge transfer and decentralized inter-agent knowledge transfer summarize crowd knowledge transfer methods from the perspective of interactions among crowd agents. Constrained by the requirements of privacy security, data acquisition ability, etc., it is often challenging for an agent to obtain sufficient data for a specific task, and models cannot be learned well due to limited training data. The target agents can not only achieve self-optimization by using the knowledge of their surrounding agents for reference, but also realize crowd evolution by sharing knowledge with each other. 

The intra-agent knowledge transfer summarizes crowd knowledge transfer methods from the perspective of agent itself, i.e., attributes of agents (such as data scale, resource, and sensing ability). In AIoT scenarios, agents have different functions, various number/types of basic sensors, and chips with distinct capabilities, which makes each agent form different aspects and levels of knowledge in the process of interacting with contexts, e.g., experiences, policies, or skills.

\subsection{Centralized Inter-agent Knowledge Transfer}

\subsubsection{Federated Transfer Learning} \label{chapter:4.1.1}

\begin{figure}
    \centering
    \includegraphics[width=0.49\textwidth,height=5.5cm]{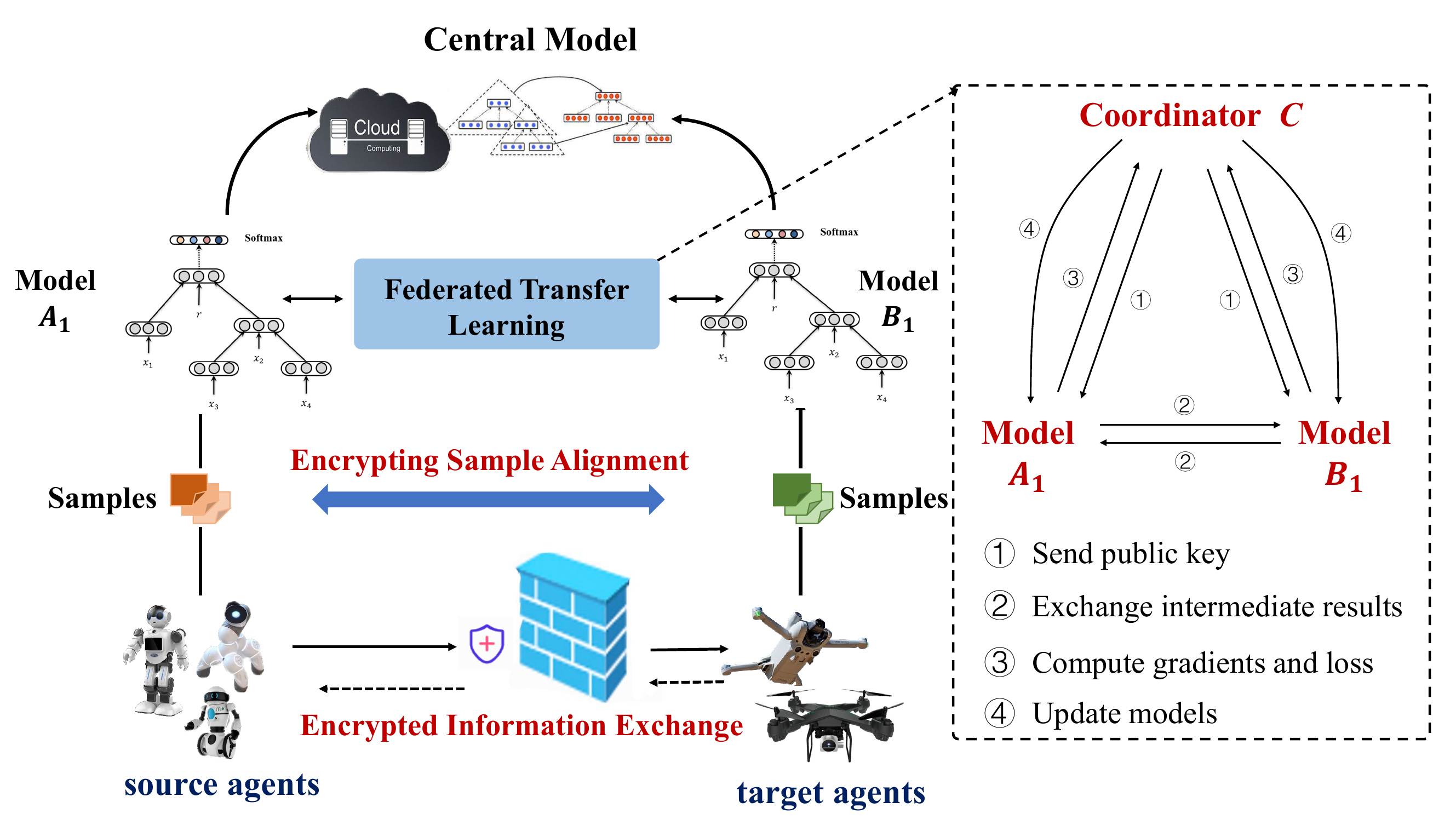}
    \caption{The general framework of Federated Transfer Learning. }
    \label{fig_FTL}
    \vspace{-3mm}
\end{figure}

With the increasing demand for data privacy and the improvement of relevant laws and regulations \cite{voigt2017eu}, there are often various restrictions imposed on the exchange of data between agents. Consider a smart city scenario, where multiple types of intelligent cameras are distributed to capture visual information from specific areas. The tasks, such as traffic control and suspect tracking, require aggregating data from different cameras for algorithm training. However, when these cameras are from different companies or involve the privacy of numerous users, sharing and merging the data becomes impractical. Each camera becomes a data island, impeding the exchange of information between cameras and hindering model learning. To address this challenge, Liu et al. \cite{liu2020secure} propose federated transfer learning (FTL), which integrates transfer learning and federated learning \cite{nguyen2021federated}. The FTL emphasizes collaborative modeling and learning on diverse data distributions, offering a promising research direction in achieving the fusion of data among multiple agents. In the scope of this paper, FTL emphasizes knowledge transfer and fusion among a group of agents, enabling the joint optimization of models under specific requirements and limitations, as shown in Fig. \ref{fig_FTL}. Firstly, different devices train local models based on their own data, and then the device encrypts the model parameters and exchanges intermediate results. Finally, joint training is conducted on these devices to obtain the final optimal model, which is then returned to each device.

FTL methods have the capability to optimize crowd agents on task data with different sample spaces or feature spaces. This is achieved by transferring features from different feature spaces to the same representations through a virtual collaborator or an aggregation server. Additionally, encryption techniques such as homomorphic encryption and random masks are utilized during the model update stage to ensure data privacy and learning security. The aggregation server utilizes the combined local updates from multiple participating agents to iteratively perform model learning and global updates, minimizing the loss function. Besides, Li et al. \cite{li2019fedmd} explore the problem of federated learning when each agent has different models. They propose the FedMD for heterogeneous federated learning. FedMD allows each agent to independently design the architecture of their local models, and knowledge transformation between agents is achieved through the process of KD. Gao et al. \cite{gao2019privacy} introduce a heterogeneous federated transfer learning strategy, namely HFTL, which aims to address covariate shift in homogeneous feature spaces and bridge different agents with heterogeneous feature spaces. HFTL consists of five components: secure domain adaptation, secure feature mapping, secure federated learning, secure model integration, and local model inference. Furthermore, Nguyen et al. \cite{nguyen2022cdkt} propose two mechanisms for knowledge transfer across devices: one is global knowledge transfer, which involves transferring knowledge from client models to a global model; the other is on-device knowledge transfer, which involves transferring generalized shared knowledge from the global model to client models.

In addition, FTL shows remarkable prospects for advancement in numerous domains, including autonomous driving and smart healthcare. For example, Liang et al. \cite{liang2022federated} propose a federated transfer reinforcement learning approach, namely FTRL, where all agents can utilize knowledge from other agents to make effective actions in the current context. Communication between agents and the federated learning server occurs through Wi-Fi, and each agent performs RL tasks in its specific context, while the server periodically aggregates the models from all agents to generate a joint model for crowd knowledge transfer. Chen et al. \cite{chen2020fedhealth} introduce a FTL framework applied to healthcare in wearable devices, dubbed as FedHealth, which utilizes federated learning for data aggregation and incorporates transfer learning to construct personalized wearable models. FedHealth starts by training a cloud model on the server side using a public dataset, which is then distributed to all users. Each user can train their own model utilizing their unique local data. 
Subsequently, the user models are uploaded to the cloud server to facilitate training updates. During the model uploading process, only homomorphically encrypted model parameters are shared, ensuring the privacy of user data and information. Users can perform personalized training by integrating the cloud model with their local model, resulting in a personalized wearable healthcare model. While FTL aims to effectively aggregate data under privacy regulations and achieve knowledge transfer. However, it does impose higher demands on network and computational resources due to frequent encryption and gradient transmission. For example, Jing et al. \cite{jing2019quantifying} conduct performance tests on the open-sourced FTL model, i.e., FATE, deployed on Google Cloud and identify that inter-process communication cost is a major bottleneck of FTL. They also observe that software-based encryption methods consume excessive CPU cycles. Sharma et al. \cite{sharma2019secure} argue that the computational overhead of the FTL framework, specifically when using secure multi-party computation and homomorphic encryption protocols, is significant. To address this issue, they leverage secret sharing to enhance the efficiency and security of model collaborative training in the context of crowd knowledge transfer.

FTL enables the collaborative training of agents across different domains, organizations, or tasks without sharing the raw data to preserve privacy and confidentiality, which is particularly useful in industries like healthcare and manufacturing, where data sharing is restricted by regulations \cite{chen2020fedhealth, gao2019privacy}.  
For instance, in the dynamic landscape of modern manufacturing in smart factories, agents need to rapidly adapt to different applications. Traditional deep learning approaches often falter in new applications across different knowledge domains, primarily due to limited data availability. Furthermore, extensive data sharing among numerous industrial agents raises concerns about the potential exposure of sensitive information. Federated transfer learning emerges as a robust solution to these challenges, promoting knowledge transfer in complicated industrial settings. For emaple, FTL-CDP \cite{kevin2021federated} is a federated transfer learning framework that supports the training of heterogeneous applications for cross-domain tasks in smart factory, such as object detection and facial recognition. This framework integrates a central server with multiple smart devices from various applications. The central server oversees the base models, while the devices, under the federated learning paradigm, share their knowledge without compromising privacy. FTL-CDP not only utilizes existing base models trained across different devices to mitigate data scarcity but also accelerates the learning process on each device through transfer learning.

\subsubsection{Teacher-Student Learning}
Several agents have the ability to acquire high-performance machine learning models based on abundant training data and computation power. Nevertheless, most agents with restricted data face challenges in attaining a similar level of performance during model training. Fortunately, the knowledge acquired by advanced agents can be employed to guide the model training of junior agents, thereby enabling the latter to achieve comparable capabilities. This approach to knowledge transfer method is termed as \textit{Teacher-Student Learning} \cite{xu2021end}. Advanced agents are capable of achieving high performance are denoted as \textit{teachers}, while the junior agents are known as \textit{students}. The teacher, identified as the central authority, assumes responsibility for facilitating students’ learning experiences and enhances the training process of models by utilizing centralized knowledge transfer.

The teacher model is typically constructed using an exponential moving average (EMA) of the student models’ parameters, thus serving as an ensemble of the student model \cite{wang2023consistent}. Initially, a small set of labeled data is utilized to train the initial model. Simultaneously, the teacher leverages unlabeled data to make predictions and incorporates these inferred results as pseudo-labels into the training process of the student model. The objective is for the student model to accurately identify these pseudo-labels and consistently predict outcomes for the augmented input data.

In the realm of semi-supervised learning, researchers have developed various methods for teacher-student learning. For example, the Unbiased Teacher \cite{liu2021unbiased} is a typical approach to student-teacher mutual learning, where the teacher continuously self-updates in different epochs to optimize the student. Meanwhile, as the student’s proficiency improves, it stimulates the teacher to generate more reliable pseudo-labels, thereby further enhancing the student’s capabilities. Similarly, the Humble Teacher \cite{tang2021humble} utilizes the concept of soft labels (or pseudo-labels) to improve students’ task performance. The Dense Teacher \cite{chen2022dense} further introduces the dense pseudo-label mechanism to enrich supervised information provided by the teacher. 
Wang et al. \cite{wang2023consistent} propose the Consistent-Teacher model, incorporating adaptive sample assignment, 3D feature alignment, and GMM-based Threshold. In each training iteration, the student model is supervised using labeled data, whereas the teacher model annotates the unlabeled data and utilizes the augmented data to train the student model. The Active Teacher \cite{mi2022active} utilizes three key factors, i.e., difficulty, information, and diversity, to assess and select unlabeled samples, with the aim of improving the quality of pseudo-labels generated by the teacher model. It is trained to learn from samples that pose more classification challenges, provide richer information, and demonstrate greater semantic diversity. In addition, Matiisen et al. \cite{matiisen2019teacher} propose the TSCL model by integrating curriculum learning into the teacher-student framework. They emphasize that the student model should receive intensive training in tasks where it exhibits rapid progress. To prevent knowledge forgetting, the student model should also undergo training on tasks where its performance starts to decline. The teacher model plays the role of monitoring the student model’s training progress and deciding which tasks the student model should prioritize, operating as a partially observable markov decision process \cite{zhang2022multi}. 

For deploying teacher-student learning approaches on resource-constrained devices, the Efficient Teacher \cite{xu2023efficient} tackles practical engineering challenges associated with deploying deep models on terminal devices like AI boxed and smart cameras. By adopting the mosaic data augmentation technique from YOLO, the Efficient Teacher enhances the generation of more effective pseudo-labels. Furthermore, it introduces the pseudo-label assigner (PLA) that classifies pseudo-labels into unreliable and uncertain label regions. For each category, more refined strategies are employed to incorporate them as the loss function for supervising the student model.

In recent years, teacher-student learning has been effectively and widely embraced on various AIoT tasks \cite{wang2019progressive, chen2021enhancing}, including knowledge distillation, multimodal learning, etc. 
For example, with the widespread adoption of depth sensors such as Kinect, the combination of multimodal information offers a promising way to enhance human activity recognition performance. Many existing methods either focus solely on single data modality or fail to fully leverage the benefits of integrating multiple modalities. To address this issue, TSMF \cite{bruce2021multimodal} adopts a teacher-student learning model to fuse skeleton and RGB modalities for recognizing indoor activities. Different from existing multimodal methods, TSMF includes two modality-specific subnetworks to fuse multimodal information at the model level, where a teacher network transfers the structural knowledge from the skeleton modality to a student network for the RGB modality. The teacher network, a graph convolutional network, extracts features from skeleton data, providing not just modality-specific predictions but also spatial weights that act as an attention mechanism targeting the region of interest within the RGB modality. Meanwhile, the student network, a conventional CNN, assimilates these features to predict values based on the RGB data, which are then combined with the teacher network’s predictions to make an overall prediction.

\subsubsection{Fine-tuning of Large Language Models}
\begin{figure}[h]
\centering
\includegraphics[width=0.5\textwidth,height=4.6cm]{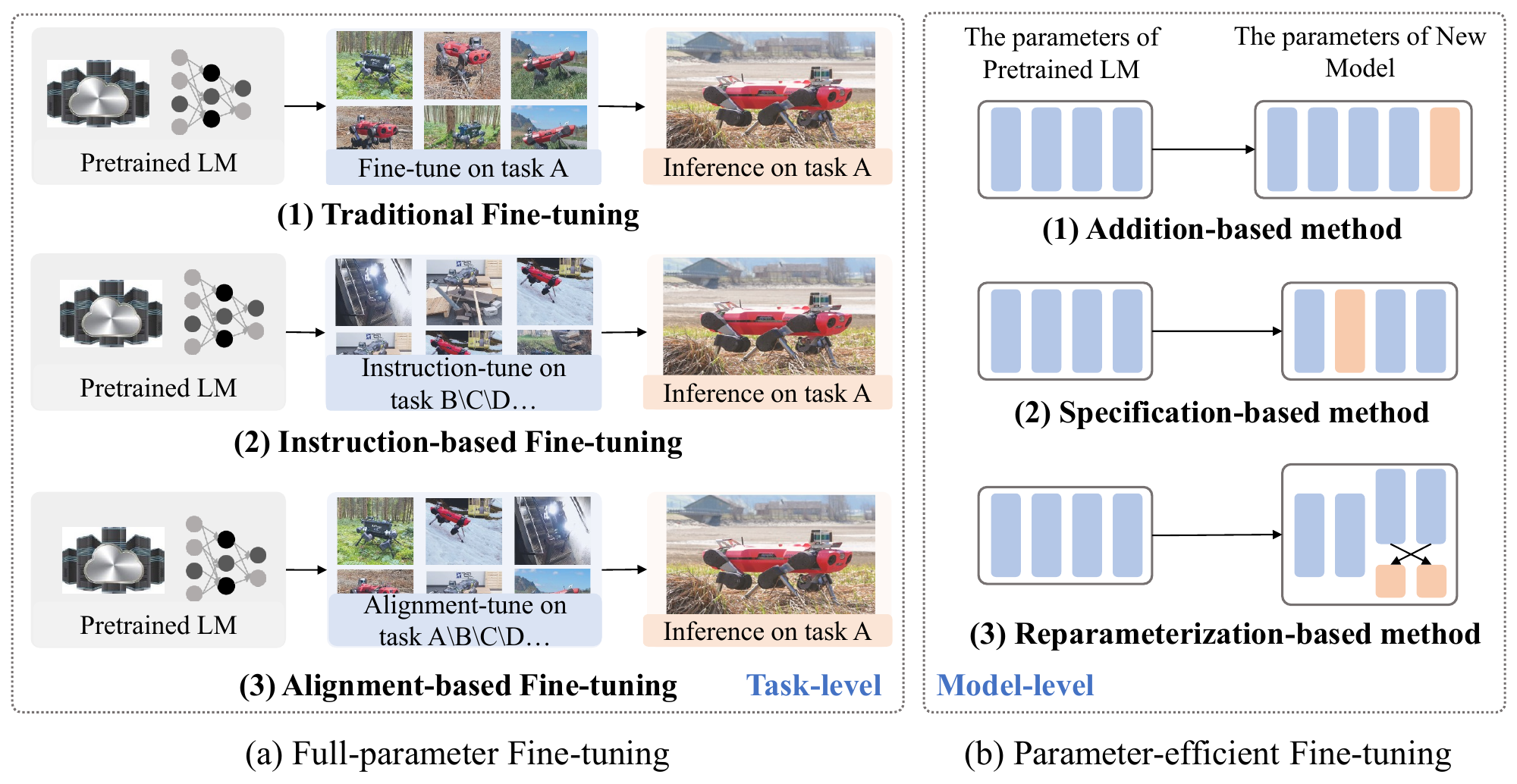} 
\caption{The framework of Fine-tuning of Large Language Models.}
\vspace{-3mm}
\label{Fig:4AIoT_A_3}
\end{figure}

Large language models (LLMs) possess a general capability to solve various tasks or heterogeneous data, but they require global training for specific tasks to achieve optimal performance. In addition, LLMs occupy significant memory and computational resources, requiring a substantial amount of energy and power consumption. However, in the field of AIoT, smart agents with limited resources such as computation, storage and power consumption require fine-tuning and adapting of LLMs to accommodate the resource constraints of the agents. Furthermore, fine-tuning of LLMs enables the model to adapt to the changing context of smart agents in real time, enhancing the performance and efficacy of the model in specific scenarios. Fine-tuning of LLMs is an approach in the derivation mode, which seeks to transfer the acquired knowledge obtained from large models trained on varied and diverse data to enhance the effectiveness of agents in specific tasks. Currently, there are two main approaches for fine-tuning: \textbf{full-parameter fine-tuning} and \textbf{parameter-efficient fine-tuning}, as shown in Fig. \ref{Fig:4AIoT_A_3}.

Full-parameter fine-tuning refers to the process of transferring the rich representations learned by a pre-trained model on a large-scale dataset to a specific task through fine-tuning. It helps accelerate the learning speed and elevate the model's performance on unknown tasks. Currently, there are several methods used for full-parameter fine-tuning, including \textbf{conventional fine-tuning}, \textbf{instruction-based fine-tuning}, and \textbf{alignment-based fine-tuning}. Conventional fine-tuning methods, fine-tune pre-trained models for a specific task to enhance its performance. However, this kind of method requires a substantial amount of high-quality data to train the model. Otherwise, overfitting may occur, resulting in decreased generalization performance. Instruction-based fine-tuning methods involve collecting examples in an instruction format to fine-tune large models. This enhances the model's ability to follow human instructions, enabling better generalization to unknown tasks. For example, BLOOMZ-P3 \cite{muennighoff2022crosslingual} fine-tunes BLOOM on a pure English P3 task set and achieves over 50\% performance improvement compared to BLOOM. Alignment-based fine-tuning employs reinforcement learning techniques to further fine-tune large models based on human feedback data. This allows LLMs to align with human preferences such as usefulness, honesty, and harmlessness according to desired instructions. For instance, InstructGPT \cite{ouyang2022training} leverages the reinforcement learning method based on the human feedback to fine-tune the pre-trained model, resulting in answers that are twice as informative and accurate as those generated by GPT-3.

Full-parameter fine-tuning methods fine-tune all parameters based on abundant computational resources and training data, which is not suitable for resource-constrained 
 devices. In contrast, parameter-efficient fine-tuning aims to reduce the amount of trainable parameters while maximizing the performance on the target task. 
 Current parameter-efficient fine-tuning methods can be classified as \textbf{addition-based method}, \textbf{specification-based method}, and \textbf{reparameterization-based method}. 
 Addition-based method inserts lightweight trainable neural network modules or adjustable parameters into the model. Only a small portion of parameters is fine-tuned to achieve efficient adaptation. 
 For example, adapter-tuning \cite{houlsby2019parameter} adds a simple neural network to each layer of the pre-trained model, enabling it to adjust to subsequent tasks. By freezing the main body of the pre-trained model during fine-tuning and training task-specific parameters only, adapter-tuning can achieve comparable performance to full-parameter fine-tuning while reducing the computational cost. Prompt tuning \cite{li2021prefix} is another method that adds trainable prompt tokens specific to the task at the input and all layers of the pre-trained model. Specification-based method refers to training specific parameters while freezing other parameters. For example, BiiFit \cite{ravfogel2021bitfit} only optimizes the biases within the model and freezes other parameters, reproducing over 95\% of the performance achieved by full-parameter fine-tuning in multiple benchmark tests. Reparameterization-based method assumes that the adaptation process of the pre-trained model is essentially low-rank or low-dimensional. Therefore, these methods aim to reparameterize the existing optimization process into a parameter-efficient manner. For example, LoRA \cite{hu2021lora} approximates the parameter weight matrix of the model by learning a low-rank matrix with small parameters. During training, only the parameters of the low-rank matrix are optimized.

Currently, fine-tuning of LLMs reduces computational resources while improving the performance of models on specific tasks, playing an important role in natural language processing, and has been widely applied in application scenarios such as urban computing and smart factories. For example, in the field of urban computing, urban scenes are constantly changing in both temporal and spatial dimensions. At the same time, due to the high cost of deploying sensors in the whole city space, the available urban data is sparse, which poses significant challenges to urban computing tasks. Fine-tuning of LLMs can converge various urban computing tasks, understand the complex interdependencies between urban data in time, space, and various tasks, and make more comprehensive and accurate predictions in sparse data \cite{zhang2024towards,chang2024llm4ts}. For example, UrbanGPT \cite{li2024urbangpt} first proposes the Spatio-Temporal Dependency Encoder module to enhance the ability of LLMs to capture temporal dependencies in spatio-temporal contexts, and then proposes the Spatio-Temporal Instruction-Tuning module to integrate spatio-temporal context signals with the inference capabilities of LLMs seamlessly to improve the model's prediction performance. Specifically, spatio-temporal-text alignment is first performed to capture complementary information and extract high-level semantic representations with more representational capabilities. Then multi-granularity temporal information and spatial details are integrated as instruction inputs to LLMs to recognize and understand spatio-temporal patterns in different regions and periods in complex spatio-temporal environments to enhance its zero-sample inference capability. Finally, spatio-temporal instruction-tuning of LLMs based on regression strategies is used to utilize contextual information to generate more accurate spatio-temporal predictions.

\subsubsection{CTDE-based Multi-agent Reinforcement Learning}
\label{section_CTDE_MARL}
The core of reinforcement learning (RL) lies in enabling an intelligent agent to acquire the ability to make suitable sequential decisions within specific time intervals, in order to address the challenges encountered in both societal and engineering contexts. This ability is developed through continuous interaction with the environment and a process of trial and error. Real-world problems, in fact, possess a high level of complexity. The majority of tasks involve large-scale systems that can be broken down into smaller sub-tasks, each assigned to distinct individuals who operate according to predefined rules or common understanding. The accomplishment of a task necessitates simultaneous participation from multiple intelligent agents. These agents will disperse and carry out tasks within their designated sub-spaces. However, at the task level, it is essential for these agents to collaborate with one another, as the outcomes of their respective sub-decisions profoundly influence each other. This type of system is commonly referred to as a multi-agent system (MAS).

Within a multi-agent system, the interrelationship between agents becomes crucial for overcoming the challenges posed by an environment where complete knowledge is not readily accessible. In conclusion, the application of multi-agent reinforcement learning (MARL) to facilitate the knowledge transfer among agents holds immense practical significance and is of utmost urgency. In fact, most of MARL follows the \textit{centralized distributed execution} (CTDE). During the training stage, crowd agents share a centralized training environment, from which they learn jointly through shared experiences; however, in the execution stage, each agent acts independently in its own decentralized environment. For instance, Qmix \cite{rashid2020monotonic} employs a combination of nonlinear networks to combine the Q-values of each agent. This combination ensures the coherence between the joint action value of Qmix and the monotonicity of each individual agent. This guarantee maintains consistency between centralized training and decentralized execution. The concept of value function decomposition effectively addresses the credit assignment challenge in multi-agent systems and significantly improves learning performance in these environments. Similarly, MADDPG \cite{lowe2017multi} enables the Critic component of each agent to acquire action information from all other agents, thereby facilitating centralized training and decentralized execution. In training, a Critic that possesses knowledge of the global state guides the training of the Actor component. In testing, only Actors with access to local observations execute actions. This algorithm eliminates the need for establishing and implementing communication rules while effectively mitigating non-stationary problems in multi-agent environments. The COMA \cite{foerster2018counterfactual} embraces the concept of centralized training and decentralized execution, integrating counterfactual baselines to resolve credit assignment issues. In a cooperative agent system, an agent's contribution to action performance is evaluated by comparing the desirability of an action to a specifically selected default action. Furthermore, Gupta et al. \cite{gupta2017cooperative} integrate the parameter sharing framework with the DQN, and DDPG algorithms and deployed them in a multi-agent environment that employs local observations.

In CTDE-based MARL, crowd knowledge transfer primarily occurs during the centralized training stage. Agents learn from the collective experiences of all agents in the shared environment, with multiple agents' value functions or policies being jointly trained, thereby enabling them to acquire knowledge from different perspectives and interactions. For example, M3DDPG \cite{li2019robust} is particularly adept at handling scenarios with multiple autonomous vehicles navigating through dynamic and potentially adversarial environments, such as busy urban traffic systems. In the learning phase, the algorithm uses deep neural networks to predict the best actions for each vehicle based on its current state observations. This predictive model is trained on data gathered from simulations of traffic scenarios. Importantly, the training process involves not just learning to cope with fixed rules of the road but also adapting to the unexpected maneuvers of other drivers, which are treated as adversarial inputs in the training phase. To enhance its robustness, the MDDPG algorithm employs a min-max strategy. This approach adjusts each agent's policy to perform optimally even in the worst-case scenario, presumed to be caused by the most challenging actions of other drivers. Essentially, the policy training aims to find a balance where each agent's decision maximizes its reward in the face of the worst possible strategies from others. Through iterative training, testing, and refining of strategies in a variety of traffic conditions and scenarios, the agents learn to make more sophisticated decisions that enhance their ability to navigate safely and quickly.

\subsection{Decentralized Inter-agent Knowledge Transfer}
\subsubsection{Distributed Multi-agent Reinforcement Learning}
The CTDE-based MARL described in section \ref{section_CTDE_MARL} utilizes the learned policies in a truly decentralized execution environment after training, where each agent just accesses local observations. While such methods demonstrate strong convergence, crowd agents often encounter scenarios in the complex and dynamic AIoT contexts that were unseen during the training stage. Consequently, researchers have delved into a technique that more closely aligns with real-world application environments, namely distributed MARL. 

Unlike CTDE-based MARL, distributed MARL is decentralized during both the training and execution stages. Each agent operates within its own observations throughout the training and execution processes, but could adjust its policy based on interactions with other agents. For example, Heinrich et al. \cite{heinrich2016deep} introduce the Neural Fictitious Self-Play (NSFP), a method that combines deep reinforcement learning with fictitious self-play to learn strategies in imperfect-information games without prior knowledge. NSFP employs distributed training across independent agents, efficiently handling large-scale multi-agent environments. Sun et al. \cite{sun2020tleague} introduce TLeague, a robust framework for competitive self-play based distributed MARL, designed to handle the intensive data demands of training high-performance AI for complex games. TLeague facilitates scalable distributed training on a mixture of CPU and GPUs, supporting both single-machine and cluster deployments with Kubernetes, enhancing MARL's accessibility and efficiency in real-world applications. The architecture includes modular components for \textit{Actor}, \textit{Learner}, and \textit{InferenceServer}, optimizing the parallel training process and providing flexibility to extend for various multi-agent problems and new MARL algorithms. Xu et al. \cite{xu2023distributed} present a Distributed-Training-Distributed-Execution (DTDE) MARL scheme for power control in heterogeneous networks (HetNets). They introduce a penalty-based Q-learning (PQL) algorithm that allows each access point in a HetNet to independently make power control decisions based on local information, promoting more efficient cooperation among agents.

Knowledge transfer in distributed MARL occurs through the process of agents interacting and observing within their respective environments. It often relies on communication mechanisms between agents \cite{kim2020communication, liu2023deep, yuan2023dacom}, facilitating the exploration of more diverse learning trajectories and a broader range of task solutions. Taking the electric vehicle (EV) charging schedules as an example, Zhang et al. \cite{zhang2023distributed} adopt the DTDE-based Stackelberg MARL framework. During the training phase, each agent, both EVA and EVs, independently trains local neural networks using only locally available data. This includes their observations and actions, as well as the actions of interacting agents, without the need to share sensitive private information. This approach addresses major concerns related to privacy and communication overhead seen in centralized systems. The EVA, acting as the leader, sets dynamic pricing signals based on the observed state and projected demands, which are then broadcasted to the EVs. In response, each EV, as a follower, decides its charging strategy based on the current price, aiming to minimize its charging costs while adhering to operational constraints such as battery capacity and required charge levels. In the execution phase, the framework facilitates real-time decision-making where each agent, equipped with its policy model, reacts based on its current observation. The distributed execution ensures that all computations, including forward and back-propagation through the neural networks, are localized, significantly enhancing the computational efficiency and scalability of the system. This setup eliminates the dependency on a central coordinator and high-precision modeling, making it robust against the non-stationarity typical of large-scale EV charging scenarios.

\subsubsection{Imitation Learning}
Imitation learning (IL) aims to extract knowledge from demonstrations/trajectories provided by human experts or agents in order to replicate their observed behavior \cite{zheng2022imitation}.  It has garnered significant attention in diverse domains, including autonomous driving \cite{codevilla2018end, pomerleau1988alvinn} and robot simulation \cite{finn2016guided, nair2017combining}. Crowd agents can independently tackle encountered tasks by spontaneously imitating the behaviors exhibited by neighboring agents, enabling decentralized knowledge transfer through imitation. In contrast to RL, IL belongs to supervised learning and involves acquiring labeled training data from expert demonstrations, whereas RL employs reward functions to guide the learning process of agents without relying on labeled data. At present, the prevailing IL methods primarily comprise behavioral cloning-based IL and inverse reinforcement learning-based IL.

\begin{figure}
    \centering
    \includegraphics[width=0.45\textwidth,height=7.4cm]{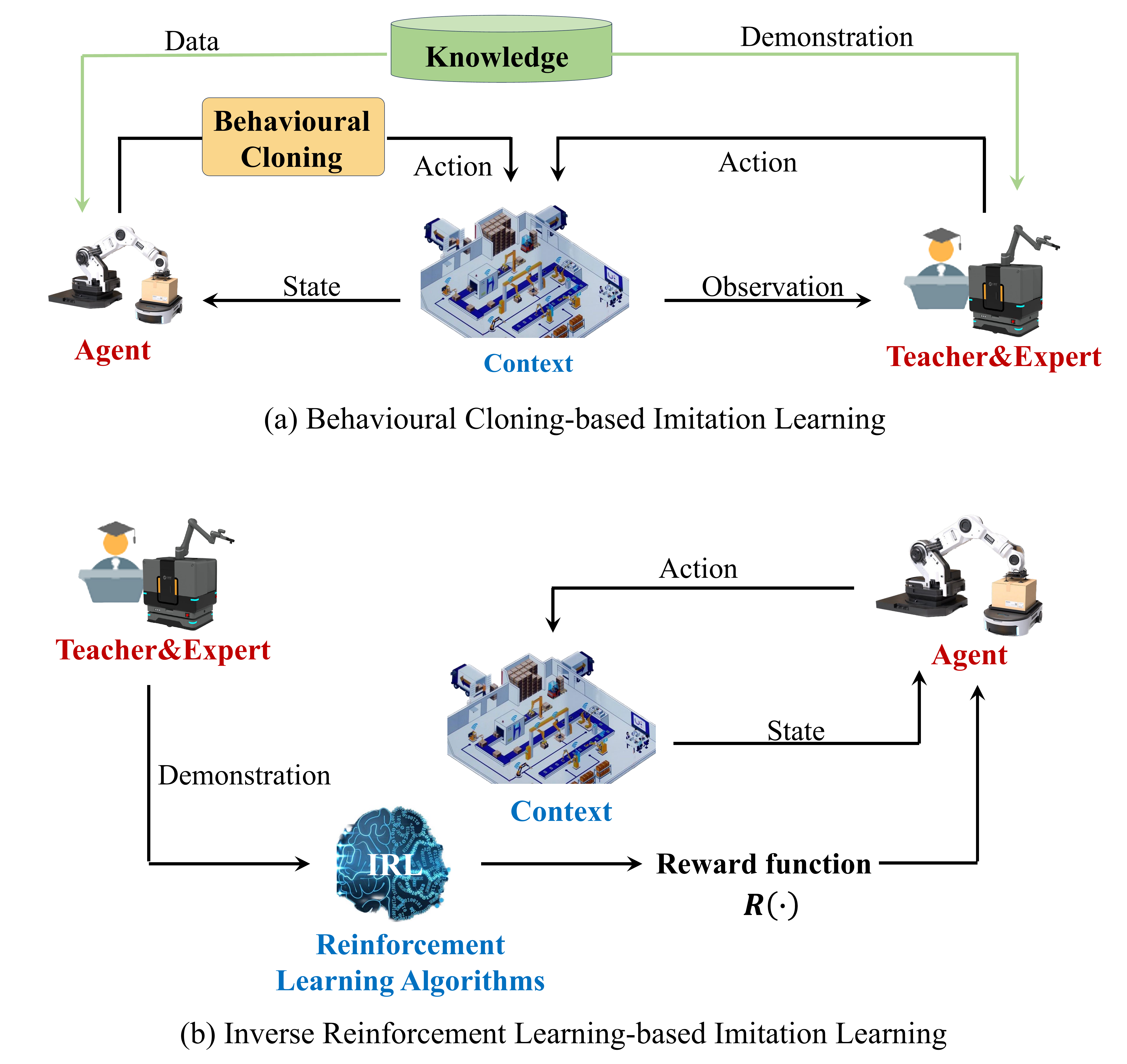}
    \caption{The general framework of Imitation Learning. }
    \label{Fig_IL}
        \vspace{-3mm}
\end{figure}

Behavioural cloning-based IL directly utilizes demonstrations provided by experts/teachers to map states or contexts to the needed actions and optimizes them through supervised learning, as shown in Fig. \ref{Fig_IL}(a). For instance, Dagger \cite{ross2011reduction} leverages data aggregation to enhance the agent’s ability to generalize in unseen scenarios. It engages in continuous interaction with the environment by amalgamating the policy derived from behavioral cloning with freshly generated data. It subsequently solicits expert guidance to obtain policy examples on the novel data, leading to the creation of an expanded dataset. Subsequently, Dagger undergoes iterative training on the expanded dataset using behavioral cloning and continues to interact with the environment. Duan et al. \cite{duan2017one} aim to enable an agent to learn the behavior of other agents with limited observation. This necessitates the learning of meta knowledge by neural networks through one-shot demonstrations. In turn, this knowledge enables them to comprehend the intention of the demonstration and directly map it to control outputs. Therefore, they incorporate soft-attention with meta-learning to accomplish this objective, which consists of the demonstration network, context network, and manipulation network. During training, the model simultaneously takes a demo and samples a state from another demo, generating the corresponding predicted action for that state.

Furthermore, many real-life tasks may suffer from the absence of expert data or the absence of an effective teacher agent. To tackle this problem, Lynch et al. \cite{lynch2020learning} opt to facilitate knowledge transfer for crowd agents using play data rather than relying on expert data. Play data is acquired through human interactions and may not demonstrate optimal performance in particular tasks. Nevertheless, it gradually accumulates experience for the agent, instills curiosity, and fosters self-improvement. They couple arbitrary agent trajectories with natural language instructions and train a model to replicate behaviors based on the linguistic descriptions.

In the context of inverse reinforcement learning-based IL, agents iteratively recover and evaluate reward functions from expert demonstrations, necessitating stronger computational capabilities to obtain distinct reward functions and tackle the issue of sparse rewards, as shown in Fig. \ref{Fig_IL}(b). For example, Ho et al. \cite{ho2016generative} propose generative adversarial imitation learning method (GAIL), a method that integrates GAN and maximum entropy to learn policies directly from expert data. The GAIL utilizes a generator to compute occupancy measures, which is then evaluated by a discriminator to assess its similarity to the occupancy measures of the expert policy. When the discriminator is unable to distinguish between the two measures, it is deemed that the agent has successfully acquired the expert’s expertise, thereby accomplishing knowledge transfer. Importantly, GAIL does not rely on expert interaction during training and has the capability to handle distribution shifts. Moreover, Sermanet et al. \cite{sermanet2018time} employ a combination of self-supervised learning to facilitate the agent’s ability to imitate behaviors from human expert videos. The proposed method involves selecting a frame from the video as well as choosing neighboring frames as positive examples and randomly selecting distant frames as negative samples. Subsequently, they train DNNs with these selected samples and utilize RL model to achieve agent control.

Furthermore, Kim et al. \cite{kim2020domain} introduce the generative adversarial MDP alignment framework, namely GAMA, for cross-domain IL with non-temporally aligned demonstration data. They initially tackle the MDP alignment problem and subsequently employee adversarial training to align and replicate expert strategies across diverse domains. Similarly, Raychaudhuri et al. \cite{raychaudhuri2021cross} develop a state mapping function that translates between the expert domain and agent domain using demonstrations from a series of proxy tasks originating from two distinct domains. They subsequently refine the acquired mapping on the target task. After acquiring the mapping function, it can utilize expert demonstrations on the target task as input to generate agent demonstrations. Finally, IL methods are employed to acquire the agent’s strategy in the target task.

Imitation learning has been utilized in many complex and unstructured environments, such as autonomous driving \cite{le2022survey} and aerial robotics \cite{tejaswi2022constrained}.
For example, Teng et al. \cite{teng2022hierarchical} develop a hierarchical two-stage imitation learning model for end-to-end autonomous driving. The first stage involves training a Bird's Eye View (BEV) model from the Carla Simulator, which helps the system understand its surroundings. The second stage introduces an interpretable imitation learning model that combines the BEV features from the first stage with steering angles generated by the Pure-Pursuit algorithm. This stage employs a multi-branch mechanism that responds to high-level commands to individually train each output branch, enhancing the model's ability to handle diverse driving scenarios.

\subsubsection{Decentralized Federated Learning}
Decentralized federated learning (DFL) is an expansion of the federated learning (FL), focused on not depending on a single central server for the training process. Instead, it distributes the model training process among multiple agents \cite{gabrielli2023survey}. It enhances the decentralization of power, prioritizes data privacy protection and distributed computing, while reducing the risks of single-point failures and central control. This paper classifies existing methods into two broad categories: \textit{traditional decentralized federated learning} and \textit{blockchain-based decentralized federated learning}.

Traditional decentralized federated learning: This category of methods achieves DFL through commonly used distributed computing techniques. For example, IPLS \cite{pappas2021ipls} is a fully DFL framework that incorporates elements of the interplanetary file system (IPFS). By utilizing IPLS and connecting to the respective private IPFS network, any agent can initiate or join the training process of ML models. It is designed to scale as the number of agents grows, ensuring resilience to intermittent connections and dynamic agent arrivals and departures. IPFS demands minimal resources while guaranteeing rapid convergence of the training model accuracy to that achieved by a centralized FL, with precision drop of less than 1\%. In addition, ProxyFL \cite{kalra2023decentralized} adopts the approach of each agent maintaining two models: The proxy model enables effective information exchange among agents, eliminating the necessity for a central server. This approach overcomes a significant constraint of conventional FL by enabling model heterogeneity, granting each agent the freedom to have a private model with any architecture. Furthermore, the proxy communication protocol employed in ProxyFL incorporates differential privacy analysis, providing stronger guarantees for privacy. Tang et al. \cite{tang2022gossipfl} propose a communication-efficient DFL framework called GossipFL. It introduces innovative sparsification and Gossip matrix generation algorithms. The sparsification algorithm enables each agent to communicate exclusively with a highly sparse counterpart. This algorithm facilitates each agent’s communication with just one peer during each communication round, enabling the exchange of highly compressed models. Consequently, this reduces upstream and downstream communication traffic, and ensures convergence while better utilizing bandwidth resources.

Blockchain-based decentralized federated learning: The methods in this category utilize blockchain functionality to achieve DFL. Each device participating in the learning process has the ability to act as a leader and guide the aggregation process for a specific learning round. This decentralized structure eliminates the requirements for a central server and enhances the system’s fault tolerance. For instance, Kim et al. \cite{kim2019blockchained} propose the BlockFL where the blockchain network facilitates the exchange of local model updates among agents, performing verification and providing corresponding rewards. BlockFL addresses the concern of single-point failures and expands its federated scope to encompass untrusted agents on the public network, achieved through a validation process of local training results. Additionally, by providing rewards proportional to the size of the training samples, BlockFL incentivizes increased collaboration from devices with a larger training sample size. DeepChain \cite{weng2019deepchain} introduces a collaborative training method with incentives for training DNNs. It encourages participation and the exchange of local gradients, prioritizing privacy protection and the verifiability of the training process. Through incentives and transactions, DeepChain encourages agents to act honestly during gradient collection and parameter updates, fostering fairness during collaborative training. They formalize this incentive mechanism based on blockchain technology, emphasizing compatibility and activity, and demonstrate a high likelihood of agents being appropriately incentivized. Hu et al. \cite{hu2020gfl} propose the Galaxy FL framework, namely GFL. GFL utilizes the consistent hashing algorithm to construct a ring topology of agents, with the aim of reducing communication pressure and enhancing topology stability. Moreover, GFL introduces the ring-decentralized federated learning (RDFL) algorithm to improve bandwidth utilization and the performance of DFL.

Decentralized federated learning is a decentralized network architecture that eliminates the need for a central server in contrast to centralized federated learning, which has been used in many scenarios to enable real-time applications \cite{pokhrel2020decentralized, xiao2021fully}. 
For example, in real-time UAV applications such as autonomous monitoring and target tracking, the challenges of high latency and significant resource consumption are critical. Transmitting raw data to the cloud, which requires substantial bandwidth and energy, is often impractical due to the constrained resources of UAV networks. Additionally, privacy concerns may restrict access to data generated by individual UAVs. To address these issues, applying decentralized federated learning within UAV networks can be extremely advantageous. Decentralized federated learning eliminates the need for central aggregation of the global model, thereby preserving bandwidth and enhancing privacy. For example, DFL-UN \cite{qu2021decentralized} is a decentralized federated learning for UAV networks, which does not need a central entity for global model aggregation. In DFL-UN, each UAV updates its local model using its own data and then shares the model weights with neighboring UAVs. The process includes three steps: initially, UAVs send their local model weights with a target UAV. Subsequently, the target UAV combines these weights with its own to create an updated local model. Finally, this newly formed local model is broadcast back to the neighboring UAVs for further aggregation and local updates.

\subsection{Intra-agent Knowledge Transfer}
\subsubsection{Multimodal Learning}
In AIoT scenarios, there are typically various data acquired in multiple ways such as sensory, visual, audio, etc. For example, in autonomous driving scenarios, cameras and LiDAR data capture information about surrounding vehicles, buildings, etc., and on-board sensor data, such as speed and steering angle, can be utilized to perceive the vehicle's own information, and the comprehensive fusion of these perceptual information can provide autonomous driving systems with more comprehensive perception and more accurate decision-making. Therefore, facing the challenge of heterogeneous data, the multimodal learning approach plays an important role, which can eliminate the redundant information from different modalities, represent the target in a more comprehensive and multidimensional way to enhance the accuracy and robustness of the overall system \cite{yan2022mtffn}.

\begin{figure*}[htbp]
\centering
\includegraphics[width=0.7\textwidth,height=5.2cm]{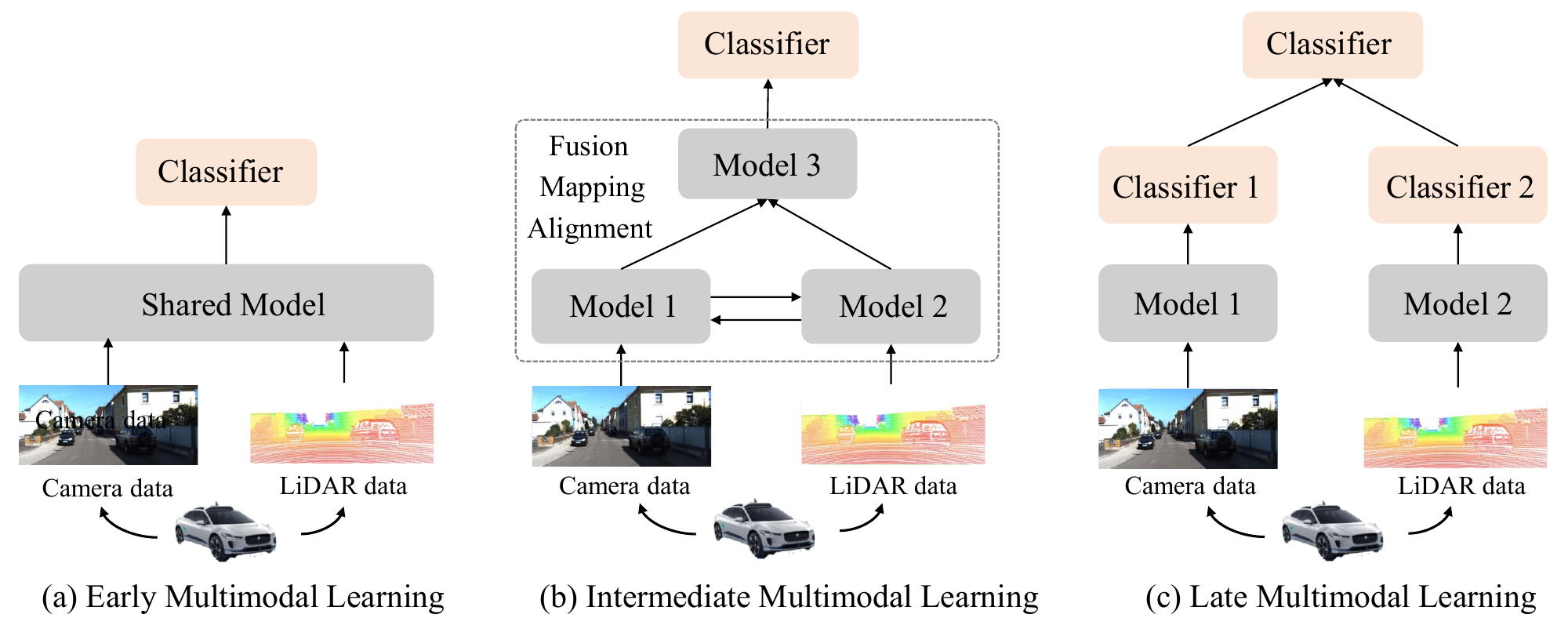} 
\caption{The framework of Multimodal Learning.}
\label{Fig:4AIoT_C_1}
\vspace{-3mm}
\end{figure*}

The multimodal learning method integrates information from different modalities, belongs to the fusion mode. It is based on the criterions of complementarity and consistency, and can be classified into three categories: \textbf{early multimodal learning}, \textbf{intermediate multimodal learning} and \textbf{late multimodal learning}. As shown in Fig. \ref{Fig:4AIoT_C_1}, The early multimodal learning method mainly refers to the transfer and fusion between data. For example, Lv et al. \cite{lv2020survival} predict survival in patients with colon adenocarcinoma based on RNA Sequencing data, miRNA Sequencing data and DNA methylation data. The intermediate multimodal learning method refers to the transfer and fusion between features, and usually contains methods such as feature mapping, feature alignment, and feature fusion. The goal of feature mapping method is to map the source modality to the target modality, so as to realize the enhancement of the target modality data and performance improvement. For example, Moon et al. \cite{moon2014multimodal} map the audio data to lip-reading video data to compensate for the shortcoming of lip-reading video data. The feature alignment method aims to mine the correlation between sub-elements of multimodal data and find correspondences between elements of different modalities, so as to realize the matching of inter-modality elements. For example, Zhen et al. \cite{zhen2020deep} map the text and image data to be aligned in the same space, which makes it possible to retrieve the data from one modality based on the data from another modality. The feature fusion method aims to fuse information from source and target modalities, represent the target in a more comprehensive and fine-grained way, and improve the system performance. For example, Li et al. \cite{li2021toward} employ RGB visual stream and skeleton stream to predict ongoing human actions. And the feature fusion method aims to information transfer and fuse at the decision-making level. For example, Deng et al. \cite{deng2020multimodal} employ chemical substructures, targets, enzymes and pathways features to jointly predict the drug-drug interaction events.

The early multimodal learning method aims to train a shared model, which is a simple and controllable method. However, it needs to ensure the data homogeneity across modalities. In the late multimodal learning method, each modality is handled separately, and the model can be trained normally even if some modality information is missing. This provides high flexibility but may not fully leverage the correlation between low-level features across modalities. Additionally, training multiple modalities separately can introduce significant computational complexity. The intermediate multimodal learning method can capture the fine-grained correlation between modalities, reducing the variability between modalities, which is flexible and widely utilized.

At present, multimodal learning methods have been widely applied in fields such as human activity recognition, urban computing, connected vehicles, healthcare, etc. For example, in the field of connected vehicles, autonomous vehicles usually are equipped with sensors of different modalities, such as cameras, lidar, radar, global navigation satellite systems, etc \cite{feng2020deep}. Existing methods typically utilize the perception of single-mode data, such as cameras capturing detailed texture information of lower environments in front of the field of view \cite{yoo20203d}. However, in complex scenes, objects are easily occluded, posing serious challenges to object detection and semantic segmentation. Lidar can provide accurate depth information in the environment in the form of 3D point clouds, but it is susceptible to extreme weather conditions \cite{bijelic2020seeing}. Therefore, in the field of connected vehicles, autonomous vehicles integrate different modes of perception data, and utilize multimodal learning methods of early fusion, intermediate fusion and late fusion to perceive a more comprehensive and accurate environment, and improve the performance and safety of autonomous vehicles \cite{huang2022multi}. For example, CrossFuser \cite{wu2023crossfuser} first utilizes convolutional neural networks to perform preliminary feature extraction on image and lidar data, and then applies joint mapping and elastic entanglement methods to the features of these two modalities to improve the domain generalization performance of the model. Finally, the multi-head attention mechanism is used to fuse the processed features with the current speed of the autonomous vehicle to generate a more accurate and comprehensive environmental representation and improve the performance of the autonomous vehicle.

\subsubsection{Curriculum Learning}\label{chapter:4.3.2}
\begin{figure}[!t]
\centering
\includegraphics[width=0.33\textwidth,height=7.5cm]{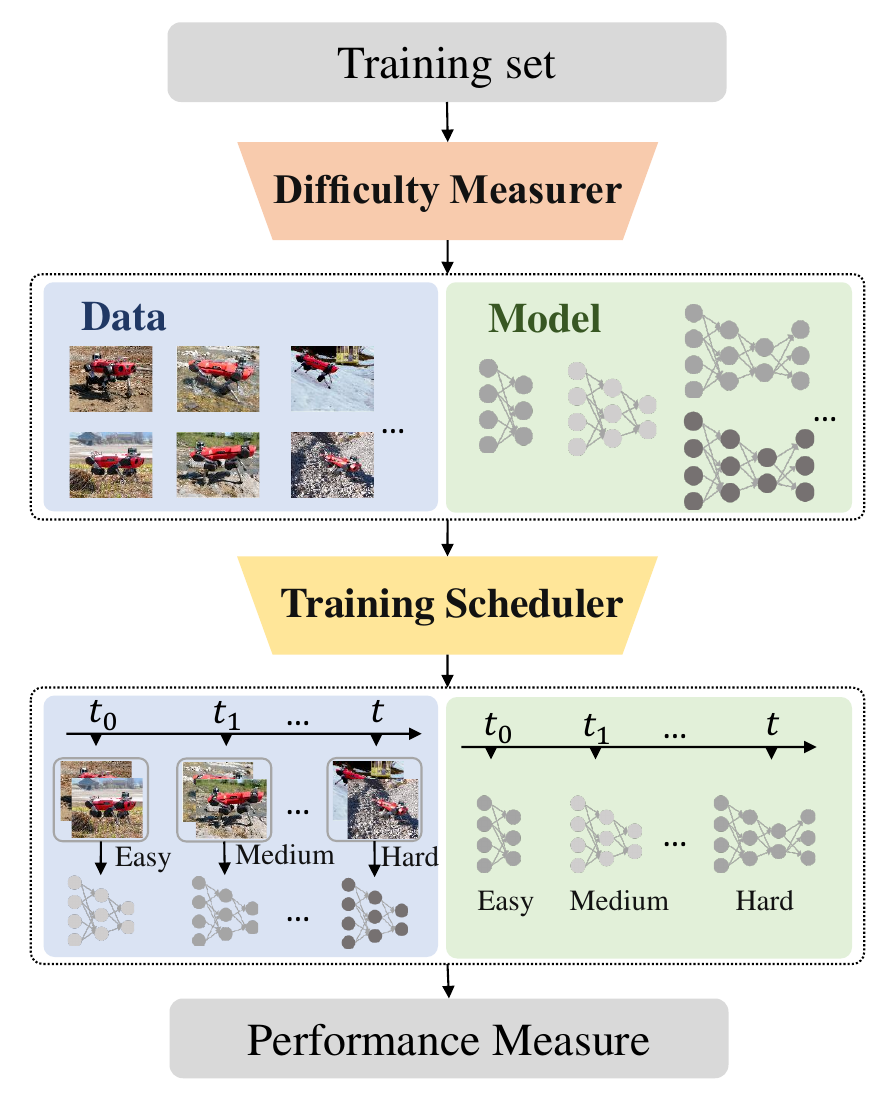} 
\caption{The framework of Curriculum Learning.}
\label{Fig:4AIoT_C_2}
\vspace{-3mm}
\end{figure}

In the field of AIoT, AIoT agents face a constant stream of complex and diverse tasks, and their environments and data distributions are constantly changing. To address the challenges of such incremental tasks, curriculum learning offers a promising solution. Curriculum learning was first proposed in 2009, and its core definition is to start learning models from easy samples and gradually advance to complex samples\cite{bengio2009curriculum}, belonging to the evolution mode. Arrange a series of "curriculum tasks" between the source task and the target task, allowing the trained object to continuously learn new knowledge, and ultimately achieving the knowledge transfer from the source task to the target task. Specifically, as shown in Fig. \ref{Fig:4AIoT_C_2}, curriculum learning first quantifies the difficulty level of training samples (data or models) in the dataset, sorts the training samples, then samples a batch of training samples for training, and finally evaluates the effectiveness of the machine learning model based on the designed evaluation indicators. In this process, it is necessary to determine the relative “easiness” of each data example (Difficulty Measurer) and arrange the sequence of data subsets through the training process (Training Scheduler) \cite{wang2021survey,soviany2022curriculum}. Based on whether these two aspects are automatically designed, curriculum learning can be divided into \textbf{predefined curriculum learning} and \textbf{automatic curriculum learning}.

Predefined curriculum learning methods \cite{liu2020norm} refer to the priori knowledge designed by human experts to be applied to guide the design of Difficulty Measurer and Training Scheduler. In this process, human experts can be regarded as teachers and machine learning models can are regarded as students. As a result, predefined curriculum learning methods fail to feed student feedback into the teacher and are not flexible enough. In addition, it is difficult to require additional expert knowledge and find the optimal combination of both the Difficulty Measurer and the Training Scheduler. Therefore automatic curriculum learning methods aim to design automatically these two aspects based on data or model, which can be dynamically adjusted by the current training. The existing automatic curriculum learning can be divided into four categories: self-paced learning (SPL), Transfer Teacher, RL Teacher and Other Automatic curriculum learning. SPL \cite{kumar2010self} utilizes students themselves as teachers to measure the difficulty of the training samples based on their losses, i.e., to decide the pace of learning based on their current situations. But when the student model is not mature enough, the SPL strategy exists the uncertainty risks. Therefore, Transfer Teacher methods \cite{hacohen2019power} leverage a pre-trained model to act as a teacher and measures the difficulty of the training sample based on the teacher's performance on the training samples. The Difficulty Measurer is automatically designed on both SPL and Transfer Teacher methods, but the Training Scheduler remains pre-defined. Thus RL Teacher methods \cite{matiisen2019teacher} employ RL as the teacher and make dynamic data selection based on the student feedback, i.e., the student makes more progress based on the teacher's one-to-one instruction, and the teacher adjusts his/her own teaching strategy based on the student's feedback so that he/she teaches better. While Other Automatic curriculum learning \cite{tsvetkov2016learning,hung2018adversarial} applies other optimization techniques to find the best lessons automatically, such as Bayesian optimization (BO), stochastic gradient descent (SGD), meta-learning, and hypernetwork.

Currently, curriculum learning has been extensively utilized in fields such as connected vehicles, healthcare, multi-robot systems, smart factories, etc. For example, in the field of multi-robot systems, the real-world environment is complex and dynamic, and developing a complex agent system directly from scratch for each new environment would incur significant costs. In addition, in the real world, it is necessary to consider the collaboration and competition of multi-agent systems, as well as the scalability of multi-agent systems, to achieve system efficiency and stability. Therefore, in the field of multi-robot systems, curriculum learning methods can gradually transfer knowledge between multi-agent systems to enhance and improve the robustness and generalization of robots, as well as the efficiency and stability of multi-robot systems. The multi-agent path discovery task aims to find conflict-free paths from the starting position to the target position for multiple agents. CPL \cite{zhao2023curriculum} is a curriculum-based path discovery learning approach that enables agents to start with simple single-agent path-finding skills and gradually learn cooperative strategies through network parameter inheritance. Specifically, CPL consists of three phases: the first phase motivates agents to complete single-agent tasks through individual rewards. The second phase motivates agents to complete their respective path-finding tasks in a multi-agent environment through individual rewards. The third phase motivates agents to cooperate through team rewards to complete the whole team path-finding task.

\subsubsection{Continual Learning} \label{chapter:4.3.3}
\begin{figure}[!b]
\centering
\includegraphics[width=0.42\textwidth,height=5.0cm]{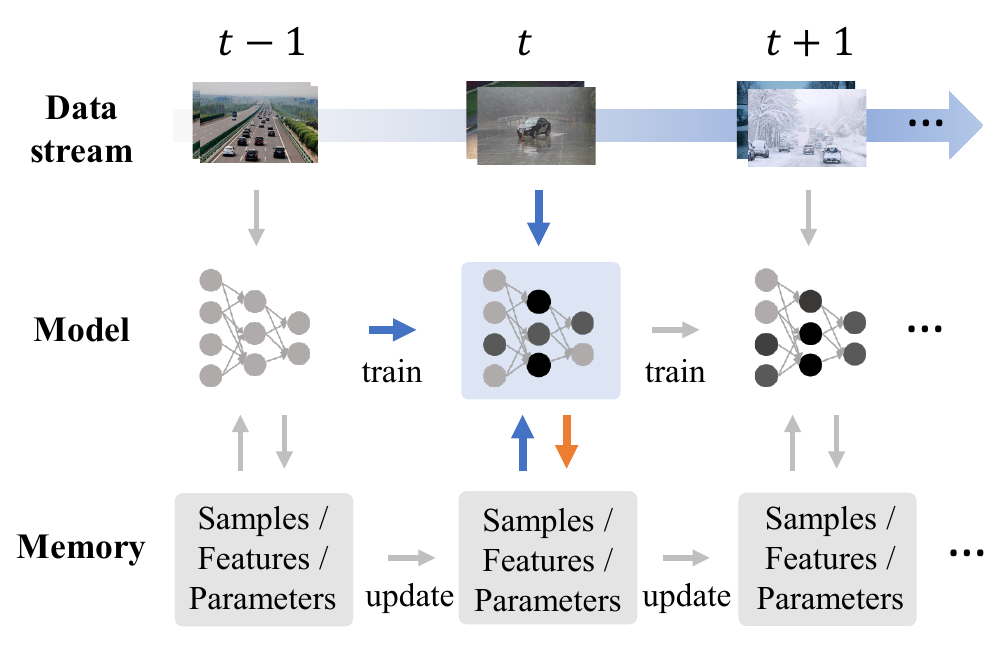} 
\caption{The framework of Continual Learning. For example, for time $t$, combing the model and memory at time $t-1$, and data at time $t$ together to obtain the model and memory at time $t$.}
\label{Fig:4AIoT_C_3}
\end{figure}

Curriculum learning (mentioned in Section \ref{chapter:4.3.2}) and other adaption methods allow models to adapt to specific tasks. However, These methods ignore the learning of old knowledge when adapting to new tasks, leading to catastrophic forgetting of knowledge \cite{french1999catastrophic}. When facing the challenge of incremental tasks, continual learning methods maintain the knowledge of old tasks during the learning process and avoid forgetting that knowledge when learning new tasks, and enable knowledge to accumulate and continue over time, as shown in Fig. \ref{Fig:4AIoT_C_3}. So, continual learning methods also belong to the evolution mode. Based on the form of preserved knowledge, continual learning methods can be classified into three categories: \textbf{reply-based continual learning}, \textbf{regularization-based continual learning}, and \textbf{parameter isolation-based continual learning}.

Reply-based continual learning methods typically store the original samples from previous tasks or generate pseudo-samples, and then add the stored samples to the training dataset to prevent forgetting when learning new tasks. When training a new task, the subset of stored samples undergo explicit retrained. For example, iCaRL \cite{rebuffi2017icarl} is an incremental learner that stores a subset of samples for each class. During model training, it is assumed that a fixed memory is allocated, and samples closest to the mean vector of each class’s characteristics are selected and stored in memory. While in the test stage, iCaRL calculates the class mean based on all samples and uses the Nearest-Mean-of-Examplars algorithm for classification tasks. Nevertheless, there may be a tendency to overfit the stored historical samples, and the scalability of the number of classes/tasks is limited due to the need for additional computing resources and storage space for raw inputs, which is impractical on resource-constrained terminal devices. GEM \cite{lopez2017gradient} utilizes a first-order Taylor series approximation to project the estimated gradient direction to the feasible region of the gradient from previous tasks, which could constrain the update of the new tasks without interfering with previous tasks. Similarly, A-GEM \cite{chaudhry2018efficient} maps the estimated gradient direction to a specific direction using a first-order Taylor series approximation. This method could be extended to an online continual learning paradigm without task boundaries, further improving the efficiency of model updating.

Regularization-based continual learning methods consolidate knowledge from historical tasks when learning new tasks through integrating extra regularization terms into the loss function of neural networks (the intensity of regularization is proportional to the level of knowledge retention), successfully avoiding the need to store raw input samples and prioritizing privacy protection. For example, LwF \cite{li2017learning} preserves knowledge from previous tasks through distillation losses, and then utilizes the outputs of previous models as soft labels to learn new tasks. EBLL \cite{rannen2017encoder} expands LwF by retaining the important low-dimensional features. In addition, some studies begin by estimating the prior parameter distribution of the latest model based on parameters from historical models, and utilize this prior parameter distribution to guide model training for new tasks. Note that during the training process, when sharp changes are made to important parameters, penalties are imposed on the optimization of the model. For instance, EWC \cite{kirkpatrick2017overcoming} incorporates the uncertainty of network parameters into the Bayesian framework, and calculates a prior distribution based on the Bayesian estimation of previous tasks (the posterior of the previous task constitutes the prior of the new task). The IMM \cite{lee2017overcoming} can estimate the Gaussian posterior probability of task parameters, and this probability’s mixed distribution is approximated by each single Gaussian distribution.

Parameter isolation-based continual learning methods specify distinct network parameters for each task to prevent any potential forgetting. Implementation options encompass growing new branches, freezing parameters of previous tasks, or allocating a separate model copy for each task. For instance, PackNet \cite{mallya2018packnet} sequentially allocates a subset of parameters to tasks using a binary mask. It trains DNNs without modifying parameters from previous tasks and prunes insignificant parameters that are deemed unnecessary. Afterward, PackNet retrains the pruned subset of important parameters to ensure continuous learning capabilities. For example, Serra et al. \cite{serra2018overcoming} introduce a hard attention mechanism, namely HAT, which utilizes stochastic gradient descent to learn a binary mask for each task. When the binary mask in HAT equals 0, it prevents updates to network weights associated with the current task, keeping them unchanged. However, when the mask equals 1, HAT facilitates the adaptation of the network to the new task by allowing optimization of the corresponding weights.

Continual learning methods have been widely applied in fields such as human activity recognition, urban computing, connected vehicles, multi-robot systems, etc. For example, in the field of urban computing, with the continuous arrival of urban data and the constant changes in urban scenes, existing methods mainly focus on solving static data, making it difficult to model the continuous evolution of urban data. Therefore, continuous learning aims to help urban computing models quickly learn new knowledge without forgetting old knowledge, thereby improving the generalization, adaptability, and accuracy of urban computing models. For example, PECPM \cite{wang2023pattern} is an efficient and effective continuous learning framework for achieving continuous traffic flow prediction without accessing historical data. Specifically, a pattern bank based on pattern matching is first proposed to store representative patterns of the traffic flow network. Then a pattern extension mechanism is applied to detect evolving patterns and new patterns, extending the new patterns to the pattern bank to adapt to the new traffic flow network. Finally, a method for integrating pattern preservation mechanism and pattern traceability mechanism is proposed, which integrates new knowledge and consolidates the old knowledge to achieve more accurate continuous traffic flow prediction.

\subsubsection{Test-time Adaptation} \label{chapter:4.3.4}
\begin{figure}[!t]
\centering
\includegraphics[width=0.466\textwidth,height=4.3cm]{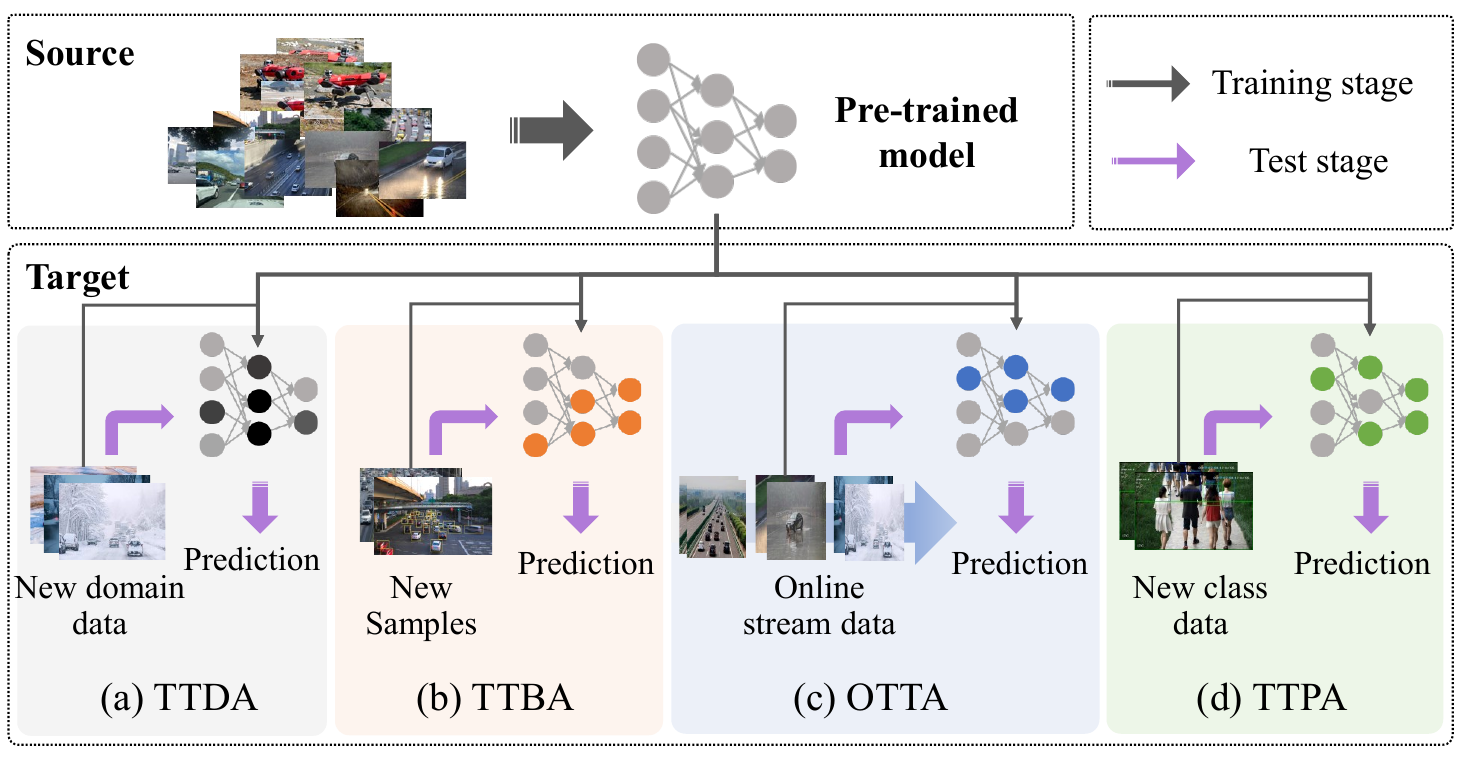} 
\caption{The framework of Test-time Adaption.}
\label{Fig:4AIoT_C_4}
\vspace{-3mm}
\end{figure}

In AIoT scenarios, the differences in capturing devices, scenes, and environments lead to inconsistent distributions between target data and training data. Domain adaptation, domain generalization, and other methods provide promising solutions. However, these methods require labeling for the target data. In the real world, the labels of test data are unknown, and test data is real-time. So, to address these challenges, the recent test-time adaption (TTA) method has emerged. It aims to adapt the source model during the testing phase to accommodate specific test data and enhance the model's performance on such data, belonging to the evolution mode. Depending on the type of test data, TTA method can be categorized into four groups: \textbf{test-time domain adaptation(TTDA)}, \textbf{test-time batch adaptation(TTBA)}, \textbf{online test-time adaptation(OTTA)}, and \textbf{test-time prior adaptation(TTPA)} (as shown in Fig. \ref{Fig:4AIoT_C_4}).

TTDA, also known as source-free domain adaptation, aims to utilize the unlabeled data in the target domain to adapt the trained model on the source domain to the data distribution in the target domain, to improve the performance on the target domain. The core idea is to learn some auxiliary tasks or feature representations from the unlabeled data in the target domain by means of self-supervised learning or self-generated models, etc., to reduce the distributional differences between the source and target domains. For example, SHOT \cite{liang2020we} freezes the source classifier and fine-tunes the target-specific feature encoding module using information maximization and self-supervised learning methods to produce a target data representation that is identical to the source data representation. Further, SHOT++ \cite{liang2021source} selects low-entropy instances in each class to perform intra-domain alignment based on the confidence of the prediction result, making the target data distribution closer to the source data distribution.
Considering that differently distributed data may be a single instance or instances, the domain distribution alignment approach is difficult to apply, and thus TTBA is proposed to make the off-the-shelf model adapt to individual instances. For example, 
TTT \cite{sun2020test} transforms an individual unlabeled test sample into a self-supervised learning problem that updates the model's parameters before making predictions, improving the robustness of the model to distribution bias. And further, Tent \cite{wang2020tent} requires no source data and makes the model fit the test data by minimizing the entropy of the source model on the test data.
All these offline test-time adaptation methods require some instances from some batches or domains, which may not be feasible for real-time streaming data scenarios. The TTT-online method sets the batch size to 1 to address this issue. However, instances may come from different distributions during testing, which can easily lead to erroneous accumulation and catastrophic forgetting issues. Therefore, the OTTA method has been proposed to tackle the aforementioned challenges. For example, CoTTA \cite{wang2022continual} proposes to use accurate weight averaging to reduce the accumulation of errors, and restore a small fraction of neurons to the weights of the pre-trained network during each training iteration to preserve the long-term knowledge from the source model and prevent catastrophic forgetting. EATA \cite{niu2022efficient} utilizes entropy to filter predictively plausible samples to improve model updating efficiency, and uses the EWC method to weight and regularize parameter importance to prevent catastrophic forgetting. And EcoTTA \cite{song2023ecotta} introduces a lightweight meta-networks that adapt the pre-trained network to the target domain, and then control the result of the meta-network to be close to the output of the pre-trained model by using the self-distillation regularization method to preserve the knowledge in the pre-trained model.
The aforementioned three approaches are applicable to the problem of test data distribution bias, however, in real AIoT scenarios, a large amount of data with different labels arrives. TTPA is proposed to solve this problem. For example, TTLSA \cite{sun2022beyond} finds that the generative model of the features is invariant, and thus it trains a well-calibrated classifier for a test-time label shift using logit adjustment. And SADE \cite{zhang2022self} first trains multiple experts from a long-tailed dataset to deal with different class distributions, and then utilizes a self-supervised approach to aggregate the learned knowledge from multiple experts to deal with unknown test class distributions.

TTA methods have been widely applied in fields such as human activity recognition, urban computing, connected vehicles, healthcare, etc. For example, in the field of connected vehicles, autonomous vehicles usually need to dynamically model the surrounding environment during driving, to facilitate real-time environmental understanding and decision-making planning \cite{chen2023end}. However, the environment is complex, uncertain, and constantly changing, making it difficult for existing models to make real-time adaptive adjustments based on the current environment, which limits the safety and accuracy of autonomous driving. Therefore, in the field of connected vehicles, TTA methods are applied to train the adaptive model with real-time unlabeled data before inferencing, to improve the robustness of the model and enhance the safety, accuracy and adaptability of autonomous vehicles. For example, multiple object tracking, i.e., detecting objects in individual frames of a video and studying the association between individual frames over time, thus multiple target tracking consists of two parts, target detection and data association, which are tightly coupled with each other. Darth \cite{segu2023darth} is a holistic test-time adaptive framework to solve the domain bias problem of multiple object tracking. Specifically, a patch contrastive learning method is firstly proposed to adapt to the data association stage and learn discriminative appearance representations in the target domain, achieving self-matching of detected pose appearances under different enhancement views for the same image. Then, a consistency detection method is proposed, which utilizes a self-supervised approach to adapt object modeling to the target domain and enhance the model's robustness to photometric changes.



\begin{table*}[!t]
\centering
\caption{A Summary of main Applications.}
\label{tab:apps}
\renewcommand\arraystretch{1.1}{
\resizebox{\textwidth}{!}{
\begin{tabular}{m{2.0cm}<{\centering}|m{3.45cm}<{\centering}|m{4.4cm}<{\centering}|m{5.2cm}<{\centering}|c}
\hline
Application Area    & Specific Task   &  Related Work    & Transfer Category    & Application Scenario                                                                    \\ \hline
\multirow{8}{*}{\shortstack{Human Activity \\ Recognition}} & \multirow{3}{*}{Gesture recognition}  & CrossSense~\cite{zhang2018crosssense}    & Intra-agent knowledge transfer      & Gait identification and gesture recognition    \\
&  & MobileDA \cite{yang2020mobileda}     & Centralized inter-agent knowledge transfer      & WiFi gesture recognition                      \\
&  & Widar3.0 \cite{zhang2021widar3}     & Intra-agent knowledge transfer           & Cross-domain gesture recognition               \\
\cline{2-5} & \multirow{4}{*}{Daily activity recognition}
& USSAR \& TNNAR~\cite{wang2018deep} & Intra-agent knowledge transfer                             & Daily activity recognition              \\
&  & STL~\cite{wang2018stratified}           & Intra-agent knowledge transfer           & Daily activity recognition              \\
&  & GILE \cite{qian2021latent}         & Centralized inter-agent knowledge transfer     & Cross-person sensor-based human activity recognition                   \\
&  & Meta-HAR  \cite{li2021meta}    & Centralized inter-agent knowledge transfer   & Daily activity recognition \\
\cline{2-5} & \multirow{1}{*}{Healthcare}   & FedHealth \cite{chen2020fedhealth}    & Centralized inter-agent knowledge transfer                             & Wearable activity recognition            \\ \hline

\multirow{9}{*}{\shortstack{Urban \\ Computing}} & \multirow{3}{*}{Traffic} & RegionTrans \cite{wang2018cross}  & Decentralized inter-agent knowledge  & Pedestrian Flow Prediction                    \\
&
& STMetaNet \cite{pan2019urban}; Meta-MSNet \cite{fang2020meta}  & Centralized inter-agent knowledge transfer  & Traffic Flow Prediction                    \\
&
& cST-ML \cite{zhang2020cst}; Dac-ml \cite{zhang2021dac}  & Centralized inter-agent knowledge transfer; Intra-agent knowledge transfer    & Traffic Flow Prediction                    \\

\cline{2-5} & \multirow{2}{*}{Environment}
& TL-ResNet \cite{tariq2021transfer} & Decentralized inter-agent knowledge   & Air Quality Prediction                    \\
&
& Wu et al. \cite{wu2023meta}  & Centralized inter-agent knowledge transfer; Intra-agent knowledge transfer    & Air Pollution  Prediction                    \\
\cline{2-5} & \multirow{2}{*}{Security}
& DeepFire \cite{khan2022deepfire} & Centralized inter-agent knowledge transfer   & Fire Detection                    \\
&
& Tasnim et al. \cite{tasnim2022novel} & Intra-agent knowledge transfer  & Crime Event Prediction                    \\
\cline{2-5} & \multirow{2}{*}{Service}
& Citytransfer \cite{guo2018citytransfer}; DeepStore \cite{liu2019deepstore} & Intra-agent knowledge transfer  & Store Site Recommendation                    \\
& 
& Axolotl \cite{gupta2022doing}; MERec \cite{wang2023meta}; & Centralized inter-agent knowledge transfer  & POI Recommendation                    \\
\hline

\multirow{6}{*}{\shortstack{Connected \\ Vehicles}} & \multirow{3}{*}{Object detection} & Neupane et al. \cite{neupane2022real} & Centralized inter-agent knowledge transfer   & Classification and tracking \\
&   & Abdulateef et al. \cite{abdulateef2022vehicle}  & Centralized inter-agent knowledge transfer 
 & Pinpoint the location of license plates and license numbers \\
&    & Rajathi et al. \cite{rajathi2022cnn}    & Centralized inter-agent knowledge transfer   & Classification of vehicles                         \\  
\cline{2-5} & \multirow{3}{*}{Path planning}  & ATN \cite{chen2019learning} & Intra-agent knowledge transfer & On-road navigation\\
&   & Shu et al. \cite{shu2021driving} & Decentralized inter-agent knowledge transfer  & Real-time decision-making for autonomous vehicles  \\
&   & Li et al. \cite{li2022driver}   & Intra-agent knowledge transfer   & Planning appropriate driving trajectory for route following      \\                      
\hline

\multirow{6}{*}{\shortstack{Multi-Robot \\ Systems}} & \multirow{3}{*}{Mobile robot navigation}  & Shuhuan et al. \cite{wen2021multi}        &  Decentralized inter-agent knowledge transfer      &  Path planning   \\  
&       & Xianjia et al. \cite{yu2022towards}               & Centralized inter-agent knowledge transfer         &  Vision-based autonomous navigation      \\
&      & Thomas et al. \cite{chaffre2020sim}          &  Intra-agent knowledge transfer         & Depth-based robot navigation  \\
\cline{2-5} & \multirow{3}{*}{UAV collaboration}  & Hongming et al. \cite{zhang2020federated}           & Centralized inter-agent knowledge transfer   & UAV-assisted exploration                                                                                                   \\
&   & Yi et al. \cite{liu2020federated}                & Centralized inter-agent knowledge transfer &  UAV-based air quality sensing                                                                                                     \\
&   & Lin et al. \cite{zhang2022forest}               & Centralized inter-agent knowledge transfer    & UAV-based forest fire recognition                                                                                                          \\                                               \hline
 \multirow{6}{*}{Smart Factory} &   \multirow{2}{*}{Manufacturing defect detection}         & Jiahui et al. \cite{cheng2021tl}              & Centralized inter-agent knowledge transfer   & Few-shot surface defect detection                            \\
&   & Xian et al. \cite{lee2023xdnet} &  Decentralized inter-agent knowledge transfer           & Cross-domain visual inspection                               \\
\cline{2-5} & \multirow{2}{*}{Machinery fault diagnosis} &   Yao et al. \cite{hu2022few}          & Centralized inter-agent knowledge transfer           & Fault diagnosis of bearings            \\
&   & Jun et al. \cite{zhao2023adaptive}         & Decentralized inter-agent knowledge transfer         & Cross-domain fault diagnosis  \\
\cline{2-5} &\multirow{2}{*}{\makecell{Anomaly detection of \\ industrial processes}}  &   Jeongyong et al. \cite{park2023mendel}         & Centralized inter-agent knowledge transfer  & Time series anomaly detection                                     \\
&   &  Wentao et al. \cite{mao2023deep}         & Decentralized inter-agent knowledge transfer     & One-class anomaly detection      \\ 
\hline
\end{tabular}}
}
\end{table*}

\section{Applications}
In this section, we showcase various AIoT applications to demonstrate the effective utilization of CrowdTransfer methods in enhancing model performance within real-world scenarios, as shown in \ref{tab:apps}.

\subsection{Human Activity Recognition}

Human activity recognition (HAR) refers to the task of inferring and predicting human intentions and activities using sensor readings. Recent advancements in HAR have led to numerous applications, including smart homes \cite{ishimaru2017towards}, healthcare \cite{li2016deep}, and sleep state detection \cite{zhao2017learning}, among others. HAR plays a vital role in human life as it captures and analyzes people's behavior, enabling computing systems to monitor and assist individuals in their daily activities. Owing to the intricate relationship between sensor readings and activities, machine learning algorithms have gained popularity as a viable approach for activity recognition tasks. Given that the human body is commonly equipped with multiple sensors (e.g., smartwatches, mobile phones, etc.), researchers often incorporate the concept of knowledge transfer in HAR, contrasting with traditional single-source single-target transfer learning solutions, as discussed subsequently.

\subsubsection{Gesture recognition}
Zhang et al. \cite{zhang2018crosssense} address the issue of domain shift by employing a mixed-expert approach. They utilize multiple specialized perception models or the collective intelligence of experts to capture the mapping from various WiFi inputs to desired outputs. This approach expands WiFi sensing to new environments and increases the problem size, with applications in gait recognition and gesture recognition. Yang et al. \cite{yang2020mobileda} propose a cross-domain knowledge distillation method, namely MobileDA. In MobileDA, a teacher network trained on a server extracts knowledge for a student network operating on an edge device, all while maintaining the simplicity of the deep model. It aims to learn transferable features and is applied to WiFi gesture recognition tasks. Zhang et al. \cite{zhang2021widar3} investigate cross-domain knowledge transfer and introduce the Widar3.0 system, which applies cross-domain knowledge transfer to a WiFi-based gesture recognition system. Widar3.0 extracts domain-independent features of human gestures at lower signal levels, capturing their unique dynamics and making them applicable across domains. Based on this, a universal general-purpose model is developed with a single training, allowing it to adapt to different data domains and achieve zero-effort cross-domain recognition. 

\subsubsection{Daily activity recognition}
Wang et al. \cite{wang2018deep} explore the scenario in which the activity information is missing on the arm and investigate how to utilize information from other body parts (referred to as the source domain, e.g., the torso or legs) to enhance model construction. They propose the unsupervised HAR algorithm, namely USSAR, which efficiently selects the $K$ most similar source domains from a given list. Additionally, they introduce a transfer neural network, dubbed TNNAR, to facilitate knowledge transfer in HAR. TNNAR can capture both temporal and spatial relationships between activities during knowledge transfer. Qian et al. \cite{qian2021latent} address the issue of DG by proposing a HAR method known as GILE. It is designed to automatically disentangle domain-independent and domain-specific features, with the former expected to remain unchanged across individuals. Moreover, a novel independent incentive mechanism is introduced in the latent feature space to further eliminate the correlation between these two types of features. Importantly, the model can be directly applied to diverse target domains without the need for retraining or fine-tuning. Consequently, the cross-population generalization ability of the model is greatly enhanced. In addition, Li et al. \cite{li2021meta} emphasize that each user may have different activity types, and even the same activity type can exhibit varying signal distributions. To improve the generalization of the global model to new users with heterogeneous data, a Meta-HAR framework is proposed. This framework learns the signal embedding network in a federated manner and additionally feeds the learned signal representation into personalized classification networks for each user, enabling activity prediction.

\subsubsection{Healthcare}
Chen et al. \cite{chen2020fedhealth} introduce FedHealth, a FTL framework for healthcare applications in wearable devices. This framework utilizes federated learning to aggregate data and employs transfer learning to construct personalized wearable models. 
Specifically, FedHealth offers precise personalized healthcare services while safeguarding the privacy of data subjects. Initially, a cloud model is trained on a public dataset on the server-side. Subsequently, the cloud model is distributed to all users, enabling each user to train their own models on local data. Then, the user models are uploaded to the cloud server to assist with training updates. During the upload process, only homomorphically encrypted model parameters are shared, without revealing any user data or information. Finally, each user can generate personalized wearable healthcare models by integrating the cloud model and their local model for personalized training. Ju et al. \cite{ju2020federated} apply FTL to brain-computer interface research, proposing a framework for EEG classification. Compared to the current state-of-the-art deep learning frameworks, this framework can extract more effective discriminative information from multi-subject EEG data.

\subsection{Urban Computing}

\begin{figure}[h]
\centering
\includegraphics[width=0.495\textwidth,height=6.8cm]{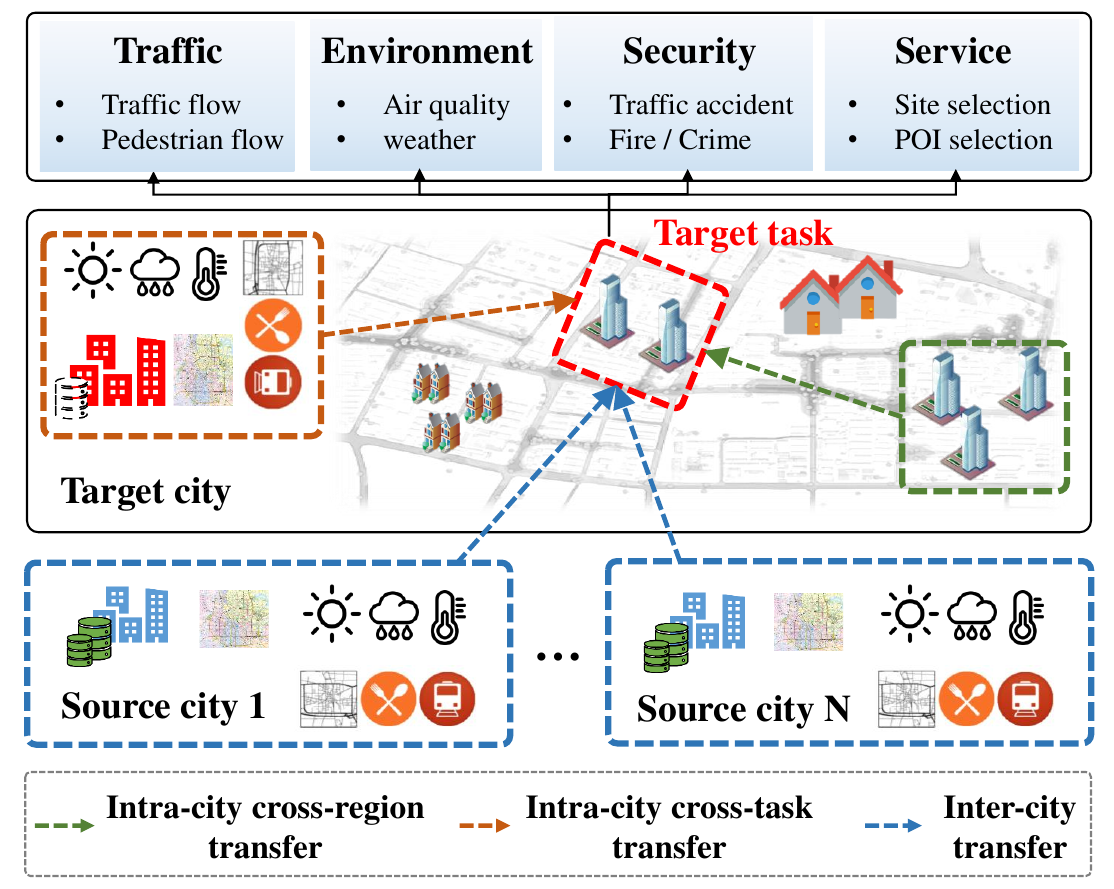} 
\caption{The knowledge transfer in urban computing.}
\label{Fig:5Application_2}
\vspace{-3mm}
\end{figure}

According to the definition of Zheng et al. \cite{zheng2014urban}, urban computing is the process of acquiring, integrating and analyzing massive heterogeneous data generated by sensors, equipment, vehicles, buildings and human beings in the urban space to solve the main problems faced by the city (such as air pollution, increased energy consumption and traffic congestion). Today, smartphones, vehicles, and infrastructure (e.g., traffic cameras, air quality monitoring stations) continuously generate large amounts of urban data in heterogeneous formats, such as GPS points, social media check-ins, and road traffic. Therefore, in scenarios where there is a limited amount of data available for the target task in the target city, it is possible to leverage data from other cities, regions, or tasks to assist in learning the target task in the target city, reducing training time and resource consumption for the target task model, and improving the model's prediction and generalization performance. Specifically, as shown in Fig. \ref{Fig:5Application_2}, the target task in the target city can be transferred from rich source city data (blue dashed lines), from rich similar regions within the target city (green dashed lines), or from other tasks within the target city (orange dashed lines). Through these transfer learning methods, the target tasks in the target city with scarce data can be effectively addressed. As the content of transfer varies in different application scenarios, this section categorizes urban computing scenarios into four types: \textbf{Traffic}, \textbf{Environment}, \textbf{Security}, and \textbf{Service}, and provides specific introductions for each of them.

\subsubsection{Traffic}

Traffic flow prediction in cities is an important problem in intelligent transportation, which aims to accurately predict the pedestrian flow, vehicle flow, and other information in different areas in cities. This enables better flow control, congestion management, and ensures urban public safety. For example, Wang et al. \cite{wang2018cross} model the similarity between regions in the source city and the target city, facilitating the selection of similar regions in the source city to assist in predicting the pedestrian flow in the target city. Due to the complex spatio-temporal correlations in urban traffic, such as spatial correlations based on spatial locations and temporal correlations based on time series, which may vary with time and location. So Pan et al. \cite{pan2019urban} propose a deep meta-learning model to capture the complex and diverse spatio-temporal correlations for predicting traffic flow in all areas in the city. Fang et al. \cite{fang2020meta} design a combination module of two meta-learning models to incorporate diverse origins of external data in both temporal and spatial aspects. Furthermore, traffic flow is highly dynamic, varying with geographical locations and time, and is easily influenced by external factors. Therefore, \cite{zhang2020cst} and \cite{zhang2021dac} adopt a Bayesian meta-learning method to learn a general traffic dynamics prediction strategy that can rapidly adjust to prediction tasks with only a few samples based on historical traffic data.

\subsubsection{Environment}
Environment is an important issue for public all over the world, which aims to accurately predict information such as air quality, water quality, and weather in different areas in cities, facilitating better environment monitoring between regions, helping in developing alert systems, and enabling timely urban decision-making. For instance, Wei et al. \cite{wei2016transfer} introduce a flexible multi-modal transfer learning method that transfers information-rich multi-modal data and labels from a source city to a target city with limited information. This alleviates the problem of label scarcity and data insufficiency in the target city, improving the accuracy of air quality prediction. Tariq et al. \cite{tariq2021transfer} mine the similarities between different subway station environments, so they introduce knowledge from data-rich subway platforms to assist in predicting the PM2.5 levels in stations with scarce data. Ma et al. \cite{ma2020transfer} argue that similar sequences exhibit the same patterns. Accordingly, they propose a transferred LSTM-based iterative estimation model that utilizes the knowledge from similar sequences to impute consecutive missing values with a higher missing rate, resulting in more accurate predictions of PM2.5 concentrations in the air. Similarly, Chen et al. \cite{chen2021transfer} first use dynamic time warping method to calculate the similarity between data from different water quality monitoring stations. And then they transfer sufficient and long-term data from similar water quality monitoring stations to the target monitoring station with missing data, facilitating more accurate predictions of the water quality in the target station. However, the availability of source data may not be sufficient. Therefore, Wu et al. \cite{wu2023meta} propose a meta-learning spatial-temporal adaptive method that captures different data patterns, providing dynamic adaptive predictions for air pollution problems across time and space.

\subsubsection{Security}
Security is an important issue for public in cities, impacting people's lives. City security aims to accurately predict information such as traffic accidents, fires, and crimes in different areas in cities, enabling better regional public safety control, facilitating real-time alerts and decision-making. For instance, Khan et al. \cite{khan2022deepfire} introduce a drone-based forest fire detection system that utilizes fine-tuned transfer learning methods to achieve more accurate image classification, thereby enhancing forest fire monitoring and enabling timely warnings. The occurrence of traffic accidents is influenced by complex dependencies between spatial and temporal features. Therefore, An et al. \cite{an2022hintnet} introduce a hierarchical knowledge transfer method called HintNet. It divides regions into multiple hierarchical sub-regions based on accident risks and models spatial heterogeneity and temporal sparsity separately. This approach better captures irregular heterogeneous patterns and sparse patterns, leading to more accurate predictions for urban traffic accidents. Crime prediction is a complex problem, as different factors can contribute to different types of criminal incidents. Therefore, Tasnim et al. \cite{tasnim2022novel} propose an effective multi-module learning method to predict crime events in cities. It first combines temporal-based attention LSTM and spatio-temporal based stacked bidirectional LSTM for feature-level fusion. Then, the fused features are further combined with spatio-temporal based attention-LSTM and stacked bidirectional LSTM to obtain the final prediction results, enabling more accurate crime event predictions in cities.

\subsubsection{Service}

Urban services are crucial for improving people's quality of life and promoting the sustainable development of cities. For example, point of interest (POI) recommendations can provide personalized location suggestions based on people's interests and needs. Tourism recommendations can suggest travel routes based on factors such as reviews and historical travel records. Through these services, people can better adapt to urban life, enjoy a higher quality of life, contributing to the sustainable development of cities.
For example, Guo et al. \cite{guo2018citytransfer} explore the correlations between different entities within a city and between entities in different cities, and accordingly, propose to leverage the data from other entities within the city and data from other cities to enhance the performance of chain store site recommendations in the target city.
Liu et al. \cite{liu2019deepstore} introduce a feature-level multi-modal learning approach utilizing multi-source data such as store data, neighborhood user data, and POI data, to model complex feature interactions and provide more accurate store site recommendations.
When a source city has insufficient data, Metastore \cite{liu2021metastore} leverages meta-learning to transfer knowledge from various source cities to the target city, enabling optimal store site recommendations in the target city. 
Similarly, Gupta et al. \cite{gupta2022doing} utilize the meta-learning method to transfer knowledge from various regions within a city to the target region, achieving accurate POI recommendations. Wang et al. \cite{wang2023meta} employ the meta-learning method to transfer knowledge from multiple cities to the target city, capturing shared patterns across cities to transfer more pertinent knowledge for precise POI recommendations. 
However, due to data sparsity and pattern diversity among different users in multiple cities, Chen et al. \cite{chen2021curriculum} not only consider the differences at the city level but also take into account the differences at the user level. They employ a curriculum learning approach to assist in improving the performance of meta-learning methods for POI recommendations.

\subsection{Connected Vehicles}
The fundamental source of economic growth in any country relies on well-planned and resilient transportation systems based on spatial information. In any case, most cities in the world still face the rapid growth of traffic flow and the complexity of traffic management, resulting in a low quality of life in modern cities. However, with advances in internet bandwidth and machine learning in recent years, autonomous vehicles (AVs) are poised to improve, reshape and revolutionize future ground transportation. It is expected that one day ordinary cars will be replaced by smart cars that can make decisions and perform driving tasks autonomously.

\begin{figure}[!b]
\centering
\includegraphics[width=0.499\textwidth,height=5.7cm]{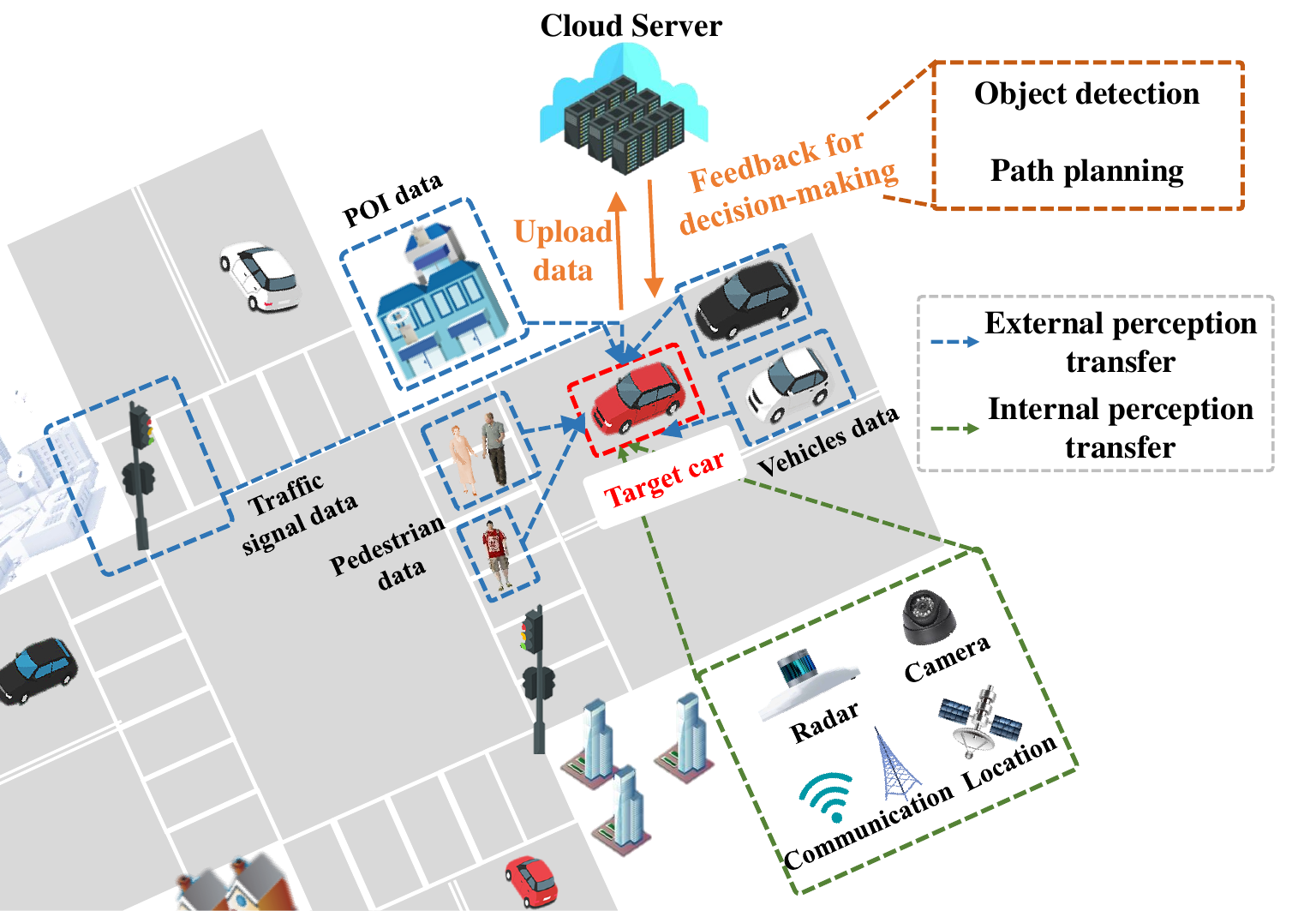} 
\caption{The knowledge transfer in connected vehicles.}
\label{Fig:5Application_3}
\end{figure}

More and more deep learning-based autonomous driving solutions have been proposed. However, due to political constraints in different places, vehicles usually cannot obtain a large amount of training data in every region or scenario. In addition, connected vehicles typically operate in different environments and scenarios, such as city roads, highways, and more, resulting in performance differences for connected vehicles across different domains. And, above all, real-time capability is an important consideration in connected vehicles. Traditional approaches necessitate a substantial time investment and computational resources to train models, which is impractical for real-time applications. To address these challenges, transfer learning approaches have been introduced to assist in improving the performance and adaptability of connected vehicles while enhancing their time efficiency, as shown in Fig. \ref{Fig:5Application_3}. 

\subsubsection{Object detection}
Object detection in connected vehicles aims to identify and locate specific objects, such as vehicles, pedestrians, and traffic signals, from complex scenes. This allows for timely perception of the vehicle's surroundings, ensuring appropriate decision-making and actions to enhance vehicle safety. For example, Neupane et al. \cite{neupane2022real} propose a real-time detection of vehicles using sensing technology in order to realize the automation of public safety positioning intelligence in road environments in intelligent transportation systems. A base model with a larger knowledge base is first trained on a hand-prepared vehicle dataset. The base model is further fine-tuned based on the camera-prepared small dataset. The two-step training procedure allows the model to leverage learning from cross-domain datasets. Furthermore, this process reduces the number of training samples and iterations, enabling the model to be immediately implemented in real-world systems and minimizing the impact of domain shift between cross-domain datasets. Then, the fine-tuned model is combined with a multi-vehicle tracking algorithm to obtain the lane count, classification and speed calculation of vehicles in real time. Similarity,
Abdulateef et al. \cite{abdulateef2022vehicle} introduce a deep convolutional neural network pre-trained based on data from vehicle license plate, to automatically achieve license plate localization and vehicle quantity estimation. And Rajathi et al. \cite{rajathi2022cnn} introduce the pre-trained neural networks to capture the rich representation of images in connected vehicles, facilitating more accurate and real-time classification and detection in connected vehicles.

\subsubsection{Path planning}
Path planning in connected vehicles aims to consider multiple factors such as road traffic conditions, road types, speed limits, traffic signals, etc., to find the optimal driving path. This can improve driving efficiency, reduce energy consumption and emissions, and enhance vehicle safety. 
For example, Chen et al. \cite{chen2019learning} propose a auxiliary task network method to learn on-road driving for self-driving vehicles, which uses the image segmentation task inspired by human experience and also additionally references optical flow information and vehicle kinematics features to accelerate model training and provide more accurate navigation. In addition to multi-modal learning methods, transfer between similar tasks can also be used to improve real-time decision-making performance of vehicles. For example, Shu et al. \cite{shu2021driving} introduce a transfer deep reinforcement learning approach to conduct real-time decision-making for autonomous vehicles in intersection environments. The decision at the unsignalized intersection includes left turn, right turn, and running straight, so this paper transfers the three learned decision strategies from one driving task to another similar driving task, aiming to reduce time consumption and achieve real-time decision-making for autonomous driving vehicles. However, the deep reinforcement learning approach requires millions of trial-and-error iterations, consuming a significant amount of time. 
So, the behavioral cloning-based imitation learning method attracts a lot of attention. For example, Li et al. \cite{li2022driver} leverage the task knowledge distillation method for automatic trajectory learning in autonomous vehicles, which transfers the driving strategies achieved from human driving behavior to domain shift scenarios.
To reduce the impact of negative transfer learning, \cite{li2022reducing} introduces a clustering-based transfer learning approach to address the dynamic multi-objective optimization problems for path planning of unmanned air/ground vehicles.

\subsection{Multi-Robot Systems}

In recent years, the rapid advancement of both robot technology and communication technology has ignited significant interest in multi-robot systems. 
These systems, as opposed to single-robot solutions, offer the potential to tackle more intricate tasks through collaborative efforts, such as conducting search and rescue missions in disaster-stricken environments \cite{brambilla2013swarm}.
One notable development in this field has been the application of reinforcement learning methods, which enable robots to acquire skills through iterative learning from trial and error. However, these approaches encounter several pressing challenges.
Firstly, acquiring data from real-world environments proves to be prohibitively expensive for robots. Consequently, the majority of multi-agent systems resort to training within simulated environments. Nevertheless, disparities between real and simulated data distributions pose significant obstacles, rendering direct utilization of pre-trained models ineffective.
Secondly, real-world environments are marked by their complexity and dynamic nature. Developing a complex multi-agent system from scratch for each new environment incurs substantial costs. 
As a result, it is necessary to leverage the knowledge transfer learning methods in multi-agent systems to enhance and improve the robustness and generalization of robots.

\subsubsection{Mobile robot navigation}
One important application of multi-robot collaboration is path planning and navigation tasks.  
For instance, multiple robots can collaborate to efficiently achieve navigation objectives in a new environment to boost overall system efficiency. 
However, the environments encountered in navigation tasks are different, and it is a major challenge to improve the system's generalization across different settings. 
To tackle this challenge, Shuhuan et al. \cite{wen2021multi} propose a dynamic proximal meta policy optimization approach, which enables robots to quickly learn the strategies of obstacle avoidance for autonomous navigation. 
During training, dynamic proximal policy optimization is employed to learn strategies from various tasks. 
Subsequently, in the testing phase, the learned meta-policy is transferred to new environments as initial parameters.
In real-world scenarios, mobile robots need not only to adapt rapidly to new environments but also to remember the learned knowledge of old environments to enhance their navigation capabilities across different scenarios. 
For example, Bo et al. \cite{liu2021lifelong} introduce a lifelong learning approach for robot navigation. It begins by identifying suboptimal actions based on initial state recognition and continues to improve navigation performance through real-time data collection. 
Considering the data privacy in inter-robot communication, Boyi et al. \cite{liu2019lifelong} further propose a lifelong federated reinforcement learning model, which fuses and transfers the knowledge learned from multiple robots. 
Experimental results indicate that these methods significantly enhance the efficiency of reinforcement learning for robot navigation.
Models trained in simulated environments often struggle to perform well in real-world settings due to distribution differences between real data and simulated data. 
To address this issue, Thomas et al. \cite{chaffre2020sim} present a soft-actor critic training strategy for depth-based mapless navigation. This strategy improves policy generalization through domain randomization methods, reducing the need for extensive model fine-tuning in real environments. 
Considering the interactions between different robots, Xianjia et al. \cite{yu2022towards} leverage federated learning to facilitate distributed collaboration among robots in obstacle avoidance tasks.

\subsubsection{UAV collaboration}

Unmanned aerial vehicles (UAVs) offer substantial potential in various applications. However, due to their inherent limitations in payload capacity and flight duration, many complex tasks necessitate the coordinated efforts of multiple UAVs, such as rescue operations and target surveillance. In UAV systems, achieving effective collaboration among UAV swarms presents challenges, as increased communication can lead to delays and higher computing costs.
To tackle these challenges, researchers have explored innovative solutions centered around knowledge transfer.
Hongming et al. \cite{zhang2020federated} employ federated learning to optimize UAV operations in UAV-assisted exploration scenarios. This approach reduces communication costs between UAVs and ground control centers while simultaneously enhancing the performance of image classification tasks. Specifically, each UAV initially trains a local model based on collected images and then uploads this model to the ground to construct a global model.
Additionally, Hamid et al. \cite{shiri2020communication} leverage the federated learning method to facilitate parameter exchange between UAVs at regular intervals, ultimately improving the overall system performance. 
Similarly, Yi et al. \cite{liu2020federated} propose a federated learning-based framework for air quality sensing, which not only enhances detection capabilities but also expands the range of air quality monitoring conducted by UAVs in intricate 3-D environments. 
Beyond the challenges of distributed learning among UAVs, data scarcity is a critical concern, particularly for tasks with limited access to training data, such as fire detection and rescue missions. 
To reduce the cost of data collection, Lin et al. \cite{zhang2022forest} utilize transfer learning methods to adapt a pre-trained ResNet model from the ImageNet dataset to a forest fire dataset. 
 They further fine-tune specific convolutional layers to effectively extract relevant features from fire-related data
For the UAV-based bridge inspections, considering diverse factors like location, color, and lighting
Mostafa et al. \cite{aliyari2021uav} harness transfer learning techniques to reduce the impact of data distribution shifts caused by complex factors on the detection model, enabling it to accurately identify crack locations in bridges.
In dynamic environments encountered during rescue missions, Muhammad et al. \cite{atif2021uav} leverage transfer learning to enable models to quickly adapt to new geographical information and user distributions in different environments.

\subsection{Smart Factory}

A smart factory refers to a facility that utilizes various modern technologies, such as the Internet of Things (IoT) and artificial intelligence (AI), to achieve intelligent management and production.
Its primary goals are to reduce manual intervention, enhance work efficiency, and facilitate efficient decision-making \cite{buchi2020smart}.
For example, IoT technology connects various objects within the factory, such as equipment, products and workers, and then collects a variety of data through cameras, RFID, and other sensors.
The vast amount of collected sensory data can be processed on terminals, edge devices, or cloud platforms using data mining and machine learning techniques for various tasks, such as fault diagnosis and anomaly detection.
However, industrial systems are dynamic and continually evolving.
For example, task requirements, manufacturing equipment, operating environments, and network resources may change over time, as well as new equipment or tasks may be introduced.
Traditional deep learning approaches, relying solely on pre-trained models from the cloud server, often struggle to deliver satisfactory performance in complex and dynamic scenarios.
To address the above challenges in industrial environments, it becomes crucial to apply knowledge transfer learning methods to enhance model performance by leveraging prior knowledge.

\subsubsection{Manufacturing defect detection}

Manufacturing defect detection is a critical concern in the manufacturing industry. 
Compared to traditional manual detection methods, utilizing deep learning methods can reduce the cost while enhancing the accuracy and efficiency of detection.
However, acquiring a significant amount of labeled data for most products can be challenging. Therefore, many researchers turn to transfer learning methods to mitigate the requirement for extensive labeled data and to enhance the accuracy of defect detection.
For example, Max et al. \cite{ferguson2018detection} first pre-train the model based on two large-scale labeled datasets using convolutional neural networks (CNNs), and then fine-tune the model on a small-scale labeled dataset using transfer learning techniques.
Considering the differences between different data distributions, Yulong et al. \cite{zhang2022tire} propose a dual-domain adaptation model to simultaneously reduce the marginal distribution and conditional distribution gap between the target domain and the source domain, learning domain-invariant and discriminative features.
Furthermore, to enable the model to adapt rapidly to new data, Xian et al. \cite{lee2023xdnet} introduce a cross-domain meta-learning framework to learn generalizable features.
Surface defect detection is a crucial application in quality control. 
While most deep learning-based defect detection methods are effective in detecting common defects with abundant samples, they tend to perform poorly in detecting rare defects due to the limited availability of data.
To address this issue, Jiahui et al. \cite{cheng2021tl} propose a transfer learning-based few-shot surface defect detection method. 
The TL-SDD method consists of two stages: the model is trained on a dataset with a sufficient number of common defect data to learn common features, and then the pre-trained model is fine-tuned on a small number of rare defect data to learn some unique features based on learned knowledge.  
In defect detection tasks, achieving high detection accuracy is paramount, but it is also essential to minimize detection time to quickly identify defective products.
Hui et al. \cite{li2023transfer} propose a real-time surface defect detection framework based on transfer learning with multi-access edge-cloud computing (MEC) networks. This framework not only improves detection accuracy in data-sparse scenarios but also reduces detection latency through MEC networks.

\subsubsection{Machinery fault diagnosis}
Fault diagnosis aims to automatically identify the health status of machinery from collected data using machine learning methods, thereby reducing maintenance cycles and improving diagnostic accuracy.
Traditional machine learning methods often rely heavily on abundant labeled data to train robust diagnostic models.
However, obtaining sufficient training data in fault diagnosis tasks can be a challenging endeavor due to several factors.
Firstly, machines usually operate in a healthy state, with rare occurrences of failures, leading to a sample imbalance issue between normal and faulty data. 
Secondly, for each new equipment, it is impractical to collect data and train a model from scratch due to the high cost. 
Therefore, various approaches have turned to transfer learning to mitigate these issues, lowering the training cost while improving model performance.
In practice, the scarcity of fault samples prevents the learning of effective diagnostic models.
To address the issue of sparse data, Yao et al. \cite{hu2022few} propose a few-shot transfer learning method with attention for fault diagnosis of bearings. 
To improve the model's performance on target devices, they introduce an attention mechanism and fine-tune the pre-trained model using real data to reduce global feature differences. 
To reduce the cost of data labeling, Bo et al. \cite{zhang1805adversarial} employ domain adaptation methods to transfer knowledge learned from a source domain to unlabeled data in a target domain, enhancing the model's generalization across different working conditions.
Furthermore, Jun et al. \cite{zhao2023adaptive} introduce the adaptive siamese-based meta transfer learning networks for cross-domain fault diagnosis. 
This approach integrates meta-learning, adaptive batch normalization, and fine-tuning into a unified framework, which allows the model to quickly adapt to data in the target domain based on learned meta-knowledge.
In many applications, due to data privacy, data from different devices cannot be directly uploaded to the cloud for model training through transfer learning. 
To address this issue, some approaches have introduced federated transfer learning to improve model performance while safeguarding data privacy \cite{zhang2021federated, zhang2022data}.
For example, Junbin et al. \cite{chen2022federated} propose a federated transfer learning framework with discrepancy-based weighted federated averaging for bearing fault diagnosis. In this framework, each device locally trains multiple local models using its data and then uploads them to the cloud. Considering the distribution differences in data from different devices, they introduced a maximum mean discrepancy (MMD)-based dynamic weighted averaging algorithm to obtain a good global model suitable for all local devices.

\subsubsection{Anomaly detection of industrial processes}
Monitoring industrial processes automatically is crucial to detect anomalies timely, so the intervention could be conducted to improve the efficiency and quality of industrial control systems. 
Currently, deep learning plays a critical role in anomaly detection within industrial processes, primarily by identifying patterns of abnormal changes in real-time data.
Different processes in industrial control systems have unique functionalities and features, making it challenging to use pre-trained detection models directly, 
To address this issue, Jeongyong et al. \cite{park2023mendel} introduce the transfer learning technique aimed at efficiently constructing anomaly detection models tailored to different domains within industrial control systems (ICS). 
The proposed method first applies principal component analysis (PCA) to each model to acquire features that are compatible with those of other models.
Subsequently, it establishes reasonable mappings between the reduced features of different models to facilitate effective knowledge transfer.
Detecting energy consumption in industrial processes is crucial for energy efficiency within industrial systems. 
However, collecting a sufficient amount of representative data within a short timeframe, particularly for newly established processes, often proves to be unfeasible. 
To address this issue, Chuqiao et al. \cite{xu2021anomaly} propose a cluster-based deep adaptation network (CDAN) model based on transfer learning for detecting anomalies in spinning power consumption.
The CDAN model incorporates a cluster-based adaptation layer positioned between the feature layers of source and target networks. This layer effectively reduces feature disparities present in different environments. 
In addition, obtaining labeled information for a large amount of data is challenging, thus the model needs to operate in unsupervised scenarios. 
To enhance the performance of the detection model for unsupervised applications, Gabriel et al. \cite{michau2021unsupervised} propose an unsupervised transfer learning (UTL) approach for anomaly detection.
Unlike existing supervised transfer learning methods, this approach leverages adversarial deep learning to align features across different domains. 
Furthermore, a loss function is designed to encourage extracted features to include inherent discriminative information from each dataset.
Although transfer learning techniques have achieved promising results in anomaly detection, challenges persist in one-class classification. 
To address this issue, Wentao et al. \cite{mao2023deep} propose a deep one-class transfer learning algorithm based on domain-adversarial training. 
This approach integrates a hyper-sphere self-adaptive constraint into the domain-adversarial neural network. Moreover, an alternative optimization method is derived to seek optimal network parameters while promoting the construction of hyper-spheres in both source and target domains, aiming to make them as similar as possible. 
This adaptive transfer of one-class detection rules within domain-invariant feature representation significantly enhances end-to-end anomaly detection for one-class classification.



\section{OPEN ISSUES AND FUTURE DIRECTIONS}
\begin{figure}[h]
\centering
\includegraphics[width=0.485\textwidth,height=4.35cm]{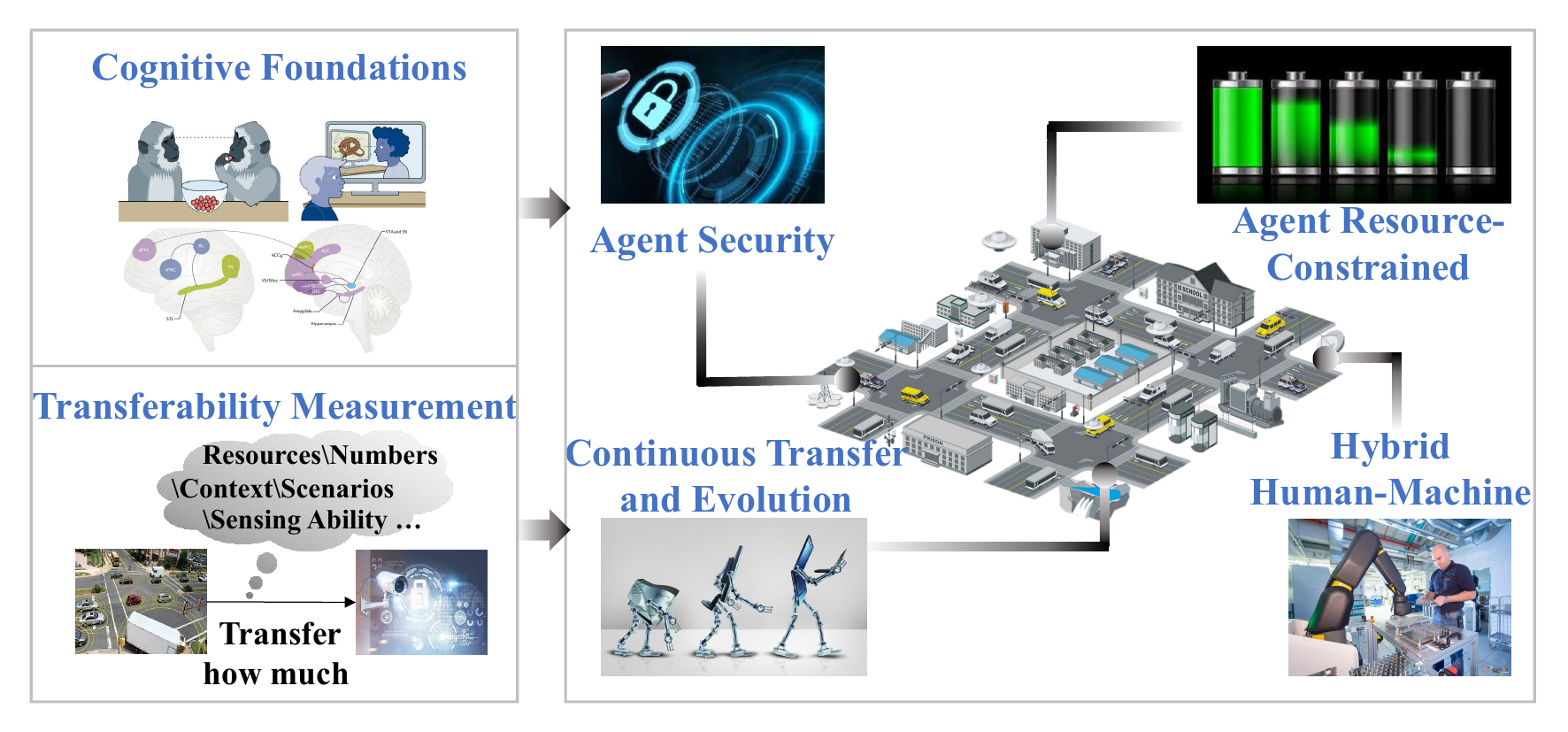} 
\caption{Open and further directions of crowd knowledge transfer.}
\label{Fig:6Open_0}
\end{figure}

Crowd transfer learning extends the traditional knowledge transfer and plays a significant role in the AIoT scenario. Although its key challenges and techniques have been investigated in Section \ref{TL_3} and \ref{TL_4}, there are still some open directions to be further researched. In this section, we mainly introduce them from the following six aspects (See Fig. \ref{Fig:6Open_0}): \textit{Cognitive foundations of crowd knowledge transfer}, \textit{transferability measurement mechanisms}, \textit{learning in resource-constrained IoT devices}, \textit{decurity in crowd knowledge transfer}, \textit{continuous crowd knowledge transfer and evolution}, and \textit{hybrid human-machine intelligence}.

\subsection{Cognitive Foundations of Crowd Knowledge Transfer}

\begin{figure}[!b]
\centering
\includegraphics[width=0.5\textwidth,height=5.41cm]{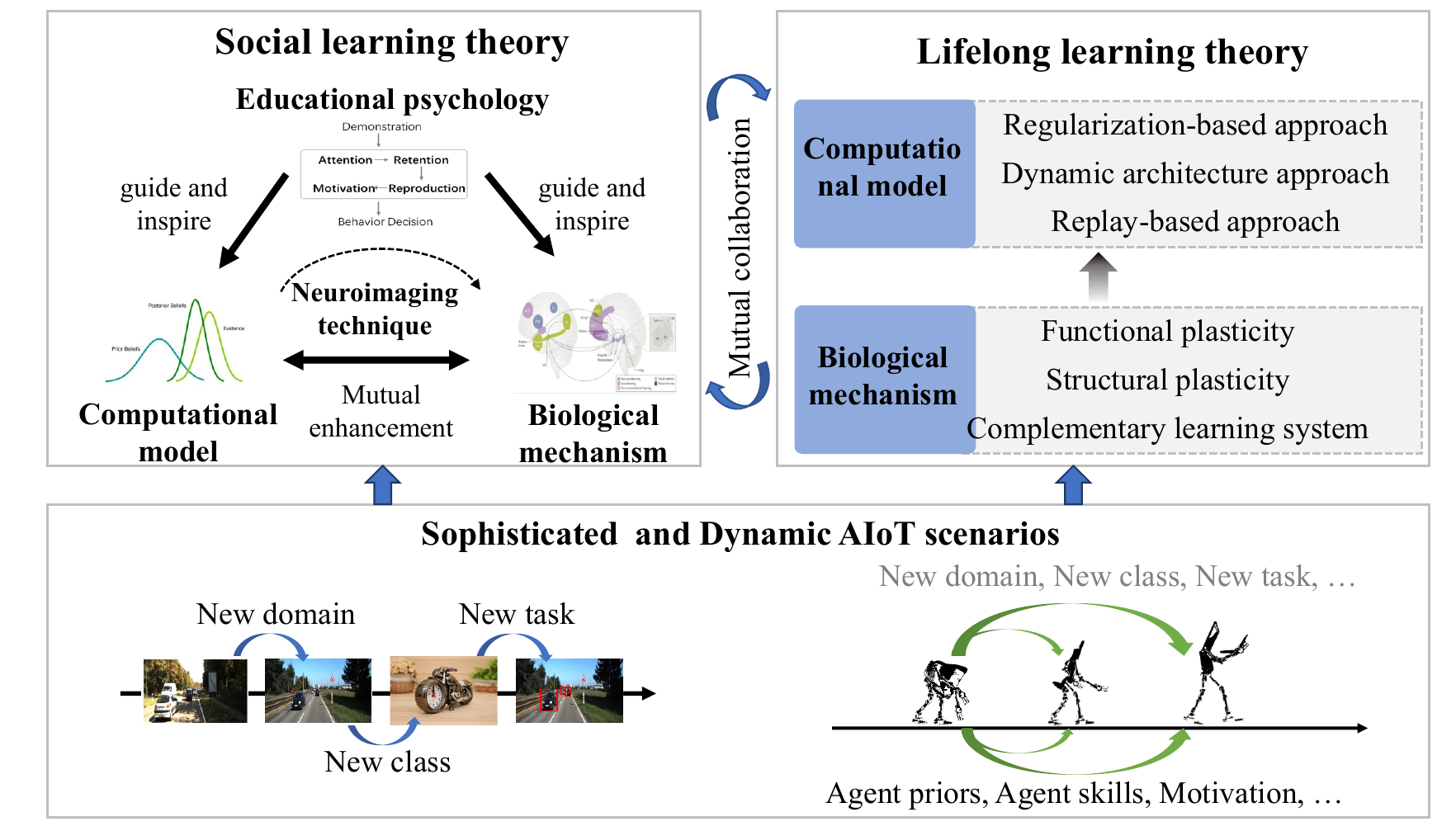} 
\caption{The cognitive foundations of crowd knowledge transfer.}
\label{Fig:6Open_1}
\end{figure}

The cognitive foundations of crowd knowledge transfer can provide theoretical guidance for the crowd knowledge transfer, better design the crowd knowledge transfer model, and enhance the performance of the transfer model. From the perspective of the sources of knowledge transfer, the research about the cognitive foundations of crowd knowledge transfer can include \textbf{continuous learning theory} (transfer of own past knowledge) and \textbf{social learning theory} (transfer of others’ knowledge), as shown in Fig. \ref{Fig:6Open_1}. 

\subsubsection{Continuous learning theory}
Continuous learning\cite{chen2018lifelong,hadsell2020embracing,parisi2019continual} can continuously process data streams in the real world, retaining and even integrating and optimizing old knowledge while absorbing new knowledge. One of the most significant challenges is catastrophic forgetting, that is, training a model with new data interferes with the previously learned knowledge, decreasing the model performance\cite{robins1995catastrophic}. Humans and other well-adapted animals can make appropriate decisions about new events based on previous continuous learned events\cite{chen2018lifelong}. Most of this ability to acquire, refine, and transfer knowledge is attributed to brain mechanism \cite{kudithipudi2022biological}. 

First of all, there is a multisensory system in which the visual cortex and auditory cortex of the brain can perceive visual and auditory signals respectively\cite{borden2022thalamic}, and then the thalamus dynamically controls the transmission of perceptual signals from the periphery to the cortex, integrating multisensory information\cite{lotter2023revealing}. In this process, to adapt and reflect changes in the external environment, the nervous system produces structural and functional changes and maintains these changes for a certain period of time\cite{power2017neural}. For structural changes, the central nervous system (CNS) can produce new neurons for new events to adapt to new tasks and skills, while for functional changes, neurotransmitters released by neuromodulatory neurons can help overcome catastrophic forgetting, and improve the understanding of uncertain environments and new tasks. For example, inspired by the plasticity of functional structures, NISPA\cite{burak2022nispa} achieves the goal of continuous learning through sparse neural networks with fixed density. And SparCL\cite{wang2022sparcl} realizes cost-effective continuous learning on edge devices through the sparsity of network structure. During the process, modularity plays an important role\cite{zeng2018reconstituted,bassett2011dynamic}. The system can perform specific functions without interfering with the rest of the system when the structure is modular, which reducing the dependencies of components, improving the system robustness and prompting the system adaptability and evolution. For example, Gallen et al.\cite{gallen2019brain} propose that brain modularity is a key biomarker that can predict cognitive plasticity, indicating that modular networks have stronger adaptability and are better at learning and problem-solving. When new memories are stored, the brain needs to generalize various experiences while training situational memories. The complementary learning system (CLS) theory\cite{mcclelland1995there,gepperth2016bio} suggests that the hippocampus uses fast learning methods to encode sparse representations to reduce interference. This learned information is then put back into the neocortical system so that overlapping representations of knowledge are retained for a long time. Furthermore, the generalization-optimized complementary learning systems (Go-CLS) theory \cite{sun2023organizing} suggests that the predictability of episodic memory needs to be considered to transfer partial hippocampal–cortical memory and optimize the generalization accuracy of the neocortex. Gepperth and Karaoguz\cite{kumaran2016learning} propose GeppNet+STM model inspired by CLS theory, in which GeppNet, a long-term memory learning module including self-organizing map and linear regression classification, can avoid the interference of old and new knowledge and alleviate the problem of catastrophic forgetting, and STM, a short-term memory learning module, can learn new knowledge and be transferred to long-term memory units at certain intervals to form long-term memory.

\subsubsection{Social learning theory}
Social learning theory (STL) means that people observe and learn or imitate the behavior of others, and reinforce or weaken this behavior according to self-established criteria\cite{olsson2020neural,heyes1994social}. Bandura\cite{bandura1986social} divides observational learning into four processes: attention, retention, reproduction, and motivation. The attention refers to observing the characteristics of the demonstrators’ behaviors, the observers’ own cognitive characteristics, and the relationships between the observers and the demonstrators. The retention indicates the representation demonstrators’ behaviors in symbolic form and maintenance of the brief exemplary demonstration in long-term memory. The reproduction represents the conversion of symbols and representations in memory into appropriate behaviors, reproducing the modeled behaviors observed in the past. The motivation means the observer is rewarded for performing the modeled behavior. Correspondingly, reinforcement learning models and Bayesian models have been proposed to portray social learning processes\cite{mathys2014uncertainty,hackel2015instrumental}: uncertainty representation, information integration, subjective expectations, and expected errors. Although both types of computational models are applicable to non-social learning, models that take into account social influences (e.g., individual theory of mind, inter-individual interactions, etc.) tend to provide a better description of social learning\cite{devaine2014social} The computational models of social learning processes explain the cognitive computations behind social decision making, but these only exist in the theoretical assumptions. Recently, neuroimaging techniques\cite{dunbar2012social,mizzi2022resting} have been increasingly applied to the field of social learning to provide a biological basis for the plausibility of computational models and to facilitate researchers' understanding of the specific roles played by specific brain regions in the social learning process.

The uncertain representation of social information refers to prior knowledge about the characteristics and intentions of others, as well as self-cognition. The social environment is high-dimensional and uncertain, Niv et al.\cite{niv2015reinforcement} study neural mechanisms of representation learning, and find that the intraparietal sulcus, precuneus, and dorsolateral prefrontal cortex can be involved in selecting what dimensions are relevant to the target task, and then effectively solve the problem of high-dimensional environments in reinforcement learning. When the social environment is constantly changing, dynamic contexts need to be considered in the process of information integration to integrate different sensory information. For example, in autonomous driving scenarios, cameras have high resolution to meet general needs, but in complex lighting environments, the reliability of autonomous vehicles is relatively low. At this time, LiDAR can provide high-precision spatial information and play a significant role\cite{cui2021deep}. Such information integration process is mainly reflected in the dorsolateral prefrontal cortex, inferior parietal lobule and anterior cingulate cortex\cite{de2017social}. After integrating multi-sensory information, neural mechanisms of decision making mainly involve related regions such as the ventromedial prefrontal cortex, orbital frontal cortex, and ventral striatum. For example, Báez-Mendoza et al.\cite{baez2021social} track the social interaction among three rhesus monkeys, demonstrate the core of neurons in the dorsomedial prefrontal cortex in group behavior, and reveal the cellular mechanisms that support social group interactions. Inspired by this, pheromone is introduced to multi-agent reinforcement learning systems to work out the large-scale multi-agent coordination problem\cite{cao2022pheromone}. Finally, the calculation of the expected error requires a comparison of the expected mental utility with the actual observed results, mainly involving brain regions such as the ventromedial prefrontal cortex, orbital frontal cortex, putamen, and ventral striatum\cite{hackel2015instrumental,zhang2020brain}. Neurons in the ventral tegmental area projecting to the nucleus accumbent release dopamine, which serves as a reward for altering neuronal activity in this region. During social interactions, additional dopamine can be released by the hypothalamic, that is the social reward\cite{hu2021amygdala,hung2017gating}. Inspired by this, Jaques et al.\cite{jaques2019social} introduce a social influence intrinsic reward to promote multi-agent communication and encourage multi-agent collaboration, thus completing collaborative tasks more quickly and accurately.

Although the continuous learning theory and social learning theory have been applied in several fields, they still face many challenges:
\begin{itemize}
\item \textbf{Spontaneous exploration and autonomous decision-making}. For complex and dynamic AIoT scenarios, the agent should be dynamically adaptive, continuously and spontaneously exploring the environment and making autonomous decisions, rather than the traditional continuous neural network models that are merely data-driven learning. Inspired by the fact that infants can constantly explore through curiosity\cite{barnett2001scientist,de2018curiosity}, curiosity-driven reinforcement learning can imitate infants’ curiosity and constantly explore, continuously reward themselves, and ultimately find the final goal\cite{burda2018large}. Therefore, how to better explore the environment, prevent uninterrupted exploration, and quickly achieve the specific goal based on curiosity and intrinsic motivation mechanism is an important research question. 
\item \textbf{Multisensory mechanisms assisted multimodal integration}. AIoT data in complicated scenarios is multimodal, for example, in autonomous driving scenarios, multimodal data includes sensing data from devices such as LiDAR and cameras, map data including road topology and lane information, vehicle sensor data such as speed and steering angle, as well as external data like weather and traffic flow. Comprehensive analysis, fusion, and mining of these data can provide autonomous driving systems with comprehensive environmental perception and decision support. In the human brain, multimodal mechanisms play a similar role. Therefore, how can agents better select, align and fuse information from various modalities with the aid of multisensory mechanisms in the brain? 
\item \textbf{Synergistic enhancement of social learning theory and continuous learning theory}. In the face of constantly arriving new tasks, how can agents, with the help of social learning theory and continuous learning theory, continuously learn information about inter-agent communication, interaction, and decision making without forgetting the knowledge related to previous tasks and collaborate more effectively to accomplish the current complex tasks?
\end{itemize}

\subsection{Transferability Measurement Mechanisms}
\begin{figure}[h]
\centering
\includegraphics[width=0.485\textwidth,height=4.1cm]{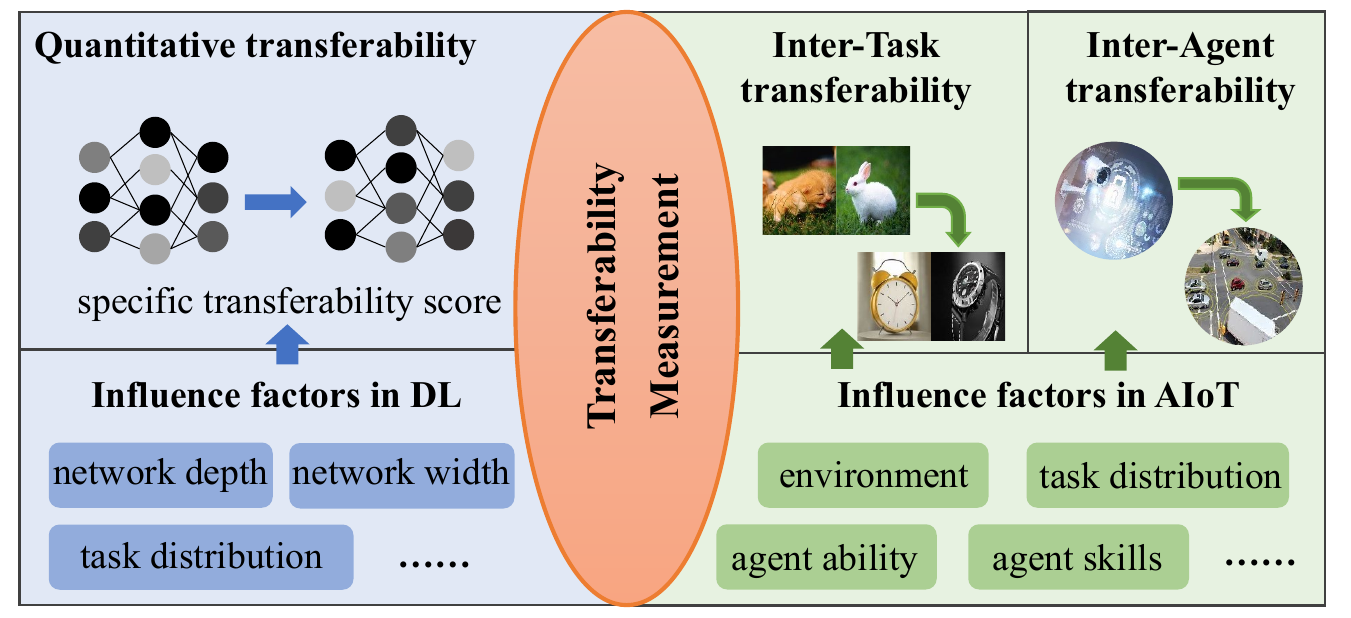} 
\caption{Core contents of transferability measurement mechanisms.}
\label{Fig:6Open_2}
\vspace{-3mm}
\end{figure}

Crow transfer learning enables to transfer similar knowledge in source agents to target agents with less data from different perspectives and levels, improving the prediction of the target agents. A mass of approaches, such as cross-agent transfer and cross-context transfer, have been proposed to improve the positive transfer from source agents to target agents or source scenarios to target scenarios, etc., and have been successfully implemented to diverse fields such as urban computing, connected vehicles, multi-robot system, UAV swarm system, smart factory, etc. Unfortunately, knowledge transfer without distinction may lead to negative transfer\cite{wang2019characterizing}, resulting in a large challenge in crowd knowledge transfer: which source agents enable contribute to the performance of the target agents, and how much its knowledge can be transferred to the target agents? Knowledge transferability provides a promising solution, which is the ability to achieve transferable knowledge from source agents and reuse this information to reduce the generalization error of target agents\cite{jiang2022transferability,bao2019information}, as shown in Fig. \ref{Fig:6Open_2}.

The current research on knowledge transferability in deep learning areas aims to explore the factors affecting transferability and specific metrics of transferability. For example, Yosinski et al.\cite{yosinski2014transferable} first confirm that the initial three layers of the neural network represent general features, and the transfer performance is relatively good, and the transferring network can accelerate the learning and optimization process. Azizpour et al.\cite{azizpour2015factors} explore several factors affecting the transferability of ConvNet representations: source ConveNet architecture (network depth, network width, etc.), source task distribution, and target ConveNet architecture (network layer, dimensionality reduction, fine-tuning, etc.). BSP\cite{chen2019transferability} confirms that feature transferability is dominated by the eigenvectors with top singular values. All these are qualitative factors, and some research specific quantitative measures of transferability. For example, H-score \cite{bao2019information} estimates the task transferability in classification problems through the statistical and information-theoretic methods. NCE\cite{tran2019transferability} and LEEP\cite{nguyen2020leep} model the correlation between the label distribution of the two classification tasks to obtain the transferability score of the two tasks. And OTCE\cite{tan2021otce} estimates the domain difference and task difference based on the Optimal Transport (OT), and then employs the linear combination between them as the final transferability score.

Different from the simplicity of neural networks in deep learning, each agent in AIoT areas can participate in the process of sensing, calculating and decision making, so the transferability measurement is more complex, and task distribution, environment, agent skills/strategies etc. can affect the transferability measurement. At present, the transferability measurement in AIoT areas can be divided into \textbf{inter-agent transferability measurement} and \textbf{inter-agent transferability measurement}. The former aims to boost the performance of target tasks through transferring knowledge such as data/label distribution of source tasks, while the latter aims to enhance the effectiveness of target agents via reusing knowledge such as agent skills/strategies. For example, Qin et al.\cite{qin2022multi} mine the common structure between tasks based on the similarity of state transition and reward function of different tasks, achieving the transferability measurement between tasks. When there are large differences in task knowledge, it will not be efficient to directly measure the similarity between tasks for transferring. The skill transfer method\cite{liu2020skill} is designed to abstract the previously learned knowledge at a high level, for example, Hausman et al.\cite{hausman2018learning} utilizes the connection between reinforcement learning and variational inference to learn the hierarchical random strategy, simulating the complex related structure and multimodal information in the action space, capturing the skill embeddings of the agent, and achieving the transferability measurement between agents.

In AIoT areas, it is usually more necessary for multi-agent cooperation to achieve complex intelligence and improve the task efficiency. Stanley et al.\cite{stanley2022transferability} develop a transferability metric to determine the most similar agents based on the physical differences of the agents and noise differences, then can better get the mapping relationship between the human and robot, generating human-like robot movement. Due to the differences in communication, reward, environment and other aspects between agents, it is difficult for agents to select some similar agents with their own physical capabilities. HAMA\cite{ryu2020multi} mines the relationship between each agent and other agents and environments through the hierarchical graph attention network, calculating the policy transferability between agents in an end-to-end form.

Current crowd knowledge transferability metrics have been applied to urban computing, multi-robot systems, and smart factory, etc. However, there are still some challenges to solve.

\begin{itemize}
\item \textbf{The ante-hoc transferability metric}. Most crowd knowledge transferability metrics are post-hoc transferability metrics, depending on the pre-trained model and requiring expensive computation. It is a crucial future research direction that how to analyze the ante-hoc transferability metrics, reduce the relevance of the pre-trained model, mitigate the time and computational resource expenditure, and select more useful tasks or agents.

\item \textbf{Diverse transferable knowledge}. In the field of AIoT, the knowledge can be the experience samples obtained by the agent itself, the knowledge of other similar tasks, and the knowledge of other agents (experience samples, strategies, skills, models, etc.), etc .\cite{sinapov2015learning,zhu2023transfer}. How to decouple and quantify the fine-grained transferability is also an important research direction.

\end{itemize}

\subsection{Learning in Resource-Constrained AIoT Devices}
\begin{figure}[h]
\centering
\includegraphics[width=0.485\textwidth,height=6.2cm]{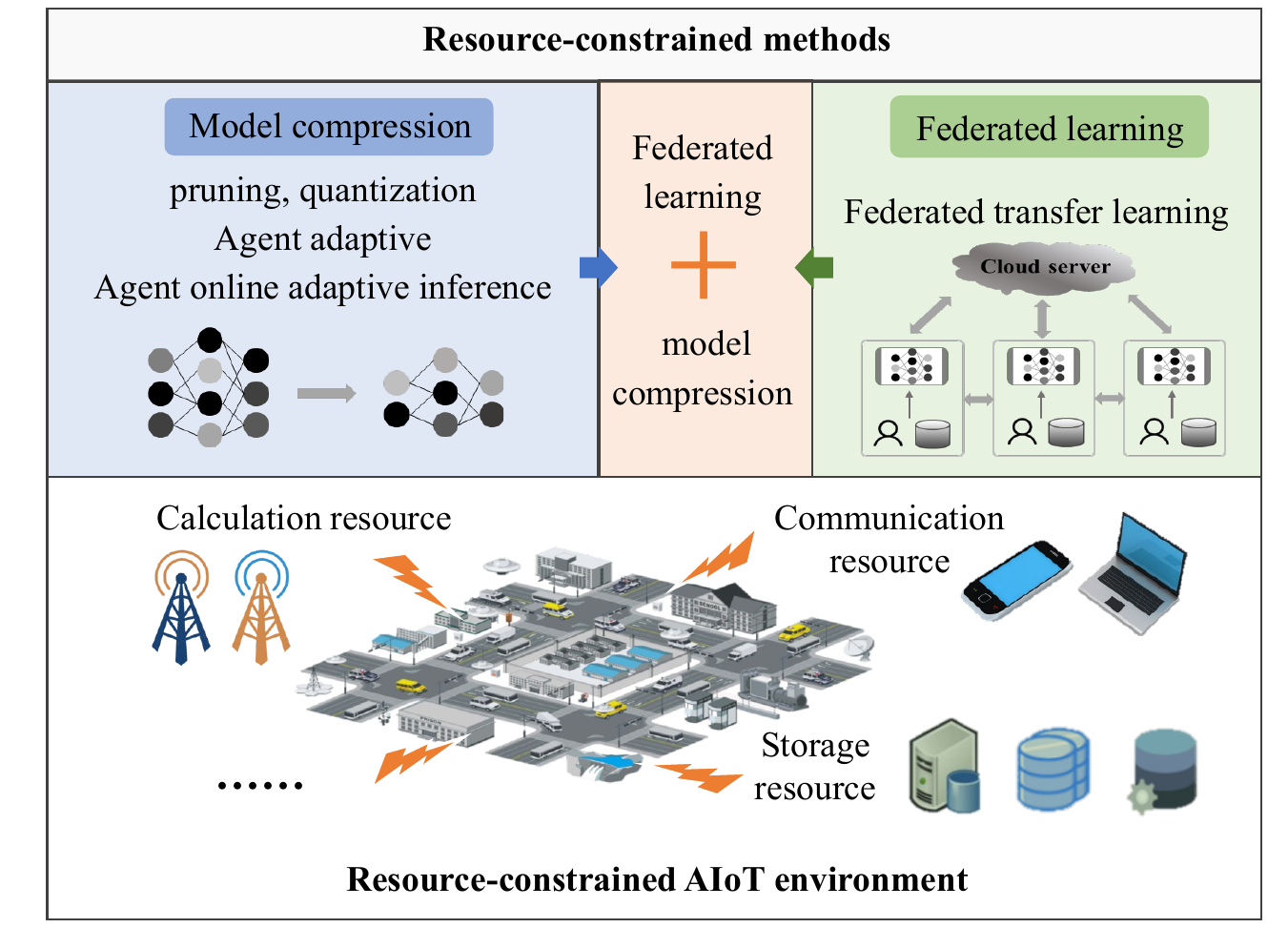} 
\caption{Core contents of learning in resource-constrained AIoT devices.}
\label{Fig:6Open_3}
\vspace{-4mm}
\end{figure}

Agents with the ability of computing and storage in the era of AIoT train their own models based on collected data instead of uploading data to the cloud server to improve training efficiency. Simultaneously, if the data is not uploaded to the cloud server, the issue of data security and privacy can be adequately addressed. However, there are resource constraints such as energy, memory and computing power in AIoT agents, resulting in poor performance in the agent side. The existing solutions can be classified into two categories from the perspective of whether the agent participates in training: \textbf{model compression methods} and \textbf{federated learning methods}, as shown in Fig. \ref{Fig:6Open_3}.

For the former, the general approach (i.e., pruning, quantization, etc.) is to compress the existing model with good performance into a model architecture with small and exquisite parameters, so that the compressed model can run on the agents with small size and small abundance \cite{zhang2021compacting}. This kind of approach does not distinguish the difference between agents, so, the agent adaptive method (i.e., knowledge distillation, neural architecture search, etc.) can directly generate a model with excellent performance and few parameters using the agent data. For example, AdaDeep\cite{liu2018demand} regards the model compression techniques as the coarse-grained DNN hyperparameters, and utilizes the reinforcement learning to automate the parameter selection based on the different task requirements and platform resource constraints, achieving the adaptive lightweight model architecture search. However, the agent resource situations (i.e., storage, computation, power, etc.) change dynamically during the real running and inference, so the online adaptive inference method can achieve the real-time balance between of accuracy and model energy consumption according to the dynamic resource situations. For example, AdaSpring\cite{liu2021adaspring} is a runtime dynamic self-evolving model compression framework. When facing dynamic terminal resource scenarios (power, storage, and other platform resources), it enables online compression policy selection and deployment of deep learning models without retraining to achieve the optimal utilization of AIoT terminal resources.

For the latter, federated learning provides a solution, which is a machine learning framework for multi-client cooperative training method with the organization of a central server. In real-world scenarios, different agents are heterogeneous in terms of their computing resources and model architecture, federated transfer learning (FTL) model provides a more promising solution\cite{liu2020secure,saha2021federated}. It aims to integrate the knowledge of crowd agents to train more individualized local models in a safe manner. The specific methods have been introduced in chapter \ref{chapter:4.1.1}. In addition, Because the unique client-server architecture of federated learning, the model often needs to be as small and effective as possible, and at the same time, it needs to ensure the client real-time and transmission rapidity, so researchers combine the federated learning with model compression approach. For example, FedGKT\cite{he2020group} trains the small CNN on client nodes, and periodically transfers knowledge to large CNN on server nodes through knowledge distillation, reducing the communication overhead.

Although compression methods and federated learning methods have achieved wider application in resource-constrained scenarios, the following challenges still exist:
\begin{itemize}
\item \textbf{The agent heterogeneity and mobility}. On resource-constrained devices, different agents have varying data distributions, computing power, memory, and other resource sizes. In the training process of federated learning, the training cycles differ for different agents due to their resource disparities. Additionally, agents are constantly on the move, and some agents may encounter difficulties in uploading their local models to the server. Therefore, how to conduct efficient and reliable federated learning in the context of agent heterogeneity and mobility is an important research issue.

\item \textbf{Combinatorial optimization at the algorithmic level}. 
On resource-constrained devices, different resource-constrained learning methods achieve performance improvements at different levels. Quantization methods reduce data computation precision to lower computational load and storage requirements, thereby optimizing performance at the data computation precision level. Pruning methods eliminate redundant and unnecessary model parameters to reduce the size of the model structure, resulting in performance optimization at the model structure level. Federated learning, on the other hand, performs model training and parameter sharing on distributed devices, improving data privacy protection and model generalization performance at the model training level. Therefore, by combining different optimization methods based on specific application requirements and device limitations, one can fully leverage their performance enhancement effects at different levels and maximize performance and efficiency on resource-constrained devices.

\item \textbf{Co-optimization at algorithmic, compilation, and hardware layers}. Regardless of the type of resource-constrained learning methods, they are all optimizations at the algorithm level. In an agent, optimizations at the compilation and hardware levels can also reduce training time and storage space. Therefore, how to synergize the optimizations at the algorithm, compilation, and hardware levels to achieve optimal performance and efficiency on resource-constrained devices?
\end{itemize}

\subsection{Security in Crowd Knowledge Transfer}
\begin{figure}[h]
\centering
\includegraphics[width=0.488\textwidth,height=5.92cm]{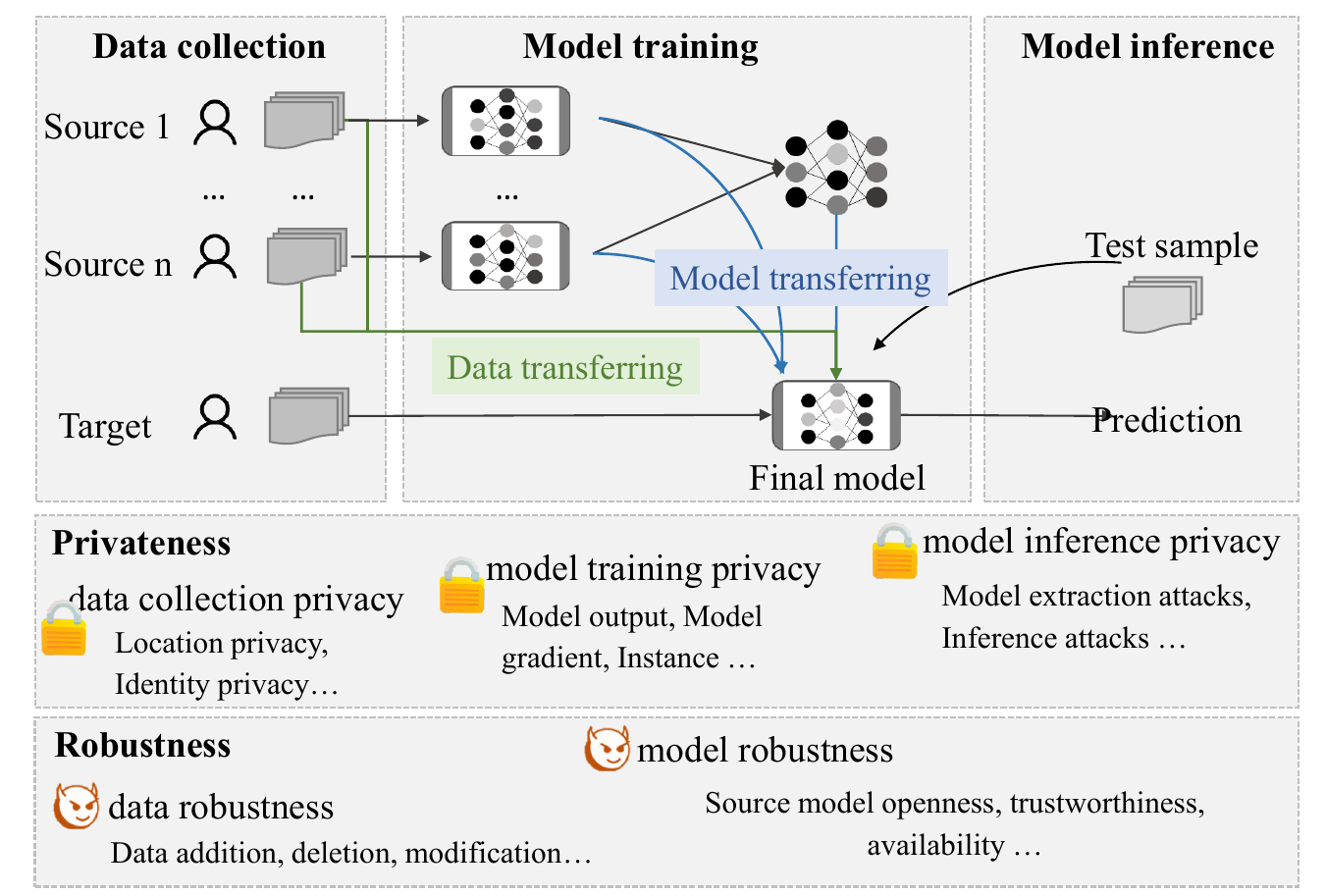} 
\caption{Core contents of security in crowd knowledge transfer.}
\label{Fig:6Open_4}
\vspace{-4mm}
\end{figure}

Despite its wide application, crowd knowledge transfer model is insecure and can be attacked. Wang et al.\cite{wang2018great} attack the transfer learning model by perturbing the input, and Rezaei et al.\cite{rezaei2019target} implement a target-agnostic attack by training the softmax layer of a pre-trained network with reverse maxima, such that any image with an arbitrary input is output as the desired class. Therefore, secure transfer is an important research problem to avoid data leakage or malicious attacks on the model that can affect the participants\cite{zhang2022remos}. The current research on the security of crowd knowledge transfer can be divided into two directions: \textbf{privateness} and \textbf{robustness}\cite{lyu2022privacy}, as shown in Fig. \ref{Fig:6Open_4}.

The privateness of crowd knowledge transfer refers to attempting to obtain private information or other benefits from various stages of transfer learning without disrupting the model training and inference process. Based on three phases of crowd knowledge transfer process: data collection phase, model training phase, and data testing phase, the privacy of crowd knowledge transfer manly contains: data collection privacy, model training privacy and model inference privacy. The data collection privacy mainly refers to the location privacy, identity privacy of agents and the privacy of perceptual data during data collection, as well as the communication privacy between agents. For example, in a human-machine collaborative smart grid system, users send their electricity consumption history information to the grid cloud for grid optimization and dispatch, however, these history records may reflect information about their daily life patterns and cause privacy leakage\cite{mcdaniel2009security}. The model training privacy means the data privacy about transferring the model output, transferring the gradient and transferring the instance, etc. For example, Zhu et al.\cite{zhu2019deep} propose the DLG methods to restore the original training data from shared gradient information. And the model inference privacy includes two categories: model extraction attacks and inference attacks. The model extraction attack refers to recovering the parameters or functions of the original model based on response information after repeatedly sending data to the API interface of the model, while the inference attack is an attack that determines whether a data record is in the model’s training dataset, given the access rights to the record and the model. For example, for the attribute inference attack\cite{ganju2018property}, the attacker can infer the characteristics of the training dataset, including the age distribution, gender distribution, ,and other related features. For the above privacy issues, cryptography-related techniques are introduced to mitigate the consequences of privacy threats, mainly including homomorphic encryption\cite{aono2017privacy}, secure multi-party computation\cite{bonawitz2017practical}, differential privacy\cite{triastcyn2019federated}, VerifyNet\cite{xu2019verifynet}, and other methods. For example, differential privacy algorithms add random noise in the process of transmitting gradients to hide or blur the actual results until they are indistinguishable, achieving protection of private data, and Geyer et al.\cite{geyer2017differentially} advocate for anonymizing the contributions of clients during training, preventing any user from inferring private data of other users from the aggregation model and avoiding the client from other clients' discrepancy attacks.

The robustness of crowd knowledge transfer refers to interfering the training or inference process of the transfer models, thus affecting the convergence speed or inference results during training, and mainly contains the robustness of data and transfer models. The robustness threat of data mainly contains the tampering of training features (e.g., malicious infiltration of poor-quality data, modification of data, deletion of data, etc.) and the tampering of training labels. For example, Chen et al.\cite{chen2017targeted} propose to generate the poisoning samples based on the Input-instance-key strategy and Pattern-key strategy, and subsequently incorporates these samples into the training dataset to accomplish the target attack. The robustness threat of transfer models mainly contains the openness of the source model, the trustworthiness of the agents and cloud servers, and the presence of a substantial number of agents for communication, and usually there are transfer model poisoning attacks\cite{xie2019dba}, free-riding attacks\cite{fraboni2021free}, witch sybil attacks\cite{fung2020limitations}, communication bottleneck attack\cite{luping2019cmfl}, etc. For above robustness issues, some defensive methods are employed to lighten the consequences of robustness threats, mainly including data sanitization\cite{koh2022stronger}, anomaly detection\cite{zhao2020shielding}, knowledge distillation\cite{lin2020ensemble}, pruning\cite{jiang2022model}, trusted execution environment\cite{chen2020training}, federated multi-task learning\cite{smith2017federated}, and other methods. For example, pruning techniques are applied to remove anomalous neurons generated by the model after a poisoning attack to purify the entire model, and PruneFL\cite{jiang2022model} enables adaptive parameter pruning in a federated learning environment, which reduces communication and computational overhead while maintaining similar accuracy of the original model.

The development of attacks and defenses in crowd knowledge transfer learning is still immature, and there are still some challenges to be addressed. 
\begin{itemize}
\item \textbf{Balancing privacy and robustness}. The privacy enhancement strategy enhances privacy protection at the expense of system robustness and accuracy, and conversely, models with high robustness are more vulnerable to privacy attacks\cite{song2019membership}, so how to protect privacy without compromising the model robustness is a problem to be considered in the future. 

\item \textbf{Privacy enhancement strategies}.The cost of the privacy enhancement strategy is high, where the transfer model training fails when there are malicious attackers, sudden dropouts of agents, and other conditions. At the same time, the privacy enhancement strategy usually requires more data transmission, more communication overhead, etc. All these exist the problem of the high cost of the privacy enhancement strategy, so how to design low-cost and lightweight privacy enhancement strategies is an important research direction for the future. 

\item \textbf{Security attack and defense system}. The current crowd knowledge transfer learning attack and defense system is not perfect, and it is difficult to achieve effective defenses when facing new attack threats\cite{xie2019dba} or diverse attack threats\cite{li2021fleam}. Therefore, how to analyze and reason all possible potential attacks and privacy issues, and combine secure encryption techniques to build a crowd knowledge transfer learning security attack and defense system is an important issue in the future.

\end{itemize}

\subsection{Continuous Crowd Knowledge Transfer and Evolution}
\begin{figure}[!b]
\centering
\includegraphics[width=0.488\textwidth,height=5.0cm]{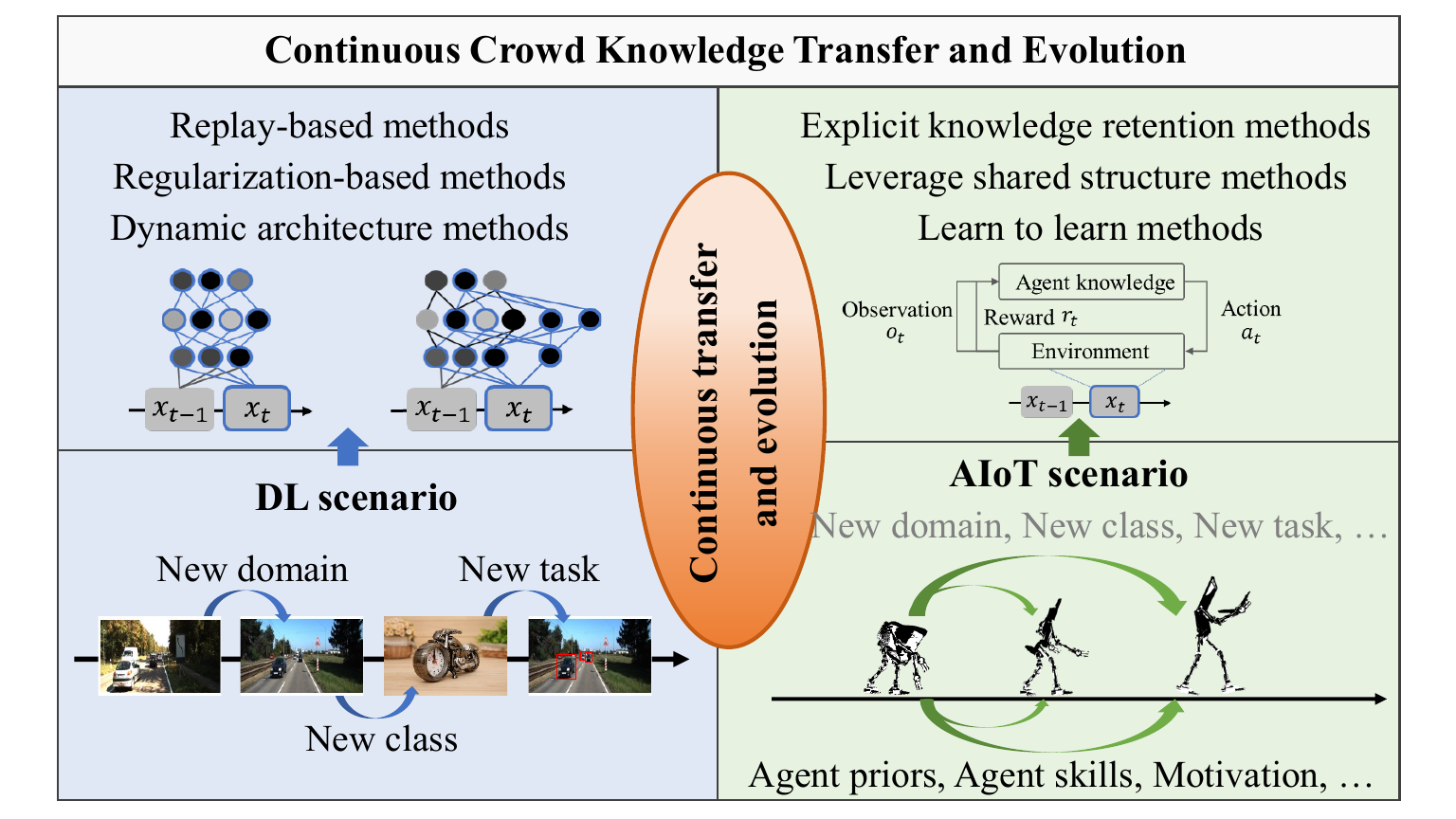} 
\caption{Core contents of continuous crowd knowledge transfer and evolution.}
\label{Fig:6Open_5}
\end{figure}

Data in AIoT scenarios are continuously generated in the form of streams. These large amounts of data, transmission delays, and data privacy and security make it challenging to process these data in the cloud\cite{deng2020edge}. Edge computing provides a promising computing paradigm, which puts computational resources closer to data sources, provides services such as computing and storage, meets data real-time analysis and intelligent processing, and reduces network bandwidth while ensuring the security and privacy of data\cite{shi2016edge}. Nevertheless, the limited storage space and computational capacity of agents fail to store prior knowledge when new classes, new tasks, and new domain data arrive, making the inference data deviate greatly from the training data, resulting in lower accuracy. Inspired by the skills of humans and animals to continuously acquire and transfer knowledge throughout their lives, researchers have proposed continual learning approaches to address these issues\cite{parisi2019continual}, as shown in Fig. \ref{Fig:6Open_5}.

Uncertain multimodal data streams are continuously arriving in AIoT scenario. Humans and animals are constantly learning in a meaningful sequence, forming increasingly complex concepts and skills\cite{tani2016exploring}. Inspired by this, agents should be able to interact continuously with complex AIoT environments to learn and understand the streaming data to gain long-term experience and skills. Current solutions can be divided into three categories \cite{khetarpal2022towards}: explicit knowledge retention methods, shared structure methods, and learn to learn methods. Explicit knowledge retention methods aim to leverage parameter storage \cite{liu2021conflict}, knowledge distillation\cite{zhang2022catastrophic}, and experience replay \cite{queeney2021generalized} to solve the problem of agent catastrophic forgetting. For example, Lampinen et al. \cite{lampinen2021towards} propose a hierarchical chunk attention memory model, which first pays high-level attention to a rough summary of the chunks, and then pays detailed attention to the most associated chunks to achieve memory for previous events. Leverage shared structure methods aim at abstracting explicit knowledge (such as modularity \cite{rosenbaum2019routing}, state abstraction \cite{allen2021learning}, skill \cite{campos2020explore}, etc.), automatically decomposing complex tasks into smaller subtasks, reusing previously solutions solving subtasks to address the issue of inaccurate explicit knowledge. For example, Devin et al. \cite{devin2017learning} propose to decompose the neural network strategy into task-specific modules and agent-specific modules, which allows task information to be shared between agents, as well as agent information to be shared between tasks, solving the problem of agent generalization with fewer samples. Inspired by the fact that a key component of continuous learning in infants is their spontaneous ability to autonomously generate goals and explore their environment driven by intrinsic motivation \cite{gopnik1999scientist}, Learn to learn methods aim to leverage the self-modifying policy \cite{schmidhuber1998reinforcement} and intrinsic motivation strategy \cite{barto2013intrinsic} to learn how to improve their adaptive and learning abilities. For example, Nagabandi et al.\cite{nagabandi2018learning} put forward a model-based fast adaptive meta reinforcement learning algorithm. It first obtains a prior model through meta-training, and then leverages this prior model and recent observations to adjust and update the overall model to fit the current environment, achieving fast online self-adaptation. However, when the environment is unknown, the above methods make it difficult to achieve good results. Ning et al. \cite{ning2020could} propose that only when an agent has independent and intrinsic thinking space like humans can it improve its learning ability, that is, intrinsic motivation can enable the agent to learn useful environment models and help it learn more complicated tasks. Therefore, Hester et al. \cite{hester2017intrinsically} proposes a reinforcement learning with an intrinsic rewards algorithm, including two intrinsic motivations: one is to explore the uncertain content of the model, and the other is to gain innovative experience that has not been trained on the model. Therefore, agents can explore new domains in a developmental and curious way, gradually learning more complex skills.

 With the continuous development of artificial intelligence, continuous crowd knowledge transfer and evolution method has been applied in multiple fields, such as urban computing, multi-robot systems, UAV swarm systems, and smart factory, etc. However, there are still some challenges in the AIoT areas:

\begin{itemize}
\item \textbf{Storage and computational resources}. The storage and computational resources of AIoT agents are limited, which makes it difficult to adapt the continuous learning tasks. Therefore, how to design more efficient model compression and parameter sharing methods for continuous learning under limited resources?

\item \textbf{Adaptive learning strategies for new scenarios}. Data in AIoT areas are constantly updated and changed, and there are scenarios of new tasks, new classes, and new domains, how to design adaptive learning strategies to enable the model to autonomously adjust learning methods according to changes in the scene;

\item \textbf{Agnostic task boundary}. Most continuous crowd knowledge transfer and evolution approaches require training and prediction in scenarios with known task boundaries, i.e., when learning a coming task, the model needs to be informed whether it is a new task or not. However, practical scenarios often face the problem of agnostic task boundary, how to clearly define the task boundary and how to solve the continuous transfer and evolution in this scenario are important issues.
\end{itemize}

\subsection{Hybrid Human-Machine Intelligence}

\begin{figure}[!b]
\centering
\includegraphics[width=0.485\textwidth,height=5.55cm]{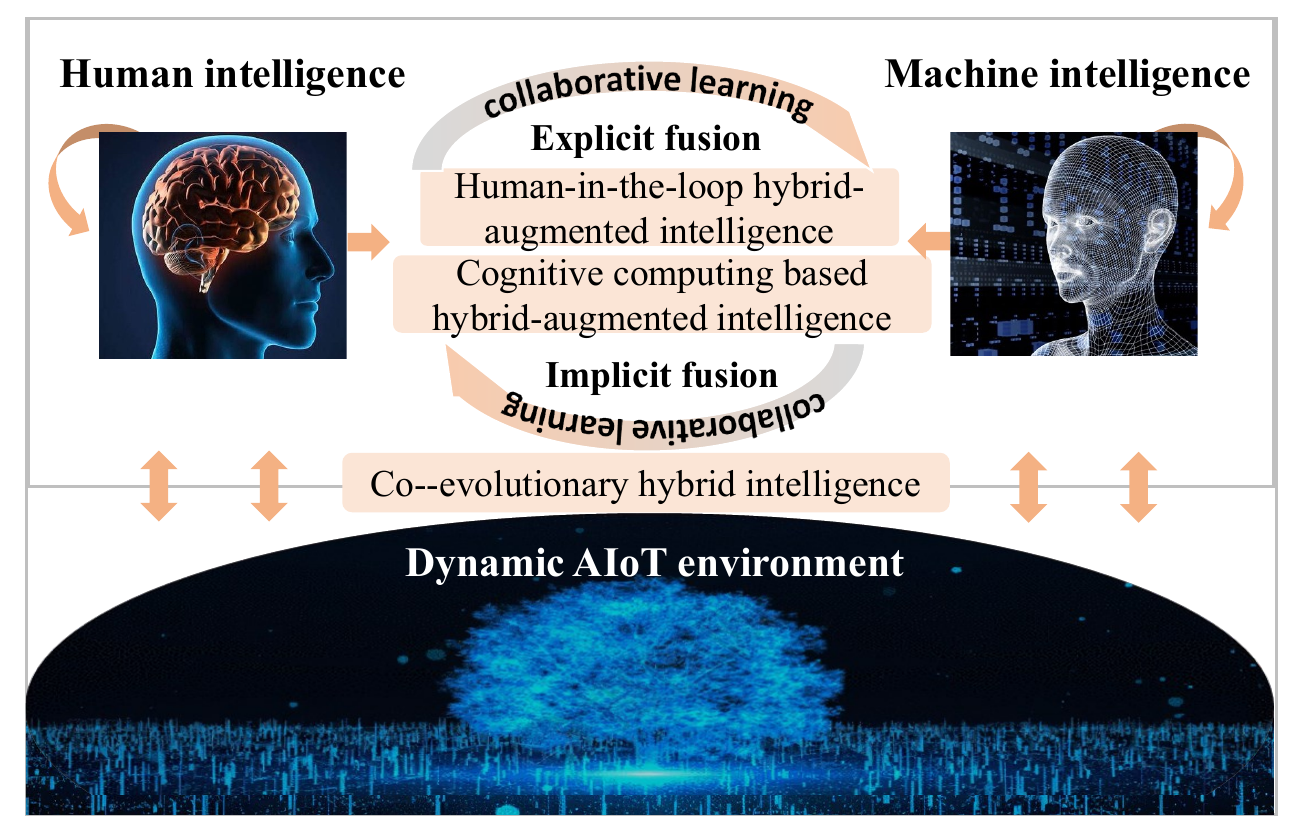} 
\caption{Core contents of hybrid human-machine intelligence.}
\label{Fig:6Open_6}
\end{figure}

Although crowd transfer learning has achieved excellent performance, it is limited by the amount of data. The addition of expert guidance can make agents learn and transfer more relevant knowledge more quickly. Hybrid human-machine intelligence additionally taking into account the advantages of human cognition and reasoning has become a new research trend \cite{zheng2017hybrid}. Hybrid human-machine intelligence, also named human-machine collaboration \cite{haesevoets2021human}, human-machine cooperation \cite{hoc2000human}, etc., is a combination of human intelligence and machine intelligence that achieves goals that neither humans nor machines can achieve. The current forms of hybrid human-machine intelligence implementation can be divided into two categories \cite{zheng2017hybrid}: \textbf{human-in-the-loop hybrid-augmented intelligence} (HITL hybrid intelligence) and \textbf{cognitive computing based hybrid-augmented intelligence} (CC hybrid intelligence), as shown in Fig. \ref{Fig:6Open_6}.

HITL hybrid intelligence is a hybrid intelligence paradigm that incorporates human involvement into the intelligence system to enhance the confidence of the system decisions, such as human-directed reinforcement learning method \cite{arakawa2018dqn}, prioritization of human preferences \cite{christiano2017deep}, and the attention mechanism of the human eye \cite{zhao2016deep}, and other human guidance. For example, Giffith et al. \cite{griffith2013policy} apply human feedback as a strategy label for agent actions to solve complex tasks.

CC hybrid intelligence improves the perception, reasoning, and decision-making capabilities of machines by simulating the cognitive mechanisms and functional patterns of the human brain. The Imitation Learning \cite{hussein2017imitation} or Learning by demonstration \cite{argall2009survey} method, is the process by which a human demonstrator conveys a strategy by personally performing the task and demonstrating the correct operation to an intelligence agent, implicitly providing prior information about the task without the need for the relevant expertise and knowledge, thereby improving the learning efficiency. For instance, Chernova et al. \cite{chernova2009interactive} present a confidence-based autonomy approach, which consists of two components: Confident Execution and Corrective Demonstration. 
When the trust level falls below the threshold,
the learner will require additional instruction to improve the confidence level of performing actions; And when experts observe incorrect actions by learners, 
they rectify those actions and incorporate the corrected actions into the training dataset to enhance future task performance.

However, the AIoT scenario has problems such as complex and variable contexts, explicit differences in data distribution, and increasing evolution of data and learning tasks, and \textbf{co-evolutionary hybrid intelligence (CHI)} \cite{krinkin2021co} is proposed to solve these problems. For example, lifelong learning \cite{aspin2000lifelong}, continual learning \cite{hadsell2020embracing}, mutual learning \cite{ansari2018rethinking}, adaptive evolution \cite{hawks2007recent} and other methods enable humans to continuously increase their cognitive, understanding, and reasoning capabilities, and machines to continuously increase their perception, inference, and decision-making capabilities, ultimately realizing the symbiosis coexistence of human intelligence and machine intelligence, and mutual development, transfer, and complementarity in the process of co-evolution. For example, Ansari et al. \cite{ansari2018rethinking} propose a two-way process of mutual learning, in which humans and machines interact, rely on, act, or influence each other in collaboration, thereby enabling humans and machines to learn and progress together.

Although the current hybrid human-machine intelligence has been applied in many fields, such as smart factory, public safety, and smart healthcare, etc., its development is in its infancy and still faces many challenges.
\begin{itemize}
\item \textbf{Various expert knowledge}. In terms of HITL hybrid intelligence, human understanding and cognitive ability (expert knowledge) for different domains are different \cite{neisser1996intelligence}, and domain experts can design information-rich rewards or rules, while laymen can only give simple instructions, so how to integrate human expert knowledge in multiple levels to better guide the learning of machines?

\item \textbf{Poor demonstration data}. In terms of CC hybrid intelligence, all methods require a significant amount of data, which consumes substantial time and resources, and may be limited to human initiative. Therefore, when the quality of demonstration data is poor, such as a lack of diversity in demonstration data or a lack of demonstration data in the face of new scenarios, the model performance is lower. So, how can the model actively explore and discover new solutions, adapt to poor demonstration data, and ultimately improve performance?

\item \textbf{The unified learning paradigm}. In terms of CHI, different paradigms such as continuous learning paradigm and adaptive evolution paradigm have been proposed to improve human cognitive abilities, machine learning abilities, and adaptive coordination between humans and machines. However, these approaches have not yet been collectively learned within a unified framework and are dependent on expert knowledge. Therefore, how to adaptively integrate explicit, implicit, and other multi-perspective knowledge of humans and machines under the same framework without relying on expert knowledge to build a more intelligent, flexible, and adaptive human-machine collaborative system, to achieve mutual enhancement, co-evolution, and sustained collaborative symbiosis between humans and machines?

\end{itemize}


\section{Conclusion}

This paper has presented crowd knowledge transfer (CrowdTransfer), a new concept of knowledge transfer for AIoT community, which leverages the prior knowledge learned from a crowd of AIoT agents to solve challenges faced by most AIoT scenarios, such as constrained resources, dynamic environments, and incremental tasks. 
Layered on traditional transfer learning, CrowdTransfer aims to facilitate self-learning, self-adaptive, and continuous-evolving AIoT agents for a variety of AIoT applications. 
In this paper, we have clarified the main characteristics of CrowdTransfer from the perspective of crowd intelligence, and introduced four transfer modes inspired by biological communities: \textit{derivation mode}, \textit{sharing mode}, \textit{evolution mode}, and \textit{fusion mode}.  
Subsequently, we have presented an overview of CrowdTransfer, and introduced the recent advances of knowledge transfer methodologies from three aspects, including intra-agent knowledge transfer, decentralized inter-agent knowledge transfer, and centralized inter-agent knowledge transfer.
Moreover, we investigate some AIoT applications in various significant domains that could benefit from CrowdTransfer, such as human activity recognition, urban computing, connected vehicles, multi-agent system, and smart factory.

Based on our thorough analysis of existing knowledge transfer studies, we have given our discussion of CrowdTransfer for open issues and future directions. 
First, the cognitive foundation is the fundament to understand and explore the theoretical guidance for the crowd knowledge transfer. 
Second, the transferability measurement mechanisms should be studied to evaluate what knowledge is transferable to avoid negative transfer.
Third, some real-world issues need to be considered to improve the performance of CrowdTransfer for practical AIoT scenarios, including learning in resource-constrained AIoT devices, data security, and continuous crowd Knowledge transfer and evolution.
Finally, the fusion of human and machine intelligence could further enhance the success of CrowdTransfer for complicated AIoT applications.






\bibliographystyle{IEEEtran}
\bibliography{ref}

\end{document}